\pdfoutput=1
\documentclass{article}

\usepackage[eandd, final]{neurips_2026}

\usepackage[utf8]{inputenc}
\usepackage[T1]{fontenc}
\usepackage{hyperref}
\usepackage{url}
\usepackage{booktabs}
\usepackage{amsfonts}
\usepackage{nicefrac}
\usepackage{microtype}
\usepackage{xcolor}
\usepackage{graphicx}
\usepackage{subcaption}
\usepackage{siunitx}

\title{Sophon: A Procedural Diagnostic for Spatial Reasoning in Vision-Language Models}

\author{%
  J. Zach~Gazak\thanks{Corresponding author: \texttt{jzg@alum.mit.edu}} \qquad
  Ryan~Swindle \qquad
  Justin~Fletcher \\[6pt]
  Exoptica, Kihei, HI 96753, USA
}

\begin{document}

\maketitle

\begin{abstract}
Open-weight Vision-Language Models (VLMs) score 85--94\% on existing spatial benchmarks (BLINK Spatial, What's Up CLEVR) yet collapse to chance on simple geometric questions when object, scene, and texture priors are stripped away. We introduce and open-source Sophon, a procedurally generated diagnostic of paired colored point-cloud stimuli across three coordinate systems (1D linear, 2D Cartesian, 2D polar) and four question types (binary, categorical, comparative, quantitative). Across four open architectures (LLaVA-NeXT, LLaVA-OneVision, Qwen2.5-VL, Idefics3), accuracy clusters at 47--52\% while frontier models (Opus 4.6, Gemini 2.5 Pro, GPT-5.4) clear 77--83\% and humans reach 84--94\%. We use linear probes and low-rank adaptation (LoRA) to trace the open-weight failures end-to-end. Probes of the vision encoder (ViT) and language decoder (LLM) recover Sophon spatial information at 83--85\% and 80--86\% accuracy, respectively, and the language tower's reasoning circuit answers text-only controls (point clouds as sets of coordinates) at +40 pp over visual stimuli. Language tower per-layer attention attribution localizes visual integration to mid-stack layers. Yet, at readout, models elevate the correct answer-set tokens but select among them randomly. We localize the mechanism by tuning individual and grouped components across every architecture's stack, including the vision tower (ViT workspace and reader), the projector, and the language tower (LLM workspace and reader). These experiments demonstrate that this mid-stack workspace holds circuitry capable of organizing the spatial signal so that the frozen reader can decode it. LoRA fine-tuning adjusts that circuitry such that 7B-class VLMs answer Sophon spatial reasoning questions at frontier aggregate accuracy. This work suggests that these models do not have a capability gap for spatial reasoning, but instead suffer from a routing bottleneck that targeted fine-tuning repairs.

\end{abstract}\section{Introduction}
\label{sec:intro}

\begin{figure}[t]
\centering
\begin{subfigure}[t]{\linewidth}
\centering
\includegraphics[width=\linewidth]{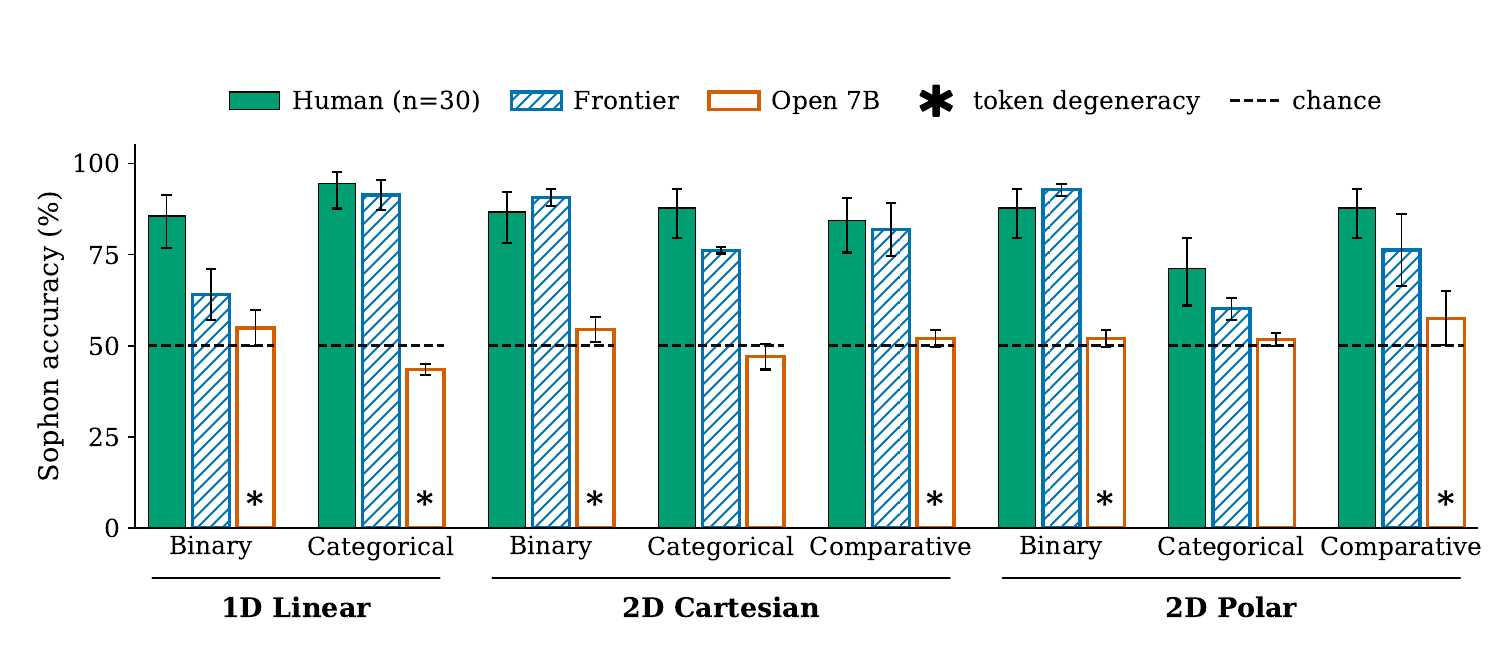}
\caption{\textbf{Three tier diagnostic gradient.} Untrained humans ($n{=}30$ Prolific, green) anchor at $71{-}94\%$.  Frontier models (Opus~4.6, Gemini~2.5~Pro, GPT-5.4) match or slightly exceed humans on the two binary cells and trail on the rest.  Open~7B~VLMs (Qwen2.5-VL-7B, LLaVA-NeXT-7B, LLaVA-OneVision-7B) cluster near chance. Cells marked $\ast$ have at least one open model emitting a constant token on $\geq$$85\%$ of items.}
\label{fig:three-tier-panel-a}
\end{subfigure}

\begin{subfigure}[t]{\linewidth}
\centering
\includegraphics[width=\linewidth]{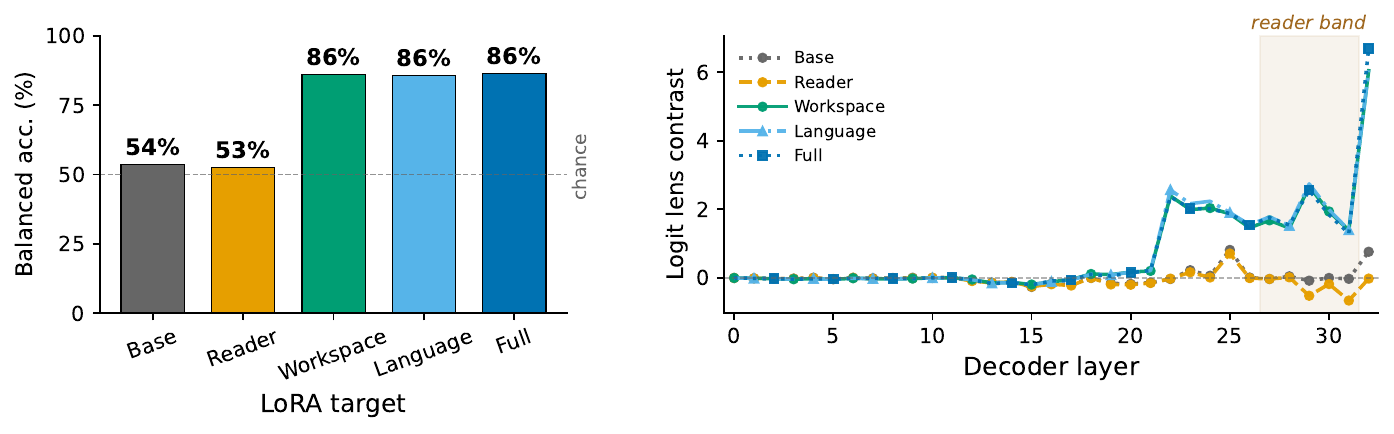}
\caption{\textbf{The fine tuning gain.}
\emph{Left:} Sophon accuracy on Idefics3-8B for five Low Rank Adaptation (LoRA) targets: \textbf{Base} (none), \textbf{Reader} (reader band only), \textbf{Workspace} (workspace only, reader frozen), \textbf{Language} (full language tower), and \textbf{Full} (both towers). \emph{Right:} per layer logit gap (GT vs.\ opposite direction) at the answer position. Reader tuning installs no contrast and recovers none of the gain. Workspace tuning recovers all of it, and its reader-band contrast appears even though those layers were never updated. Per-architecture traces in App.~\ref{app:logit-lens-cross-arch}.}
\label{fig:three-tier-panel-b}
\end{subfigure}

\caption{\textbf{The diagnostic gap and its mechanism.} (a) Despite $85{-}94\%$ on existing spatial benchmarks, open models show no spatial competence on Sophon while untrained humans solve with high accuracy. (b)~Ablation localizes the fine tuning mechanism upstream: the reader-band discriminative contrast at the readout is a downstream effect of upstream workspace change (\S\ref{sec:mechanism}).}
\label{fig:three-tier}
\end{figure}
\vspace{-0.9em}

Modern VLMs are increasingly deployed on imagery diverging from common scenes (medical, microscopy, scientific instruments, astronomy), where the spatial reasoning is less dependent upon object semantics (e.g., ``is the
cup left of the book'') and more dependent upon the rotation and scale alignment of comparatively sparse image features.  Recent work has reported that frontier VLMs fail at these tasks, even as they answer naturalistic spatial benchmarks with near human performance~\citep{wu2023gpt4vmedical,kuckreja2024geochat,tong2024eyes,rahmanzadehgervi2024blind}.

Without doubt, the visual skills of modern VLMs are revolutionary. But the next stage of vision must be adaptability. In distribution statistics get you only where those statistics apply, while genuine perception of scale, rotation, and alignment may enable models to solve image based tasks with more general geometric underpinnings, including domains the training corpus never sampled.

Open VLMs now report $85{-}94\%$ on curated spatial benchmarks such as What'sUp \citep{kamath2023whatsup} and BLINK \citep{fu2024blink}. Yet these same models fail in uninformative ways when the task is stripped of semantic scaffolding.  For example, Qwen2.5-VL-7B ($88.8\%$ on BLINK Spatial, $93.6\%$ on What'sUp~CLEVR) answers ``counterclockwise'' for every one of $200$ rotation questions we ask about colored point clouds, and GPT-4o~\citep{openai2024gpt4o} reports a rotation magnitude of $45^\circ$ for $98\%$ of rotation questions.  We argue that existing spatial benchmarks remain solvable through non-spatial cues, such as object identity, scene layout priors, and language co-occurrence \citep{liu2023vsr},
paralleling the texture bias and shortcut learning failures documented in CNNs \citep{geirhos2020shortcuts,hermann2020shapes}.  As such, failures may confound weaknesses in vision encoding, language reasoning, and their interface, while a model
that has successfully learned ``cup-on-book'' co-occurrence statistics scores well without performing the underlying spatial reasoning task.

We introduce Sophon\footnote{We name the benchmark Sophon after the sophons of Liu Cixin's \emph{Three Body Problem}: protons unfolded from higher dimensions onto lower dimensional surfaces in order to make their structure accessible.}, a procedurally generated geometric benchmark designed to isolate spatial perception from all non-spatial cues. Stimuli are pairs of colored point clouds that differ by a single geometric transformation drawn from horizontal and vertical translation, radial scale, and rotation.  Questions ask directly about that transformation (binary alignment, categorical direction, comparative magnitude, quantitative estimation) in three coordinate systems (1D, 2D Cartesian, and 2D polar).  The tasks contain no natural objects to recognize, no scenes to leverage, and no texture cues.

Beyond introducing Sophon, contributions are as follows:  we establish \emph{solvability}, document a \emph{diagnostic gap with a readout-routing bottleneck}, we \emph{localize the fine-tuning mechanism}, and show that \emph{fine-tuning routes existing perception} instead of enhancing perception.  We expand below:

\textbf{Solvability:} three 2026 flagships (Opus~4.6~\citep{anthropic2026opus46}, Gemini~2.5~Pro~\citep{google2026gemini25pro}, GPT-5.4~\citep{openai2026gpt54}) cluster at $77{-}83\%$ on categorical/binary/comparative cells, with Opus~4.6 and GPT-5.4 at scale MAE~$0.06$ and Opus~4.6 polar scale Pearson~$r{=}0.86$, ruling out prompt format idiosyncrasies and making open VLM failure diagnostic. Further, untrained humans ($n{=}30$) answer Sophon questions at $84{-}94\%$.

\textbf{The diagnostic gap with a readout-routing bottleneck}: open VLMs collapse to chance while frozen encoder probes recover spatial content at $83{-}85\%$, a text only control adds $40$\,pp at the decoder, and layer-wise probes preserve the signal through every decoder layer at $80{-}86\%$.  Each diagnostic suggests that these models are capable of Sophon tasks in isolation.  The failure is the routing of a spatial signal to the answer token.

\textbf{Location of the fine-tuning mechanism}: a logit lens trace shows a discriminative contrast emerging at late-stack (``reader'') layers in fine tuned models.  Freezing the reader, however, and training only the upstream (``workspace'') layers recovers $96{-}102\%$ of the Sophon LoRA gain in every architecture tested. Training only the reader captures none of it ($-3\%$ to $+17\%$).  The reader contrast is the readout of upstream workspace change.

\textbf{Fine-tuning routes existing perception}: Low-Rank Adaptation (LoRA) on Sophon improves external benchmarks (BLINK $+4.4$\,pp at 1k) while encoder probes drift $<1$\,pp and chain-of-thought (CoT) scene descriptions remain wrong.  The gain is answer routing, not improved perception. Further, that routing is narrow. At matched supervision budget, rotation-only training does not move rotation off chance, and Cartesian-only training transfers zero accuracy to polar questions.

\section{Related Work}
\label{sec:related}

\paragraph{Spatial reasoning benchmarks for VLMs.}
What'sUp~\citep{kamath2023whatsup}, BLINK~\citep{fu2024blink}, and VSR~\citep{liu2023vsr} probe spatial prepositions on natural and CLEVR style~\citep{johnson2017clevr} images, and a second wave targets 3D and embodied
settings~\citep{chen2024spatialvlm,cheng2024spatialrgpt,tong2024cambrian,ma20243dsrbench}. Across these benchmarks, images contain recognizable objects whose scene context constrains plausible spatial answers, so genuine perception is hard to separate from object identity and scene layout priors.  Sophon is complementary, stripping all non-spatial cues, trading naturalism for isolation (comparison in Appendix~\ref{app:benchmark-comparison}).

\paragraph{VL architectures} Contemporary VLMs follow an encoder--projector--decoder pattern.  We evaluate \textbf{LLaVA-NeXT} \citep{liu2023llava} (CLIP~\citep{radford2021clip} $+$ Mistral-7B~\citep{jiang2023mistral}), \textbf{LLaVA-OneVision}~\citep{li2024onevision} (SigLIP~\citep{zhai2023siglip} $+$ Qwen2), \textbf{Qwen2.5-VL}~\citep{bai2024qwen25vl} (windowed attention ViT $+$ 2D rotary embeddings~\citep{su2024roformer}), \textbf{Idefics3}~\citep{laurencon2024building} (SigLIP $+$ Llama-3.1-8B-Instruct~\citep{llama31_8b_instruct}) and frontier closed weight models. 

\paragraph{VL probing.} Prior work attributes visual-grounding differences to connector design via architecture ablations~\citep{mckinzie2024mm1,cha2024honeybee}; linear probing instead locates information layer-by-layer~\citep{alain2016probes}. Eyes Wide Shut~\citep{tong2024eyes} traced VLM failures to CLIP-blind encoder representations.  Encoder probes recover Sophon tasks at $83{-}85\%$ pooled ($97{-}100\%$ on translation and scale) across CLIP, DINOv2~\citep{oquab2024dinov2}, and SigLIP~\citep{zhai2023siglip}, yet end-to-end VLMs collapse to chance.  In this work, the failure lies downstream of the encoder, between representation and answer selection in the language tower. 

\paragraph{Shortcut learning, chain of thought, and low-rank adaptation.} Shortcut learning~\citep{geirhos2020shortcuts} is well documented across model class. Convolutional networks default to texture rather than shape~\citep{hermann2020shapes}, and language models default to lexical-overlap heuristics~\citep{mccoy2019hans}. The same dynamics appear in VLMs: \citet{vo2026vlms} attribute $75.7\%$ of VLM errors to bias-aligned outputs, which our constant taxonomy operationalizes. Sophon demonstrates that LoRA installs a coordinate system specific vocabulary routing that does not transfer across coordinate systems and changes readout significantly more than the underlying representation (\S\ref{sec:exp4}). We diagnose this with two-turn chain-of-thought prompting~\citep{wei2022cot} and parameter-efficient fine-tuning~\citep{hu2022lora,mangrulkar2022peft}, with all evaluations served via vLLM~\citep{kwon2023vllm}.

\paragraph{Concurrent work on the encoder-decoder gap.} Recent work converges on readout and handoff failures in VLMs across different task regimes.  Our nearest neighbor is \citet{tartaglini2025diagnosing}, who study scatter-plot understanding on the same OneVision family and localize a vision--language \emph{handoff} bottleneck.  There, individual point coordinates are decodable in the encoder but not surfaced to the language model. Sophon asks a different question using similar stimuli.  By asking only for the direction of a transform between two whole point sets, we demand a global relational judgment that never requires recovering a coordinate. This is the population regime \citet{tartaglini2025diagnosing} flag as unexplained.  Supplying ground-truth coordinates \emph{hurts} their ensemble tasks, yet \emph{solves} Sophon's (\S\ref{sec:exp3}), because Sophon's downstream computation is trivial.  We find the missing component is readout of the visual signal, not the computation over it. The early-vs-late LoRA dissociation (\S\ref{sec:exp4}) places fine-tuning's recoverable gain at the decoder readout, complementary to their per-point handoff.  \citet{fu2025hidden} report large encoder--VLM gaps on vision-centric tasks; \citet{liu2025visual} show fine-tuned visual representations underperform frozen SigLIP; \citet{kancheti2026faithful} measure a $26.1\%$ description--answer contradiction rate matching Sophon's $25{-}30\%$ and patch it with reinforcement learning. Where Kancheti et al.\ show an RL fix exists, we show what \emph{standard supervised fine-tuning} installs in its place: a small downstream readout projection driven by upstream representational change (\S\ref{sec:mechanism}). Sophon adds the procedural, shortcut-resistant stimulus (rendering-invariance controls, App.~\ref{app:rendering-battery}); the factorial coordinate-system $\times$ transformation design; and the perceiver/constant/few-anchor taxonomy separating residual-stream perceptual failure from text-anchored failure on continuous-magnitude cells. Procedural stimulus benchmarks in cognitive science share our design philosophy \citep{budny2025visual,kosoy2025decoupling,salbench2025}, though none localize failures within the VLM stack.

\section{The Sophon Benchmark}
\label{sec:benchmark}

\begin{figure}[t]
\centering
\includegraphics[width=\linewidth]{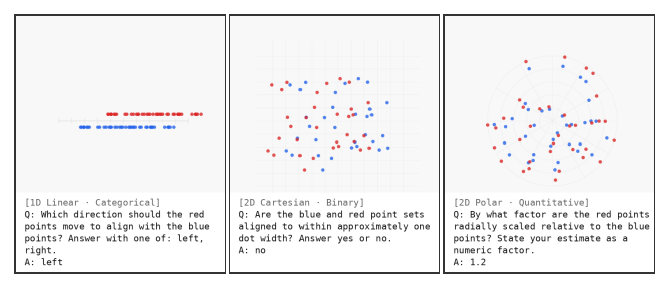}
\caption{Sample Sophon stimuli, one per (question type, coordinate system) cell: 1D linear / categorical (left), Cartesian / binary (center), polar / quantitative (right). Each panel shows its coordinate system and ground truth transform parameter. No objects, scenes, or textures appear in any stimulus (Appendix~\ref{app:examples}).}
\label{fig:sophon-examples}
\end{figure}

Sophon consists of procedurally generated point cloud images paired with spatial reasoning questions. Every image shows two overlaid sets of points,
blue and red, that share a common underlying pattern but differ by one
geometric transformation. Questions ask about that transformation in several
formats. Figure~\ref{fig:sophon-examples} illustrates three examples.

\paragraph{Coordinate Systems and Transforms:}
Sophon covers three coordinate systems, \textbf{1D linear}, \textbf{2D Cartesian} (x-y grid), and \textbf{2D polar} $(r,\theta)$.  Each system supports translation, rotation (excluding linear), and scale transforms. Images render at $512\times 512$ on a neutral gray background with axis grids.  Renders contain no content that could cue object identity or scene context.

\paragraph{Question Types:}
For each (coordinate system, transform) pair, we generate four question
types.  \textbf{Binary}: ``Are the blue and red point sets aligned?''
\textbf{Categorical}: ``Which direction should the red points move to
align with the blue points?'' with restricted 2-way or full 6-way
vocabulary (\S\ref{sec:vocab-ablation}). \textbf{Comparative}: ``Which image shows better alignment between the blue and red point sets, A or B?'' \textbf{Quantitative}: ``Approximately how many degrees should the red points rotate to align with the blue points?''

\paragraph{Positive control:}
Sophon is solvable end-to-end by three current frontier VLMs (Opus~4.6,
GPT-5.4, Gemini~2.5~Pro at $77{-}83\%$; Opus polar scale Pearson $r{=}0.86$),
and its spatial content is recoverable above $85\%$ by linear probes on
three independent encoder families (CLIP, DINOv2-large, SigLIP-large;
Appendix~\ref{app:probes-external}). Three providers and three encoder
regimes converging rules out prompt format and stimulus novelty as
confounds.

Sophon is procedurally generated; we release the generator (MIT) and
a pre-rendered evaluation suite (CC-BY 4.0), with eval seeds (42, 43,
44) disjoint from training seed 100 to prevent leakage (datasheet in
Appendix~\ref{app:datasheet}; leaderboard policy and URLs in Appendix~\ref{app:release}).

\vspace{-0.5em}
\section{Experimental Setup}
\label{sec:setup}
\vspace{-0.4em}

\paragraph{Models.}
We evaluate five closed-weight frontier models (Claude~Opus~4.6, Gemini~2.5~Pro, GPT-5.4, GPT-4o, and  Grok-4.20~\citep{xai2026grok420}) via a unified API, and five open VLMs of varying scale and family (Qwen2.5-VL-7B,  Qwen2.5-VL-32B-AWQ, LLaVA-NeXT-7B-Mistral, LLaVA-OneVision-7B, Idefics3-8B) served locally via vLLM. Fine-tuning uses LoRA with the PEFT library at rank 32 (alpha 64, dropout 0.05). The standard budget is 1{,}000 Sophon samples across question types.  We run a volume sweep (500--10{,}000) on Qwen2.5-VL.

\vspace{-0.35em}
\paragraph{Component ablations.}
We decompose each VLM into five components: the vision tower's \emph{workspace} and \emph{reader}, the \emph{projector},  and the language tower's \emph{workspace} and \emph{reader}. Each tower splits at fixed layer boundaries (for OneVision, L0--22 /  L23--27 in the decoder and L0--22 / L23--25 in SigLIP, all per-architecture bands are in App.~\ref{app:bands}). The unembedding is always frozen. LoRA targets single components (\textsc{Vision}, \textsc{Projector}, \textsc{Language}), the language  tower's bands alone (\textsc{Workspace}, \textsc{Reader}), both towers' bands together (\textsc{Workspaces},  \textsc{Readers}), and everything (\textsc{Full}). We also vary the training distribution (Cartesian only, polar only,  full mixed) for the cross-coordinate transfer experiments. The mechanism ablations (\S\ref{sec:mechanism}) contrast \textsc{Reader}, \textsc{Workspace}, \textsc{Language}, and \textsc{Full}.

\vspace{-0.35em}
\paragraph{Evaluation.}
Each run uses the shared Sophon scoring pipeline. Categorical, comparative, and binary tasks are scored as exact-match accuracy on balanced ground-truth splits, with chance level reported per cell. Quantitative tasks report mean absolute error (MAE) and Pearson correlation against ground-truth magnitudes. We call a model a \emph{perceiver} on a quantitative cell when its parsed answers track the stimulus (Pearson~$r \geq 0.5$, top-$1$ answer mass $<50\%$). \emph{Constant}, \emph{few-anchor}, and \emph{noisy} responders score on MAE while emitting answer sets decoupled from the stimulus (\S\ref{sec:quant-taxonomy}). Fine-tuned models are evaluated after merging LoRA weights into the base checkpoint.\footnote{vLLM's adapter serving silently omits vision-encoder LoRA layers.  When served unmerged, every vision-side condition degrades to its language-only counterpart.}

\vspace{-0.35em}
\paragraph{Probes.}
We extract hidden activations from every layer of five vision encoders spanning four pretraining regimes: CLIP (LLaVA-NeXT's encoder, supervised contrastive), Qwen2.5-VL's ViT (custom multimodal supervised), DINOv2-large (self supervised), SigLIP-large (sigmoid contrastive), and the SigLIP-so400m vision tower extracted from LLaVA-OneVision (the same SigLIP-so400m architecture but with OneVision weights). For each encoder we train a linear probe per layer for coordinate system identification, direction classification, and magnitude regression on a train/test split of Sophon, and report the peak layer's accuracy. The same probe protocol runs on What'sUp and BLINK images for cross benchmark comparison, and on fine tuned encoders (CLIP post-1k; Qwen-VL ViT post-1k and post-10k LoRA) to test whether Sophon training shifts the encoder.

\section{Main Results}
\label{sec:results}

\subsection{Spatial Reasoning Ability is Unequal Within and Across Model Classes}
\label{sec:exp1}

Figure~\ref{fig:three-tier} (top panel) visualizes the three tier solvability gradient on Sophon. The full per-model, per-cell heatmap (with human and chance rows, against eight accuracy cells plus rotation MAE and scale MAE columns) is in Appendix~\ref{app:diagnostic-heatmap}, and the per-model Sophon vs external benchmark summary is in Appendix~\ref{app:per-model-summary}. Three current generation flagship models cluster at $\sim$$80\%$ Sophon accuracy on categorical/binary/comparative cells (Gemini 2.5 Pro $82.6\%$, GPT-5.4 $81.4\%$, Claude Opus 4.6 $77.2\%$, sample-weighted
across the eight cells), establishing the upper end of the solvability gradient. The previous generation GPT-4o sits notably lower ($74.8\%$),
and Grok-4.20 collapses to $49.8\%$ via the constant token (same answer, always) failure mode characterized below. Open 7B VLMs that achieve $85{-}94\%$ on external benchmarks like What'sUp and BLINK collapse on Sophon to similar chance adjacent territory, often with constant token failure modes.

\paragraph{Frontier solvability.}
Beyond the categorical cells, Opus~4.6 and GPT-5.4 reach scale MAE $\approx 0.06$, and Opus~4.6 polar scale Pearson $r{=}0.86$ (\S\ref{sec:quant-taxonomy}).  Grok-4.20 collapses by emitting a constant answer token under greedy decode (temperature$=$0), a mechanistically informative failure.  A temperature sweep recovers part of the gap (Appendix~\ref{para:temp-ablation}), and the remainder cannot be attributed for an API model, where serving-side factors are unobservable. Where the weights are open, the same phenomenon is diagnosable: Qwen2.5-VL-7B shows constant-token emission on rotation specs and does not recover under temperature variation.

\paragraph{Scale alone does not close the gap.}
Quadrupling decoder parameters within Qwen-VL ($7$B to $32$B-AWQ) moves Sophon mean accuracy by $+3.6$\,pp ($51.1\%$ to $54.7\%$), well below the frontier cluster. Quantization confounds the absolute level, but the $32$B-AWQ model is near ceiling on the cells the $7$B solves ($91.5\%$ linear translation, $\sim$$82\%$ Cartesian translation), so $4$-bit AWQ is not broadly crippling.  Rotation categorical stays at chance ($48.2\%$) and quantitative rotation error explodes (MAE $111^\circ$), which matches the performance pattern of the $7$B. This single $32$B-AWQ point suggests that scale alone is insufficient in this parameter regime.

\subsection{The Readout-Routing Bottleneck}
\label{sec:exp3}

In this work, we find that Sophon-class spatial representation is present and elevated but not routed to the answer token in open-weight 7B class VLMs.  The workspace does not write it in a form the frozen reader converts into an answer.  We call this the \emph{readout-routing bottleneck}.  We decompose this failure with five convergent probes: linear readout on frozen vision encoders, a text-only control, layer-wise probes through the language decoder, per-layer attention attribution, and two-turn chain-of-thought. Each isolates a different stage of the VLM encoder--projector--decoder--readout pipeline.

\begin{figure}[t]
\centering
\includegraphics[width=0.9\linewidth]{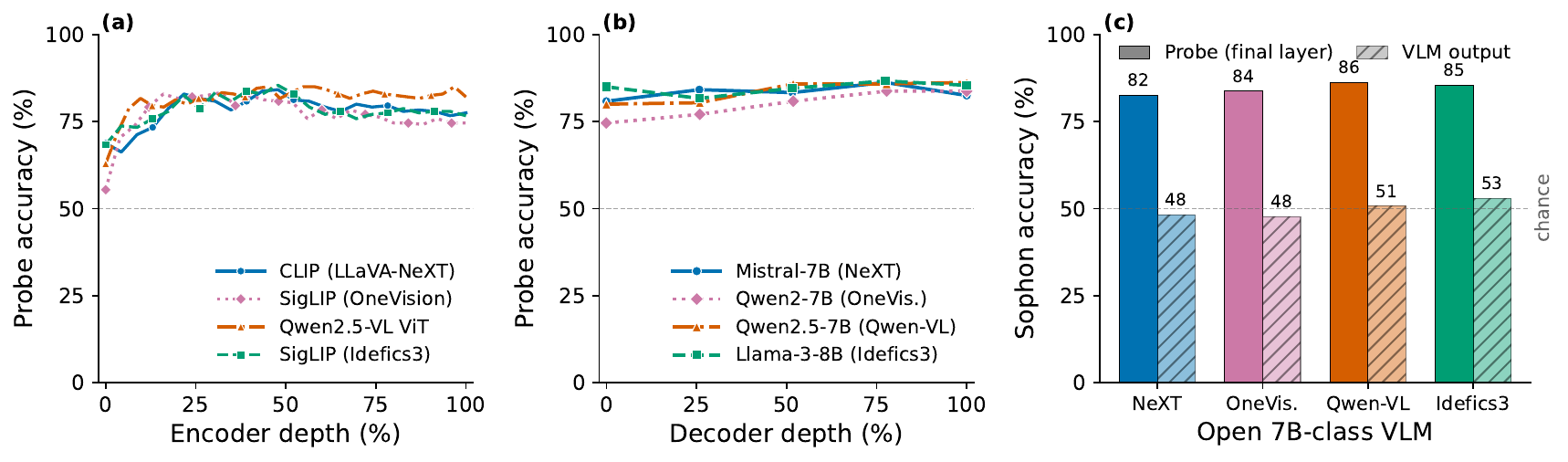}
\caption{\textbf{The readout-routing bottleneck: spatial content traverses
encoder and decoder, then collapses at the readout.}
(a)~Linear probe accuracy on Sophon direction classification at each
encoder layer.  Five frozen encoders cluster at $83{-}85\%$ in
mid-to-late layers regardless of pretraining objective (Appendix~\ref{app:probes-external}). (b)~Probe accuracy at the
answer emission position across four open architectures, sampled at
five depths each: $80{-}86\%$ at every layer, including the final
one. (c)~Final layer probe accuracy (solid) vs. end-to-end Sophon categorical accuracy (hatched) across the same four decoders. Every architecture carries the answer in its residual stream and emits at chance, a readout failure.}
\label{fig:probe-unified}
\end{figure}

\paragraph{Encoder probes: information is present.} Figure~\ref{fig:probe-unified} visualizes our bottleneck claim. Every contemporary vision encoder we test (CLIP, Qwen-VL ViT, DINOv2-large, SigLIP-large, SigLIP-so400m post-OneVision-training) reaches $83{-}85\%$ pooled direction probe accuracy in mid-to-late layers. The aggregate hides an important detail. Translation and scale probe at $97{-}100\%$, while rotation lags at $57{-}78\%$. Every base open VLM pairing one of those encoders with a language decoder collapses to $35{-}55\%$ on the same task. Post fine-tune
encoder probes drift less than a few points (Appendix~\ref{app:probes-architecture}).  The bottleneck is generic across encoder families.  While fine-tuning can remove the bottleneck, that repair is not on the encoder side (\S\ref{sec:exp4}).

\paragraph{Layer-wise decoder probes: the signal survives the language tower.}
We train per-layer linear probes for left vs right discrimination on
the answer emission position's hidden state across all four open VLM
architectures (LLaVA-NeXT, LLaVA-OneVision, Qwen2.5-VL-7B, Idefics3).
Probe accuracy reaches $80{-}86\%$ at the final decoder layer in every
architecture and is preserved within a few points across every
intermediate layer. A masking control that strips the answer set tokens
from the prompt shifts probe accuracy by less than $\pm 3$\,pp,
confirming the signal is genuinely derived from the visual Sophon stimuli rather than scaffolding from the prompt's offered choices. The spatial signal does not erode through the language tower, instead arriving intact at the answer-emitting position.

\paragraph{Attention attribution: visual integration concentrates in the language workspace.}
In Qwen2.5-VL-7B, per-layer attention from question text positions to
visual tokens peaks at L13/L14/L16 ($42{-}49\%$ mass) within a stable
L7--L18 band. The band is preserved across coordinate systems,
question types, and model performance regimes (from $0$--$94\%$
accuracy).  Attention to visual tokens at L13 actually \emph{increases}
on $0\%$-accuracy rotation. Vision tokens are integrated into the language workspace's residual stream. 

\paragraph{Text-only control: the decoder works.}
When LLaVA-NeXT is given the point coordinates as \emph{text} instead
of the image, binary accuracy rises from $47\%$ to $89\%$ and
categorical from $41\%$ to $85\%$ on the 1D linear spec
(protocol in Appendix~\ref{app:text-only-protocol}). The bypass
also lifts the frontier: Opus~4.6 moves from $76\%$ to $100\%$ on
1D linear binary and from $92\%$ to $100\%$ on 1D linear categorical
($+24$ and $+8$\,pp). And it holds at scale: Qwen2.5-VL-32B-AWQ
moves from $56\%$ to $95.8\%$ on Cartesian rotation binary, $+40$\,pp.

\paragraph{Chain of thought: vocabulary without grounding.}
Under a two-turn CoT protocol open VLMs produce fluent spatial
descriptions with \emph{correct terminology}, but often \emph{incorrectly describe} the
image content, with $\sim$$25{-}30\%$ \emph{contradiction} (description
correct, final answer disagrees). Opus 4.6's descriptions are
correct $84\%$ of the time with a $33\%$ two-turn contradiction rate
(Appendix~\ref{app:cot}), but rotation accuracy stays at $54\%$.
A single-turn control reproduces the pattern (Opus~4.6: contradiction
$56\%$ on rotation vs $4\%$ on translation, with the description
mentioning the GT direction $98\%$ of the time): the failure is a
property of readout, not workspace.

Encoder, decoder, and integration handle Sophon tasks in isolation, and
the spatial signal survives to the answer position. Base models already
elevate the correct answer set. On binary and categorical cells, the
ground-truth token is the model's top candidate when it answers
correctly and its median \emph{rank-2} candidate when it answers wrong.
It sits in the top-$10$ of a 32k--152k-token vocabulary for $100\%$ of
wrong samples in every architecture. Quantitative cells fail
differently (\S\ref{sec:quant-taxonomy}). The workspace holds the signal but does not write it in a form the frozen reader converts into an answer (\S\ref{sec:mechanism}). We hypothesize that standard VLM assembly, in which pretrained towers are joined by lightweight alignment training, leaves existing spatial capacity present but unexpressed.

\vspace{-0.5em}
\subsection{The Mechanism: Where the Fine-tuning Gain Lives}
\label{sec:mechanism}
\vspace{-0.3em}

If the failure is readout routing (\S\ref{sec:exp3}), where does
Sophon LoRA's gain live? We answer with a layer-wise logit lens at
the answer emission position and an ablation that restricts LoRA to
specific components of the stack (\S\ref{sec:setup}).

\paragraph{Logit lens: a discriminative contrast emerges near the readout.}
We project the answer position hidden state through the unembedding $W_U$ at every decoder layer and measure the logit gap between the ground truth direction token and its same axis competitor (e.g.\ ``left'' vs ``right''). In base models, within-set top-1 selection stays at chance ($48{-}51\%$) at every layer, in all four architectures. After Sophon LoRA fine tuning the gap rises late in the stack, reaching $+1.8$ to $+6.7$ at the final layer and pushing within-set top-1 accuracy to $92{-}98\%$ in the three architectures without a damaged axis. Qwen's exception is its trans\_y axis, which fine tuning degrades (Appendix~\ref{app:rendering-battery}). We call this the \emph{discriminative contrast}: a small but specific projection along the unembedding's vocabulary axis, emerging at the layers nearest the readout.

\paragraph{Ablation places the gain in the workspace.} 
The contrast appears in the reader band, but those layers do not need to be trained. The dissociation replicates across all four architectures on the balanced evaluation. \textsc{Reader} training recovers $-3\%$ to $+17\%$ of the \textsc{Language} gain. \textsc{Workspace} training recovers $96{-}102\%$ of it, with the reader band and $W_U$ frozen (Idefics3 $86.1\%$, LLaVA-NeXT $84.2\%$, LLaVA-OneVision $82.5\%$). On Qwen2.5-VL, \textsc{Workspace} ($70.8\%$) exceeds \textsc{Language} ($64.6\%$), so co-training the reader hurts there. Similarly, \textsc{Readers} (both towers' reader bands) stays at base level on both architectures with a vision reader band ($47.6\%$ OneVision, $56.2\%$ Qwen). $W_U$-only ablations agree, recovering $\sim$$10{-}19\%$ on Qwen (Appendix~\ref{app:fine-tuning}). The gain is upstream of the readout.

\paragraph{The contrast is a downstream readout, not the mechanism.}
Logit lens on the \textsc{Reader} checkpoints shows no contrast: gaps stay under $+0.7$ and within-set selection stays at chance ($48{-}51\%$) in all four architectures, indistinguishable from base. Training the reader band alone installs nothing. The \textsc{Workspace} checkpoints show the opposite: a full-strength contrast ($+2.5$ to $+6.0$, within-set $90{-}94\%$) with the reader band never updated. On three architectures the trained conditions converge to the same contrast wherever LoRA is placed. On Qwen2.5-VL, the conditions that co-train the reader land \emph{below} \textsc{Workspace} ($+0.6$ and $+1.1$ vs $+2.5$), matching their behavioral deficit. The discriminative contrast emerges when the workspace is reorganized to feed the frozen reader, and training the reader never helps.

\paragraph{The vocabulary axes are not the mechanism either.}
At each layer we compare two directions. The first is the probe direction, along which the spatial signal is linearly decodable. The second is the vocabulary axis, the unembedding difference $W_U[\text{left}]-W_U[\text{right}]$ (also up/down). Pushing the residual along this axis moves answer logits with no further computation. If fine-tuning worked by aligning the signal with the output vocabulary, these would converge. They stay $81{-}88^\circ$ apart (cosine at most $0.16$) across all four architectures and all conditions, at every layer, against a random baseline of $90^\circ \pm 1^\circ$ in these $3584$--$4096$-dimensional spaces. The deviations are real but do not track behavior. The largest alignment in the condition grid belongs to a \textsc{Reader} cell that selects at chance, while models selecting at $94{-}98\%$ sit at cosines of $0.04{-}0.11$. Increasing contrast thus requires aligning the residual stream with pretrained circuitry deeper in the readout stack.

\vspace{-0.5em}
\subsection{Fine-Tuning Routes Existing Perception}
\label{sec:exp4}
\vspace{-0.3em}

We fine tune four open architectures (Qwen2.5-VL-7B,
LLaVA-NeXT-7B, LLaVA-OneVision-7B, Idefics3-8B) on Sophon data across
the component grid of \S\ref{sec:setup}, and sweep training volume on
Qwen ($500$--$10$k samples; $1$k elsewhere). These experiments produce
three findings consistent across architectures.  Full experiment
details are found in Appendix~\ref{app:fine-tuning}.

\paragraph{The language-tower gain dominates.}
End-to-end Sophon accuracy gains $+25$ to $+35$\,pp at 1k training,
while frozen encoder probes drift $<$$1$\,pp at 1k and $\sim$$3$\,pp at
10k, an order of magnitude smaller. \textsc{Vision} LoRA plateaus near $63\%$ on the mixed evaluation regardless of volume, and stays at base level on the balanced perceptual cells. \textsc{Language} LoRA
contributes the bulk of the gain. External transfer is real but narrow: Sophon trained Qwen peaks at $93.2 \pm 0.8\%$ on BLINK Spatial ($+4.4$\,pp over base, $n{=}3$ seeds) at $1$k and \emph{regresses} at $5$k and $10$k while Sophon itself saturates at $5$k, suggesting over specialization to
the point cloud distribution. Sophon training does not damage the external benchmarks either. Across the fine-tuned grid, \textsc{Workspace} and \textsc{Full} stay within sampling noise of base on BLINK and What'sUp in three of four architectures, and the What'sUp moves are mostly small gains. LLaVA-NeXT is the exception, trading $7$\,pp of BLINK for $+6$ to $+9$\,pp on What'sUp (Table~\ref{tab:external-clean}). Per-volume, per-architecture, and per-target tower numbers are in
Appendix~\ref{app:fine-tuning}.

\paragraph{The vision-side update carries no causal load.}
We test this by zeroing LoRA updates piecewise in the \textsc{Full} tuned models. Zeroing only the language reader band's update moves
accuracy by at most $1.4$\,pp, and two of the four stripped models score
marginally above intact \textsc{Full}. Zeroing only the language
workspace band's update returns every model to base ($52{-}55\%$, an
$18$--$36$\,pp drop), with the reader band and vision side updates still
in place. Zeroing the entire vision side update changes accuracy by at most $1.3$\,pp in all four architectures, and vision side band strips stay within the same $1.3$\,pp on the two architectures tested. The language workspace is the one live band, even in the jointly trained \textsc{Full} LoRA: its language component works standalone, and its vision component neither helps
alone nor is required. The additive test agrees:
\textsc{Workspaces} (both towers' workspace bands) matches
\textsc{Workspace} (the language workspace alone) within $2$\,pp in
every architecture, and \textsc{Vision} training by itself stays at
base level in three architectures ($48{-}53\%$). On Idefics3 it lands $6$\,pp above base, but final layer encoder probes are unchanged ($\pm 1.5$\,pp), so the gain is not new perception.

\paragraph{Cross-coordinate transfer is asymmetric.}
Polar to Cartesian transfer is substantial in every architecture
($+22{-}50$\,pp on Cartesian 2D), while Cartesian to polar transfer
is at or below seed noise on Qwen and modest on LLaVA
(Appendix~\ref{app:crosscoord-heatmap}). Cartesian only training
even drives Qwen polar categorical \emph{below} baseline ($51\%$
to $26\%$): the model emits Cartesian direction tokens on polar
questions, actively interfering. Upstream change shaped by the
training set's coordinate distribution is by construction
coordinate specific (\S\ref{sec:mechanism}).

\paragraph{Rotation does not move at this budget.}
$1000$ rotation only training samples leave Qwen rotation
categorical at chance ($51\%$) and degrade translation accuracy
(Cartesian $47\%$ to $30.5\%$). The rotation barrier holds at
matched supervision budget even when training targets it
exclusively (Appendix~\ref{app:rotation-discussion}). The MAE
``improvement'' from $22.3^\circ$ to $10.5^\circ$ is anchor
replacement, not estimation: $44\%$ of fine tuned answers fall on
three tokenizer artifact values closer to the GT mean
(\S\ref{sec:quant-taxonomy}). Sophon fine tuning routes image
features to direction words while leaving the encoder representation largely unmoved. External benchmarks cannot reveal this because they permit the same shortcut.

\vspace{-0.5em}
\subsection{Perception or Rendering Priors? An Appearance Battery}
\label{sec:rendering}

Sophon's procedural design allows stimuli with identical point geometry and modified appearance. We evaluate variants spanning marker shape, dot count and sparsity, fourteen color palettes, red/blue role reversal, grid removal, background, and render resolution. These zero-shot evaluations across all four architectures and three stimulus seeds are in Appendix~\ref{app:rendering-battery}. Geometry-preserving changes barely move the trained models, while base and \textsc{Reader} models sit at chance under every variant. The perception that fine-tuning surfaces is a reading of the geometry, not a rendering prior. The two dimensions that do matter are color and color role. Novel palettes cost $11{-}30$\,pp and role reversal $17{-}26$\,pp on three architectures, but remain above chance. Fine-tuning binds direction words partly to the training appearance, and the binding weakens away from it (\S\ref{sec:exp4}). The point-count sweep separates translation, readable from a single point, from scale, whose accuracy climbs steadily with dot count (Table~\ref{tab:battery-npts}).

\subsection{Human Baseline}
\label{sec:human-baseline}
\vspace{-0.3em}

Untrained human raters ($n{=}30$ post attention filter; protocol,
ethics, and full distributions in Appendix~\ref{app:human-baseline})
score $84.4{-}94.4\%$ on seven of eight categorical/binary/comparative cells, with polar categorical the lone outlier at
$71.1\%$. On rotation magnitude we score Pearson $r$ between answer
and GT angle (correlation separates estimation from anchor
replacement, \S\ref{sec:quant-taxonomy}). Humans reach $r{=}0.69$
on polar rotation, the only perceiver class ($r{\geq}0.5$) result
across $10$ models $\times$ $4$ quantitative cells. On Cartesian
rotation no system is a perceiver (humans $r{=}0.20$, GPT-5.4
$r{=}0.37$, others below). Human \emph{scale} is not a perception
measurement (free text prompt admitted multiple readings,
Appendix~\ref{app:human-baseline}).

\section{Analysis}
\label{sec:analysis}

\vspace{-0.5em}
\subsection{Vocabulary Distractor and Other Shortcut Probes}
\label{sec:vocab-ablation}
\vspace{-0.3em}

Three studies corroborate a shortcut interpretation.
\emph{(a) Vocabulary distractor}: offering full 6-way direction
vocabulary (left, right, up, down, clockwise, counterclockwise) instead of transform appropriate 2-way restriction
collapses non rotation specs for open and frontier models
(Opus~4.6 polar radial 84.8\% to 25.5\%; Cartesian rotation control
stays flat). Two mechanisms compose: rotation tokens (``clockwise'',
``counterclockwise'') appear in non rotation reasoning when offered,
and the last keyword wins parser surfaces them
(Appendix~\ref{app:parser}). This motivates our
restricted vocab default. \emph{(b) Range hints}: appending the GT
range to scale questions \emph{improves} MAE for anchoring open
VLMs but \emph{hurts} the one model estimating genuinely (Opus~4.6
polar scale $r{=}0.86 \to 0.31$). \emph{(c) Positional ordering}:
fine tuned LLaVA-NeXT moves from $53\%$ to $90\%$ on Cartesian
rotation when the correct answer is offered last; translation
specs stay at $0\%$ (Appendices~\ref{app:vocab-distractor},
\ref{app:range-hint}, \ref{app:vocab-order}).

\vspace{-0.5em}
\subsection{How Models Answer Numbers}
\label{sec:quant-taxonomy}
\vspace{-0.3em}

\begin{figure}[t]
\centering
\includegraphics[width=0.9\linewidth]{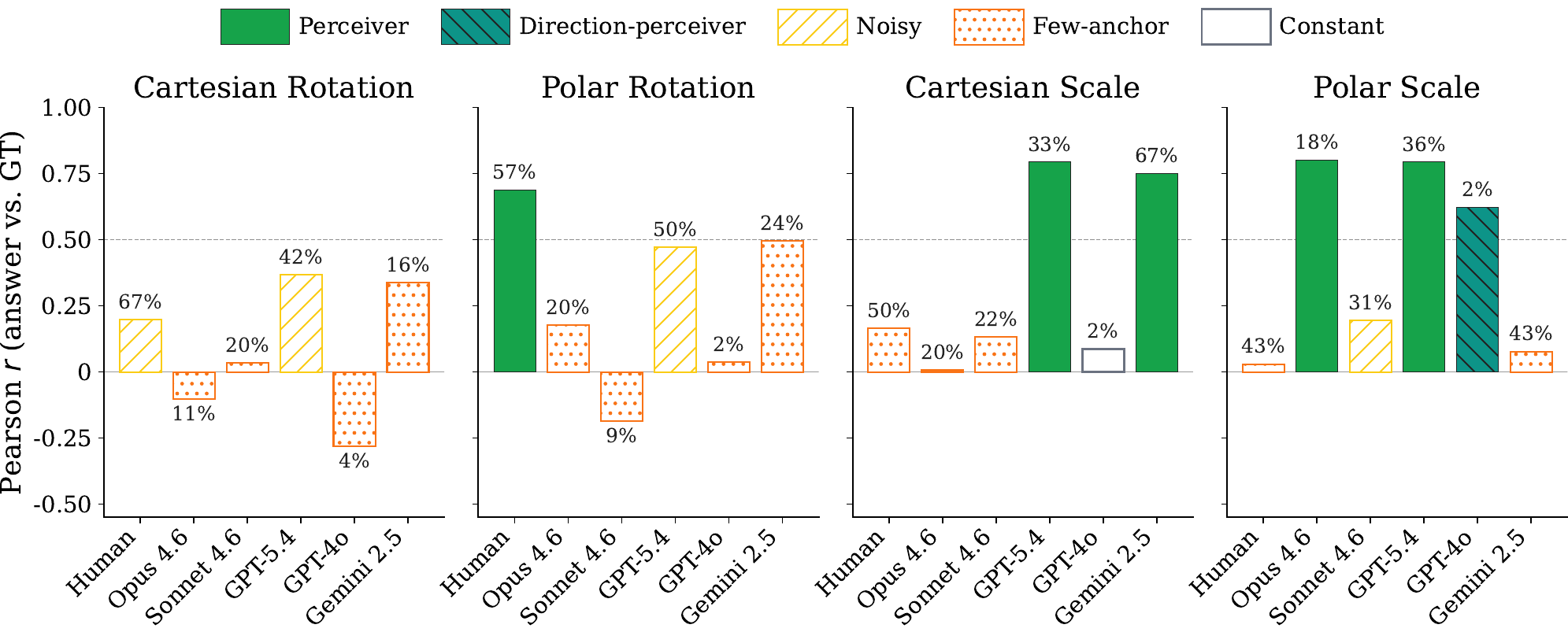}
\caption{\textbf{Quantitative answer distribution taxonomy.} Bar
height is Pearson $r$ between parsed answer and GT magnitude
($n{=}200$). Filled = stimulus engaged: \textit{Perceiver}
($r{\geq}0.5$, top-1${<}50\%$), \textit{Direction perceiver}
($r{\geq}0.5$, top-1${\geq}50\%$). Hatched/outline = engagement
failure: \textit{Noisy} ($r{<}0.5$), \textit{Few anchor}
(top-3${\geq}50\%$), \textit{Constant} (top-1${\geq}85\%$). Labels
above bars are answer uniqueness ($n_{\text{unique}}/n$). Human
scale $r$ reflects rater unit ambiguity. See all models version in
Appendix~\ref{app:quant-taxonomy}.}
\label{fig:quant-taxonomy}
\end{figure}

A low error MAE can come from genuine tracking or from emitting a
small anchor set whose mean sits near the GT mean. For continuous
cells, that distinction separates spatial perception from output
memorization. Figure~\ref{fig:quant-taxonomy} maps each (model,
cell) pair to a 5-class taxonomy on two axes: Pearson $r$ (does the
answer track the stimulus?) and top-$1$/top-$3$ mass.

\emph{(i) Scale perception is real but selective.} On polar scale
three flagships clear the perceiver threshold: Opus 4.6
($r{=}0.80$; under an explicit unit re prompt $r{=}0.70$, still
above the threshold), GPT-5.4 ($r{=}0.79$), and GPT-4o ($r{=}0.62$,
only 3 anchor values $\{0.5, 1.5, 2.0\}$ \emph{outside} the GT range
$[0.71, 1.30]$ but tracking GT direction, a
\emph{direction perceiver} distinct from continuous estimation).
On Cartesian scale only GPT-5.4 and Gemini reach perceiver; Opus 4.6 drops
to $r{=}0.01$. \emph{(ii) Rotation is hard for everyone}, with
humans holding a small lead: only humans clear the perceiver
threshold on polar rotation ($r{=}0.69$); GPT-5.4 reaches
$r{=}0.47$ (distributional but uncorrelated); on Cartesian rotation
no system perceives. The all models version
(Appendix~\ref{app:quant-taxonomy}) shows no open 7B--32B model
reaches perceiver or direction perceiver on any cell.

\vspace{-0.8em}
\section{Limitations}
\label{sec:limitations}
\vspace{-0.7em}

Sophon's abstract stimuli isolate spatial perception with priors
stripped, so our claims do not extend to naturalistic competence or to
non-spatial abilities, which we do not test. The open VLMs are
predominantly $7$B. The four architectures use MLP, patch-merger, and pixel-shuffle
projectors, so the routing bottleneck claim is
scoped to these and leaves cross-attention and Q-Former projectors
unevaluated. Seed coverage is partial ($n{=}3$ Qwen, $n{=}2$ LLaVA,
$n{=}1$ above 1k), and the human study is underpowered on quantitative
cells (Appendices~\ref{app:fine-tuning}, \ref{app:limitations}).

\vspace{-0.8em}
\section{Conclusion}
\label{sec:conclusion}
\vspace{-0.7em}

Open VLMs that score $85{-}94\%$ on existing spatial benchmarks answer
Sophon at chance, while frontier models reach $77{-}83\%$ and humans
$84{-}94\%$. Linear probes decode the spatial signal from encoder and
decoder at $80{-}86\%$, yet the models select among the answer tokens at
random (\S\ref{sec:exp3}). Fine-tuning places the repair in the language
workspace. LoRA on the workspace alone recovers $96{-}102\%$ of the
language tower gain with the reader band and $W_U$ frozen, and LoRA on
the reader band alone recovers $-3\%$ to $+17\%$
(\S\ref{sec:mechanism}). These results point to spatial supervision of
the language workspace as the intervention to pursue. At the 1k budget
the repair is coordinate specific, weakens under novel color palettes,
and does not reach rotation (\S\ref{sec:exp4}, \S\ref{sec:rendering}).

\begin{ack}
This work received no external funding. The authors declare no
competing interests. We thank the three anonymous reviewers and the
area chair for comments that shaped the camera-ready version.
\end{ack}

\bibliographystyle{plainnat}

\appendix
\makeatletter
\renewcommand{\thesection}{%
  \ifnum\value{section}>26 A\@Alph{\numexpr\value{section}-26\relax}%
  \else\@Alph{\c@section}\fi}
\makeatother

\section{Cross-Architecture Logit Lens}
\label{app:logit-lens-cross-arch}

\begin{figure}[!htbp]
\centering
\includegraphics[width=\linewidth]{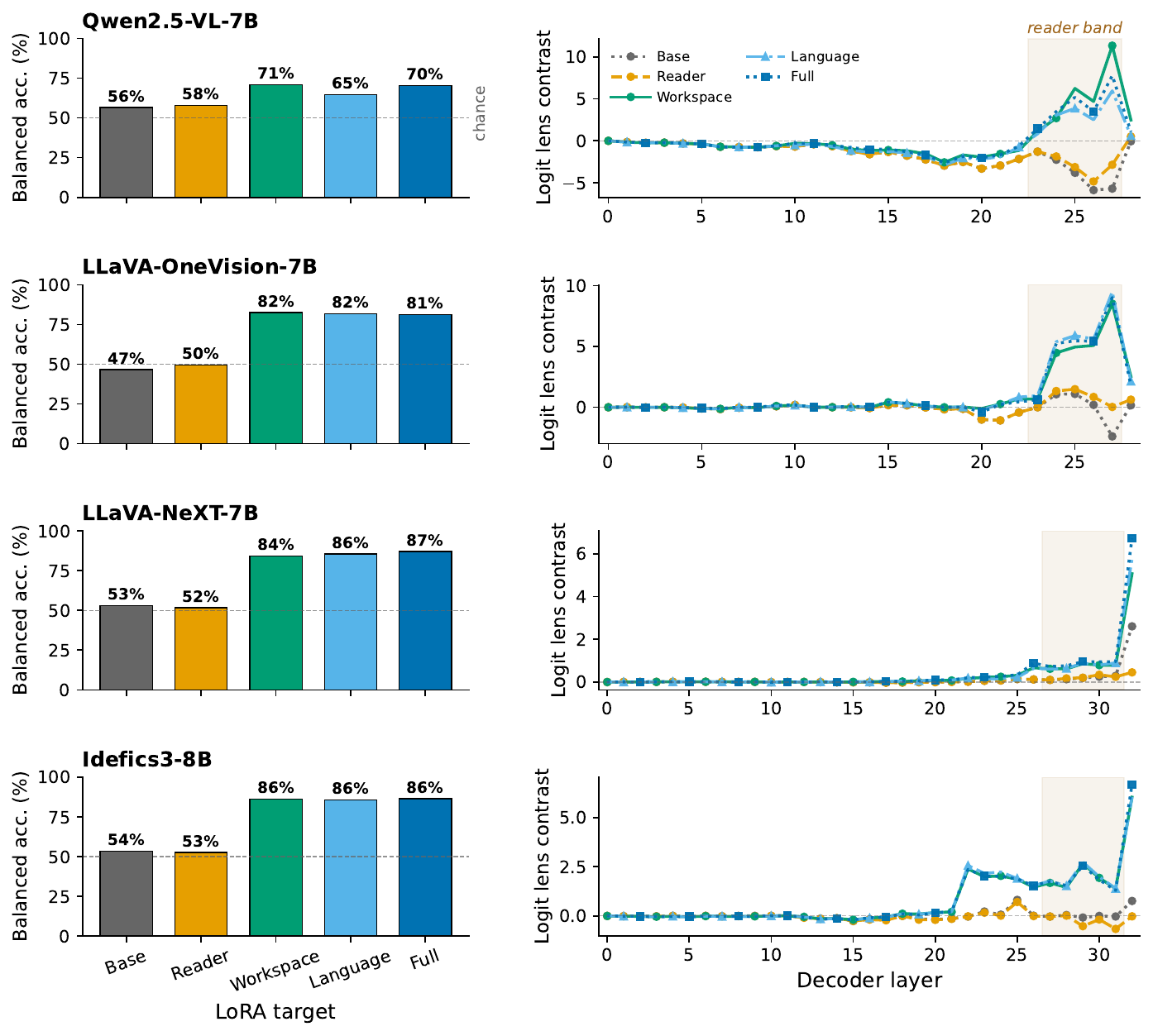}
\caption{\textbf{Per-layer GT vs competitor logit gap, five LoRA conditions, four architectures.} Each panel projects the answer position hidden state through the unembedding $W_U$ at every decoder layer and reports the mean gap between the ground truth direction token and its same axis competitor (e.g.\ ``left'' vs ``right'').  The shaded region marks the reader band. In every architecture \textsc{Base} and \textsc{Reader} sit at or below zero throughout, so reader-band training installs no contrast. The three trained conditions rise late in the stack and largely coincide, and \textsc{Workspace}'s contrast appears despite the reader band and $W_U$ being frozen. The reader band's contrast emerges when the workspace is reorganized to feed it (\S\ref{sec:mechanism}).}
\label{fig:logit-lens-cross-arch}
\end{figure}

We project the answer position hidden state through the unembedding $W_U$ at every decoder layer and measure the per sample logit gap between the ground truth direction token and its same axis competitor (e.g.\ ``left'' vs.\ ``right''). The trace is run for five conditions per architecture (\textsc{Base}, \textsc{Reader}, \textsc{Workspace}, \textsc{Language}, \textsc{Full}; \S\ref{sec:setup}). Figure~\ref{fig:logit-lens-cross-arch} shows these for the models tested.  Across all four, \textsc{Base} and \textsc{Reader} sit at or below zero through the decoder, and \textsc{Reader} stays flat inside the band it trained. The three trained conditions
rise late in the stack and largely coincide. \textsc{Workspace}'s contrast appears with the reader band and $W_U$ frozen. On Qwen2.5-VL the conditions that co-train the reader (\textsc{Language}, \textsc{Full}) land below \textsc{Workspace}, matching their behavioral deficit (\S\ref{sec:mechanism}).

\section{Sophon Stimulus Examples: Full Set}
\label{app:examples}

\begin{figure}[!htbp]
\centering
\includegraphics[width=\linewidth]{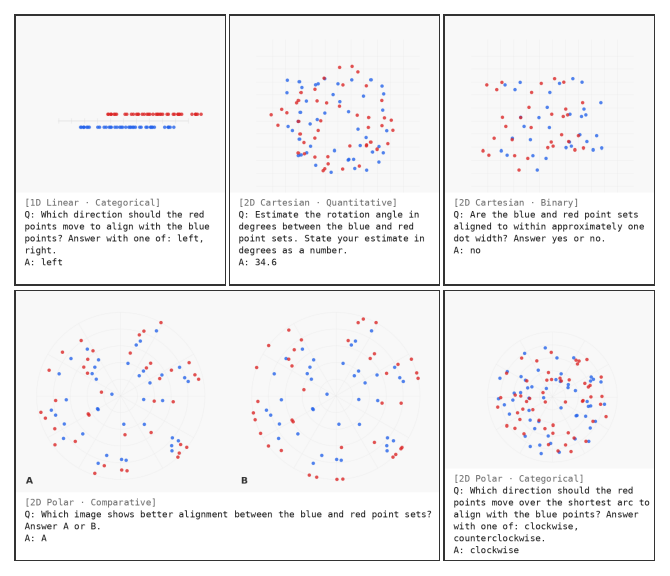}
\caption{Full Sophon sample stimuli. \emph{Top row, left to right}:
1D linear/categorical, Cartesian/quantitative, Cartesian/binary.
\emph{Bottom row}: polar/comparative (double wide A-B panel),
polar/categorical.}
\label{fig:sophon-examples-full}
\end{figure}

\begin{figure}[!htbp]
\centering
\includegraphics[width=\linewidth]{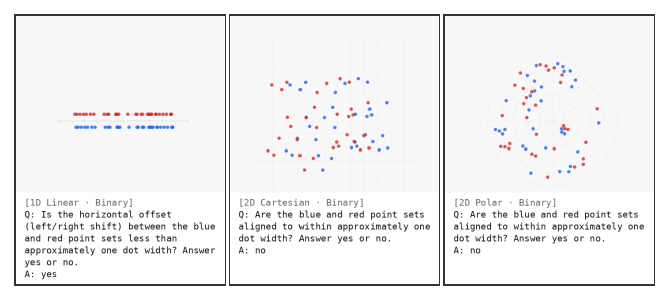}
\caption{Binary examples across all three coordinate systems.}
\label{fig:examples-binary}
\end{figure}

\begin{figure}[!htbp]
\centering
\includegraphics[width=\linewidth]{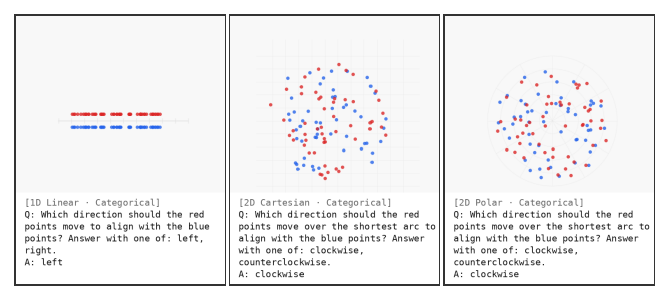}
\caption{Categorical examples across all three coordinate systems.}
\label{fig:examples-categorical}
\end{figure}

\begin{figure}[!htbp]
\centering
\includegraphics[width=0.92\linewidth]{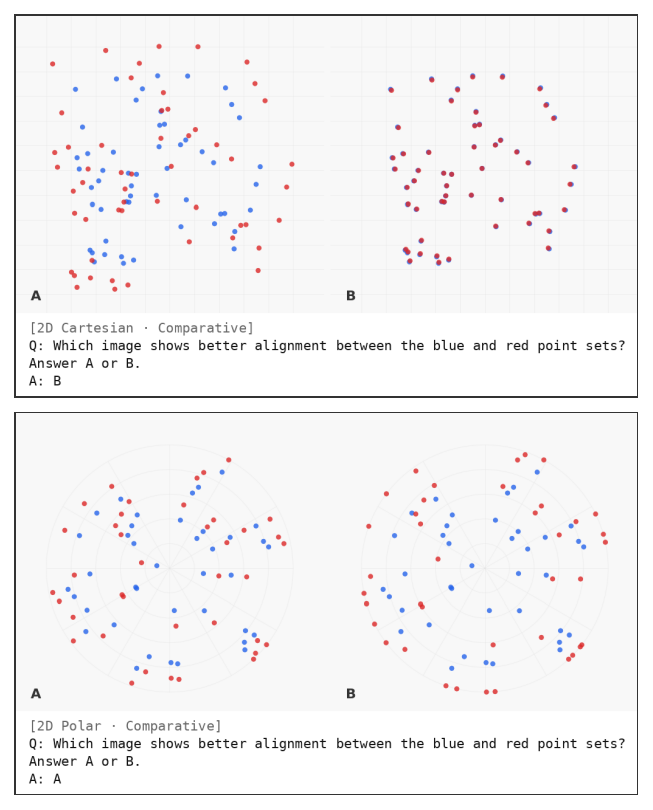}
\caption{Comparative examples across the two 2D coordinate systems (Cartesian and polar). 1D linear has no comparative cell by design (Figure~\ref{fig:diagnostic-heatmap}).}
\label{fig:examples-comparative}
\end{figure}

\begin{figure}[!htbp]
\centering
\includegraphics[width=0.92\linewidth]{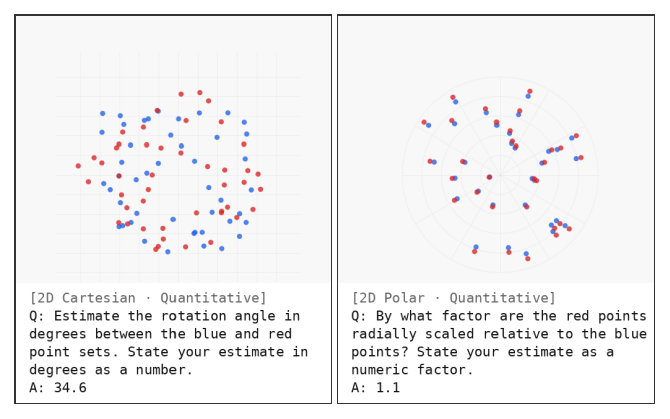}
\caption{Quantitative examples across the two 2D coordinate systems (Cartesian and polar). 1D linear has no quantitative cell by design (Figure~\ref{fig:diagnostic-heatmap}).}
\label{fig:examples-quant}
\end{figure}

The main text Figure~\ref{fig:sophon-examples} shows three sample
stimuli, one per (question type, coordinate system) cell. This
appendix collects the rest of the stimulus space. A full coverage
five tile layout is in Figure~\ref{fig:sophon-examples-full}, and
per question type preview rows across all three coordinate systems
are in Figures~\ref{fig:examples-binary}--\ref{fig:examples-quant}.
These serve reproducibility (visual verification of
what the procedural generator produces) and confound checking.
There are no objects, scenes, textures, or
recognizable shapes in any panel. Only the spatial relationship
between the blue and red point sets carries the answer.

\section{Per-Encoder Probe Results: DINOv2 and SigLIP}
\label{app:probes-external}

\begin{figure}[!htbp]
\centering
\includegraphics[width=0.96\linewidth]{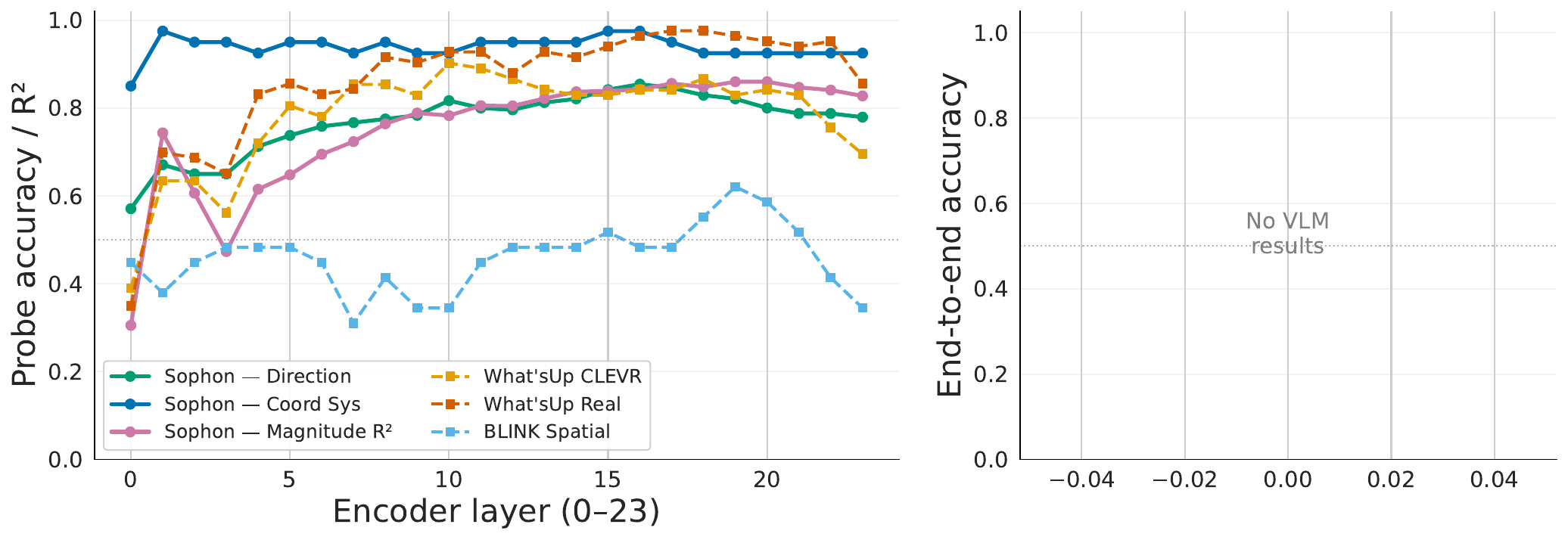}
\caption{DINOv2-large probes across Sophon and external benchmarks. Peak
accuracies on Sophon: direction 85.4\% (L16), coordinate system 97.5\%
(L1), magnitude $R^2{=}0.86$ (L20).}
\label{fig:probe-dinov2}
\end{figure}

\begin{figure}[!htbp]
\centering
\includegraphics[width=0.96\linewidth]{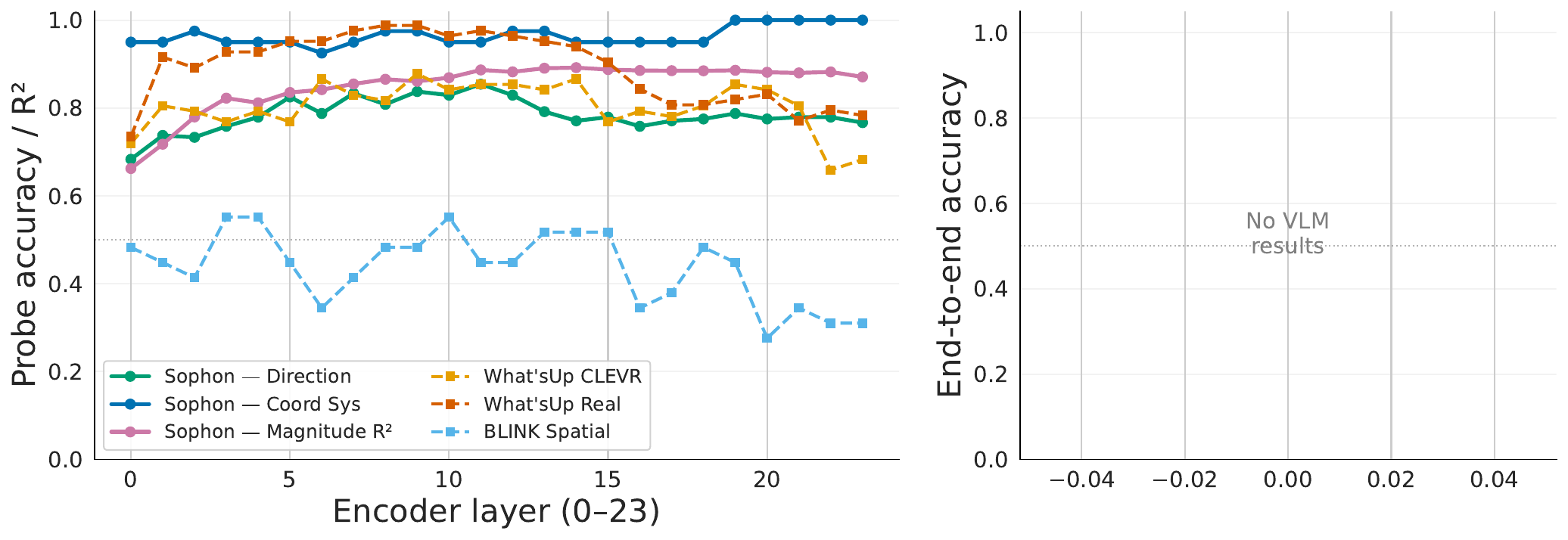}
\caption{SigLIP-large probes. Peaks on Sophon: direction 85.4\% (L11),
coordinate system 100.0\% (L19), magnitude $R^2{=}0.89$ (L14).}
\label{fig:probe-siglip}
\end{figure}

We test DINOv2-large and SigLIP-large as frozen encoder probes
because they cover pretraining objectives outside the
contrastive language image regime. DINOv2 is trained with
self-supervised distillation (no language pairing at all) and
SigLIP uses a sigmoid contrastive objective. If spatial
direction were tied to a specific contrastive vision language
training signal, these encoders would underperform CLIP on Sophon
probes. If it is a generic property of modern image encoders, they
should match. Figures~\ref{fig:probe-dinov2} and~\ref{fig:probe-siglip}
show layer-wise linear probe accuracy on frozen
DINOv2-large~\citep{oquab2024dinov2} and
SigLIP-large~\citep{zhai2023siglip} vision features, on the same
Sophon spec suite and external benchmarks as the CLIP and Qwen-VL
panels in Appendix~\ref{app:probes-architecture}. Both encoders
concentrate spatial information in mid layers (L11--L20) and clear
$\sim$85\% on Sophon direction classification, matching CLIP's
ceiling. The right column reads ``No VLM results'' because DINOv2
and SigLIP are evaluated as frozen encoders without an attached
decoder. They appear as encoder ablations only.

\section{Datasheet for Datasets}
\label{app:datasheet}

We follow the datasheet structure of Gebru et al.~\citep{gebru2021datasheets}.
Sophon is a procedurally generated benchmark, so many questions reduce to
``not applicable'' or ``zero, by construction.''

\subsection{Motivation}

\paragraph{For what purpose was the dataset created?} To isolate
geometric spatial perception in vision language models (VLMs) from
semantic shortcuts (object identity, scene layout priors, language
co-occurrence). Existing spatial benchmarks (What's Up, BLINK Spatial)
contain naturalistic stimuli where these priors confound failure
diagnosis. Sophon strips them away.

\paragraph{Who created the dataset?}
The authors, at Exoptica (Kihei, HI).

\paragraph{Who funded the creation of the dataset?}
The work received no external funding.

\paragraph{Any other comments?} Sophon is intended as a complement to,
not a replacement for, naturalistic spatial benchmarks.

\subsection{Composition}

\paragraph{What do the instances represent?} Each instance is a pair of
overlaid point clouds (blue and red) rendered into a single $512\times512$
PNG with axis gridlines on a plain neutral background. The two clouds
share an underlying point pattern but differ by one geometric
transformation (translation, rotation, scale, or compound). Each
instance is paired with a question text and a ground truth answer.

\paragraph{How many instances are there?} The released suite
(\texttt{sophon/specs/full\_suite.yaml}) has ten spec files, each
drawing 200 stimuli per seed and asking two or three question types
of each stimulus, for 25 (spec, question type) cells and 5{,}000
instances per seed. The three evaluation seeds (42, 43, 44) give
15{,}000 instances, shipped pre-rendered as release assets and
reproducible from the generator. The paper's diagnostic tables report
the cells listed in Appendix~\ref{app:per-cell}; the combined
\texttt{cartesian\_translation} spec (both axes at once) is in the
suite but not in those tables, since its per-axis specs cover it.
Training packages use seed 100 at sizes 500, 1{,}000, 5{,}000, and
10{,}000.

\paragraph{Does the dataset contain all possible instances or is it a
sample?} It is a sample drawn from a parametric generator with
configurable ranges. In every released spec the number of points per
cloud is uniform in $[30, 60]$, points are uniform within the
rendered bounds, and the margin from the axis boundary is $0.1$.
Transform magnitudes are uniform within per-spec ranges: Cartesian
translation $[-0.15, 0.15]$ per axis on $[0,1]^2$, 1D linear
translation $[-0.3, 0.3]$, rotation $[-45^\circ, 45^\circ]$ in both
Cartesian and polar, scale factor $[0.7, 1.3]$ (deviation
$[-0.3, 0.3]$) in both Cartesian and polar radial scale, polar radial
shift $[-0.2, 0.2]$, and the compound spec translation $[-0.1, 0.1]$,
rotation $[-30^\circ, 30^\circ]$, and scale deviation $[-0.2, 0.2]$
together. The exact ranges are in \texttt{sophon/specs/*.yaml}. The
sample is reproducible from the seed.

\paragraph{What data does each instance consist of?} (1) A
$512\times512$ PNG, (2) the question text, (3) the ground truth
answer (a direction word, yes/no, A/B, or a number, by question type),
and (4) a metadata record (\texttt{metadata.jsonl}) with the spec name,
coordinate system, question type, transform parameters and magnitude,
direction word and offered vocabulary for categorical cells, point
count, and a deterministic \texttt{sample\_id}.

\paragraph{Is there a label or target associated with each instance?}
Yes. Every instance has a ground truth answer derivable from the
generator's transform parameters.

\paragraph{Is any information missing from individual instances?} No.

\paragraph{Are relationships between individual instances made explicit?}
Instances within a cell share spec parameters (coord system, transform
type, question type). Cross-cell relationships are encoded in the
spec filename and \texttt{sample\_id}.

\paragraph{Are there recommended data splits?} Evaluation seeds 42,
43, and 44 are held out from training (seed 100). The cells span
three coordinate systems (1D linear, 2D Cartesian, 2D polar) and four
question types (binary, categorical, comparative, quantitative).

\paragraph{Are there any errors, sources of noise, or redundancies?}
The stimuli are deterministic given the seed. No annotation noise
exists because labels are derived from generator parameters.

\paragraph{Is the dataset self contained, or does it link to external
resources?} Self contained.

\paragraph{Does the dataset contain confidential, offensive, threatening,
or sensitive content?} No. All stimuli are abstract colored point
clouds with no identifiable people, places, objects, or text.

\paragraph{Does the dataset relate to people?} No.

\subsection{Collection Process}

\paragraph{How was the data acquired?} Procedurally generated by the
Sophon generator (\texttt{sophon-bench} 0.3.3). No human annotators
or scraped data are involved.

\paragraph{What mechanisms or procedures were used to collect the data?}
The generator deterministically produces (image, question, answer)
tuples from configured cell specs and a random seed. See
\S\ref{sec:benchmark} of the main paper.

\paragraph{If the dataset is a sample, what was the sampling strategy?}
For each cell, transform parameters are drawn uniformly within the
configured range. The point distributions per cloud are uniform within
the rendered axis bounds (with a \texttt{margin} guard).

\paragraph{Who was involved in the data collection process?} The paper's
authors. No external contractors, crowdworkers, or third party data
providers were used.

\paragraph{Over what timeframe was the data collected?} Generated
during paper development (March--April 2026). The procedural generator
allows future regeneration at any time.

\paragraph{Were any ethical review processes conducted?} Not applicable
— the dataset contains no human subjects and no personal data.

\paragraph{Does the dataset relate to people?} No.

\subsection{Preprocessing, Cleaning, and Labeling}

\paragraph{Was any preprocessing or cleaning performed?} The generator
applies standard rendering guards (point margin from axis edges, point
count constraints, transform-magnitude bounds per cell). No post-hoc
filtering is applied to the generated data.

\paragraph{Was the ``raw'' data saved in addition to the preprocessed
data?} The generator parameters and random seeds are the canonical
``raw'' representation. The rendered PNGs are reproducible from them.

\paragraph{Is the software used to preprocess/clean/label the data
available?} Yes. The generator is the benchmark package (MIT).

\subsection{Uses}

\paragraph{Has the dataset been used for any tasks already?} Yes, for
all experiments reported in this paper. These span frontier and open VLM
evaluation, layer-wise vision encoder probing, fine tuning ablations,
cross-coordinate transfer studies, and chain of thought analyses.

\paragraph{Is there a repository linking to papers using the dataset?}
A leaderboard and citation index will be maintained in the GitHub
repository.

\paragraph{What other tasks could the dataset be used for?} Projector
and connector ablations, curriculum learning studies, spatial token
augmentation experiments, image side coordinate embedding studies,
and as a unit test suite for new VLM architectures.

\paragraph{Is there anything about the composition of the dataset or
the way it was collected that might impact future uses?} Sophon's
stimuli are deliberately abstract. Conclusions drawn from Sophon apply
to the controlled abstract regime. Extending to naturalistic stimuli
requires a different stimulus pipeline.

\paragraph{Are there tasks for which the dataset should not be used?}
Sophon should not be used as a standalone proxy for ``general spatial
reasoning ability,'' which would require complementary naturalistic
benchmarks.

\subsection{Distribution}

\paragraph{Will the dataset be distributed to third parties outside the
entity creating it?} Yes, publicly.

\paragraph{How will the dataset be distributed?} (a) The procedural
generator and evaluation harness are public at
\url{https://github.com/zgazak/sophon} and install from PyPI with
\texttt{pip install sophon-bench}. (b) The pre-rendered evaluation
suite (seeds 42, 43, 44; 5{,}000 instances per seed) is attached to
the v0.3.3 release of that repository as one archive per seed, and
the generator reproduces it exactly from the spec files and seeds.

\paragraph{When will the dataset be distributed?} It is public now.

\paragraph{Will the dataset be distributed under a copyright or
intellectual property license?} The generator code is under the MIT
License and the pre-rendered data under CC-BY 4.0.

\paragraph{Have any third parties imposed IP-based or other
restrictions?} No.

\paragraph{Do any export controls or other regulatory restrictions
apply?} No.

\subsection{Maintenance}

\paragraph{Who will be supporting/hosting/maintaining the dataset?}
The authors, through the public repository at
\url{https://github.com/zgazak/sophon}.

\paragraph{How can the owner/curator/manager be contacted?}
By issue on the GitHub repository, or by email to the corresponding
author (\texttt{jzg@alum.mit.edu}).

\paragraph{Is there an erratum?} Issues will be tracked publicly in
the GitHub repository.

\paragraph{Will the dataset be updated?} Yes. We commit to a two year
maintenance window with quarterly frontier model re-evaluation against
held out seeds, to track whether new VLMs improve on Sophon over time.

\paragraph{If the dataset relates to people, are there applicable
limits on the retention of the data?} Not applicable.

\paragraph{Will older versions of the dataset continue to be
supported/hosted/maintained?} Yes. Versions are tagged in git and
prior versions remain installable.

\paragraph{If others want to extend/augment/build on/contribute to
the dataset, is there a mechanism for them to do so?} Yes, pull
requests welcomed in the public GitHub repository. Forked variants
should preserve the seed semantics for reproducibility.

\section{Release, Licensing, and Maintenance}
\label{app:release}

We release three artifacts:
\begin{itemize}
  \item the procedural \textbf{generator and evaluation harness} as a
        Python package (MIT licensed, public at
        \url{https://github.com/zgazak/sophon}, installable with
        \texttt{pip install sophon-bench}); the results in this paper
        were produced with \texttt{sophon-bench} 0.3.3;
  \item a \textbf{pre rendered evaluation suite} (held out seeds 42,
        43, 44 of the diagnostic and held out test specs) under
        CC-BY 4.0, so future users can score new VLMs without
        re rendering, with its download location listed in the
        repository README;
  \item per-paper \textbf{result JSONs and prediction JSONLs} in the
        accompanying repository, so the per-cell tables in
        Appendix~\ref{app:per-cell} and the figures in
        \S\S\ref{sec:results}--\ref{sec:analysis} can be regenerated
        end to end.
\end{itemize}

The generator and evaluation harness are public at
\url{https://github.com/zgazak/sophon}. Because every stimulus is a
deterministic function of its spec file and seed, the packaged
generator reproduces the evaluation suite byte for byte; the
pre-rendered copy is a convenience, and its location and Croissant
metadata file are maintained in the repository README.

\paragraph{Maintenance commitment.}
We commit to a \textbf{two year maintenance window} from publication
date with three concrete actions. We re-evaluate the leading public
frontier models on the held out seeds quarterly and post results to a
public leaderboard. We triage generator bugs and parser pathologies
through GitHub tracked issues. We tag generator versions so that any
cited result remains exactly reproducible after updates. Older versions of the generator and
evaluation harness will remain installable.

\paragraph{Contamination policy.}
Training data for fine tuning experiments uses generator seed 100.
Evaluation seeds (42, 43, 44) are deliberately disjoint to prevent
leakage. We discourage future work from training on the held out seeds,
and the leaderboard will require a contamination statement from each
submission. The procedural generator allows arbitrary fresh seeds for
new evaluation snapshots if a particular seed becomes contaminated by
inclusion in pretraining corpora.

\section{Component Bands and Condition Names}
\label{app:bands}

Each tower splits at a fixed cut into a \emph{workspace} (the early band) and a \emph{reader} (the remaining band). The cut sits after layer 22 in the 28-layer decoders and, proportionally, after layer 26 in the 32-layer decoders. The reader band is the final five decoder layers, the region where the fine-tuned logit-lens contrast separates from baseline. The ablations in \S\ref{sec:mechanism} validate the cut
functionally. The projector is never band-split, and the unembedding and token embeddings are frozen in every condition. All fine-tuning uses LoRA at rank 32, alpha 64, dropout 0.05.

\begin{table}[h]
\centering
\small
\begin{tabular}{l l c c}
\toprule
architecture & tower (depth) & workspace & reader \\
\midrule
\multicolumn{4}{l}{\emph{vision tower}} \\
LLaVA-OneVision & SigLIP-so400m (26) & L0--22 & L23--25 \\
Qwen2.5-VL & windowed ViT (32) & L0--22$^{c}$ & L23--31$^{c}$ \\
LLaVA-NeXT & CLIP-ViT-L (24) & L0--23 (all)$^{b}$ & ---$^{b}$ \\
Idefics3 & SigLIP2 (27) & L0--26 (all)$^{b}$ & ---$^{b}$ \\
\midrule
\multicolumn{4}{l}{\emph{language tower}} \\
LLaVA-OneVision & Qwen2 (28) & L0--22 & L23--27 \\
Qwen2.5-VL & Qwen2.5 (28) & L0--22 & L23--27 \\
LLaVA-NeXT & Mistral (32) & L0--26$^{a}$ & L27--31 \\
Idefics3 & Llama-3.1 (32) & L0--26$^{a}$ & L27--31 \\
\bottomrule
\end{tabular}
\caption{Workspace/reader bands per architecture. Layer indices are inclusive.
Projectors (LLaVA-OneVision and LLaVA-NeXT: 2-layer MLP; Qwen2.5-VL: merger
MLP; Idefics3: single linear) are never band-split.
$^{a}$The 32-layer decoders cut after L26, the same proportional depth as
L22 of 28.
$^{b}$The LLaVA-NeXT and Idefics3 vision towers lie entirely below the cut,
so they have no reader band. For these architectures \textsc{Readers} is
identical to \textsc{Reader}, and their vision towers are fully contained in
the \textsc{Workspaces} condition.
$^{c}$In Qwen2.5-VL's vision tower the banded conditions
(\textsc{Workspaces}, \textsc{Readers}) cover the MLP sublayers only. The
band configs target the language-model module names, which match the vision
SwiGLU MLP but not its fused attention (\texttt{qkv}/\texttt{proj}). The two
conditions use identical targets with complementary bands, so they remain
exact complements. The single-component \textsc{Vision} condition covers
attention and MLP.}
\label{tab:bands}
\end{table}

\begin{table}[h]
\centering
\small
\begin{tabular}{l l}
\toprule
condition & trainable components \\
\midrule
\textsc{Vision} & entire vision tower \\
\textsc{Projector} & projector \\
\textsc{Language} & entire language tower \\
\textsc{Workspace} & language workspace \\
\textsc{Reader} & language reader \\
\textsc{Workspaces} & both towers' workspaces \\
\textsc{Readers} & both towers' readers$^{b}$ \\
\textsc{Full} & both towers$^{d}$ \\
\bottomrule
\end{tabular}
\caption{Condition names. Conditions are named by what they train, and all other
parameters are frozen.
$^{d}$\textsc{Full} trains both towers in their entirety. The projector's
linear layers match none of the LoRA target names, so the projector is
trained only in the \textsc{Projector} condition.}
\label{tab:conditions}
\end{table}

\section{Diagnostic Heatmap: Per-Cell Detail}
\label{app:diagnostic-heatmap}

\begin{figure}[!htbp]
\centering
\includegraphics[width=\linewidth]{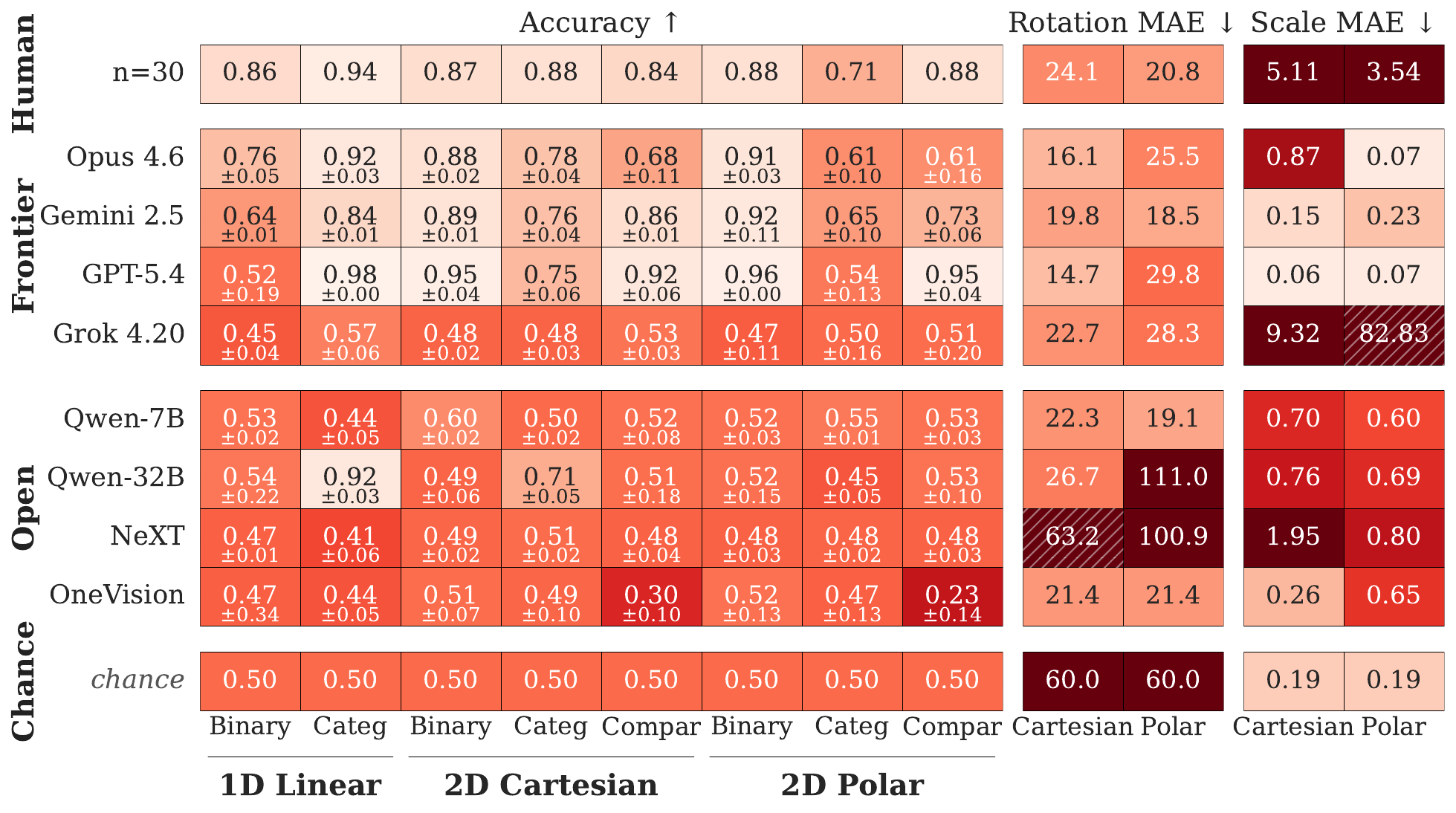}
\caption{Sophon diagnostic heatmap, grouped by tier. Rows are
organized into four bands: Human (Prolific $n{=}30$), Frontier
closed weight (Claude Opus~4.6, Gemini 2.5~Pro, GPT-5.4, Grok 4.20),
Open 7B base (Qwen2.5-VL-7B, Qwen2.5-VL-32B-AWQ, LLaVA-NeXT-7B,
LLaVA-OneVision-7B), and Chance (per-cell random baseline). Columns
are coordinate system $\times$ question type accuracy cells (1D
Linear has no comparative) plus per-coord system rotation MAE and
scale MAE columns. Each cell shows mean accuracy or MAE. Quantitative
cells are diagonally hatched when parse rate is below 25\%
(numerically unreliable). Frontier accuracy cells include the 3-seed
cross seed standard deviation $\pm s$ from the variance subset.
Frontier models match or exceed humans on most binary cells but trail
on categorical and comparative. Open 7B models cluster near per-cell
chance. Grok 4.20 collapses to a near constant token output under
greedy decode (Appendix~\ref{app:grok-degen}).}
\label{fig:diagnostic-heatmap}
\end{figure}

Figure~\ref{fig:diagnostic-heatmap} renders the full diagnostic suite
as a per-model, per-cell heatmap. The visualization makes the
three tier solvability gradient discussed in \S\ref{sec:exp1}
directly inspectable. Frontier closed weight rows match or exceed
the Human row on most binary cells but trail on categorical and
comparative. Open 7B rows cluster near per-cell chance. The Grok
4.20 row collapses across most cells via the constant token failure
mode (Appendix~\ref{app:grok-degen}). Sophon vs external aggregates
are in Table~\ref{tab:main}, and per-cell numbers per model are in
Tables~\ref{tab:per-cell-frontier} and~\ref{tab:per-cell-open}.

\section{Per-Model Summary: Sophon vs.\ External Benchmarks}
\label{app:per-model-summary}

Table~\ref{tab:main} summarizes Sophon mean accuracy alongside scale MAE
ranges and the two external benchmarks (What's~Up~CLEVR, BLINK Spatial)
that we use as the cross benchmark comparators. Empty entries
mark cells where the corresponding external benchmark was not run on
that model. The GPT-4o What's~Up and BLINK numbers anchor the
frontier vs open VLM comparison referenced in \S\ref{sec:exp1}.

\begin{table}[t]
\centering
\small
\caption{Per-model Sophon accuracy with external benchmark
context. Sophon~Acc is the unweighted mean across the 23 underlying
(coordinate system, question type) cells of the diagnostic suite
(Fig.~\ref{fig:diagnostic-heatmap} aggregates these cells into 12
visible columns, 8 accuracy cells plus 4 MAE columns), with $200$ samples
per cell.
$^\ddagger$Grok-4.20 emits a single constant token for $17$ of the
$23$ cells under our default greedy decode (per-cell breakdown in
Appendix~\ref{app:grok-degen}, temperature ablation in
Appendix~\ref{para:temp-ablation}). Its $49.8\%$ mean is the chance
hit rate of those constants on balanced GT distributions.}
\label{tab:main}
\begin{tabular}{@{}lccc@{}}
\toprule
Model & Sophon Acc & Scale MAE & BLINK Spatial \\
\midrule
\textbf{Gemini 2.5 Pro}   & \textbf{82.6\%}          & 0.15--0.23             & 88.8\% \\
\textbf{GPT-5.4}          & \textbf{81.4\%}          & \textbf{0.06--0.07}    & 86.7\% \\
Claude Opus 4.6           & 77.2\%                   & \textbf{0.06--0.09}    & 86.0\% \\
GPT-4o                    & 74.8\%                   & 0.75--0.96             & 81.1\% \\
Grok 4.20                 & 49.8\%$^\ddagger$        & 9.32--82.83$^\ddagger$ & 82.5\% \\
\midrule
Qwen2.5-VL-7B             & 51.1\%                   & 0.60--0.70             & 88.8\% \\
Qwen2.5-VL-32B-AWQ        & 54.7\%                   & 0.69--0.76             & 81.1\% \\
LLaVA-OneVision-7B        & 52.1\%                   & 0.26--0.65             & 83.2\% \\
LLaVA-NeXT-7B             & 47.7\%                   & 0.84--1.00             & 68.5\% \\
\bottomrule
\end{tabular}
\end{table}

\section{Grok-4.20 Constant Token Degeneracy: Per-Cell Detail}
\label{app:grok-degen}

The same constant token failure mode that affects open 7B models
appears for Grok-4.20. A single fixed answer covers
$\geq$$85\%$ of samples in 17 of 23 cells of the diagnostic heatmap. The
fixed answers are spec specific (binary cells default to ``yes'',
categorical cells to a transform appropriate direction word, rotation
quantitative cells to a round angle). Grok presumably scores well on
existing naturalistic spatial benchmarks, but on stimuli that strip
away naturalistic priors it exhibits the same behavioral collapse as
a 7B open VLM. This is the diagnostic property we built Sophon to surface.

\begin{table}[t]
\centering
\small
\caption{Grok-4.20 constant token degeneracy on Sophon. ``Coverage''
is the fraction of samples in that (spec, question type) cell that
received the listed top answer. Cells with $\geq$$85\%$ coverage are
shown ($n{=}17$ of 23 cells); cells below $85\%$ omitted.}
\label{tab:grok-degen}
\begin{tabular}{@{}llcc@{}}
\toprule
Cell (spec / qtype) & Constant answer & Coverage \\
\midrule
all binary specs (×8)             & ``yes''               & 97--99\% \\
Cartesian rotation / categorical  & ``counterclockwise''  & 100\% \\
polar rotation / categorical      & ``counterclockwise''  & 100\% \\
Cartesian rotation / quantitative & ``45'' (degrees)      & 88\% \\
polar rotation / quantitative     & ``45'' (degrees)      & 97\% \\
Cartesian translation x / categ.  & ``left''              & 99\% \\
linear translation / categorical  & ``left''              & 100\% \\
Cartesian translation y / categ.  & ``up''                & 96\% \\
polar radial / categorical        & ``outward''           & 97\% \\
\bottomrule
\end{tabular}
\end{table}

\begin{figure}[!htbp]
\centering
\includegraphics[width=\linewidth]{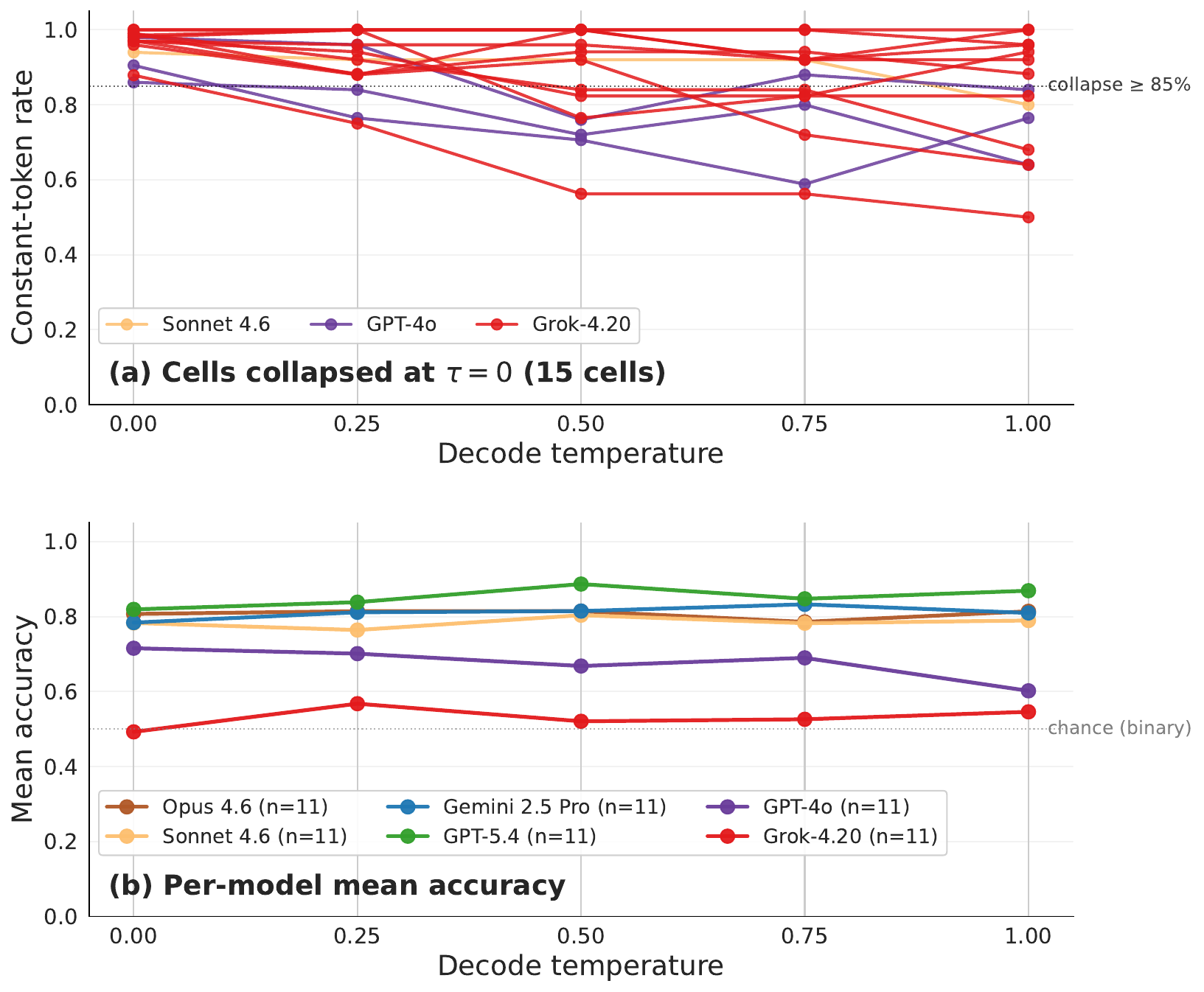}
\caption{\textbf{Decoding temperature partially releases Grok-4.20's
constant token attractor on categorical cells but not on binary cells,
and does not improve the per-model mean accuracy of any frontier VLM.}
\emph{Top panel:} constant token rate (fraction of samples assigned the
most common parsed answer) versus decode temperature for the $15$
(model, cell) combinations that collapse at temp$=$0 (const $\geq 85\%$).
Lines staying above the $85\%$ threshold indicate a persistent
attractor, and lines dropping below indicate a decode induced artifact.
All three GPT-4o (purple) cells release, as does the single Sonnet 4.6
(orange) cell. Of Grok-4.20's $11$ collapsed cells, $4$ release (the
categorical cells on Cartesian trans x, Cartesian trans y, Cartesian rotation, plus
the rotation quantitative cell) while the $7$ binary and comparative
cells stay $\geq 88\%$ at every temperature.
\emph{Bottom panel:} per-model mean accuracy across all $11$ scoreable
cells in the $5$-spec subset, plotted against decode temperature.
Frontier models that succeed at greedy decode (Opus~4.6, Sonnet~4.6, Gemini~2.5~Pro,
GPT-5.4) remain approximately flat ($\Delta \leq +5$\,pp from
$\tau{=}0$ to $\tau{=}1.0$). GPT-4o degrades $\sim 11$\,pp. Grok-4.20
stays at chance across the entire sweep. The full per-cell breakdown backing both
panels is reported in Table~\ref{tab:temp-ablation-summary}.}
\label{fig:temp-ablation-curves}
\end{figure}

\paragraph{Temperature ablation.}
\label{para:temp-ablation}
The Sophon evaluation pipeline uses temperature $0$ (greedy decode) by
default for reproducibility, the standard setting for VLM benchmarks.
Under that setting Grok-4.20 produces the constant token output
documented above. To test whether the collapse is a greedy decode
artifact or a representational failure we re-ran a representative
$5$-spec subset (\texttt{Cartesian\_translation\_x},
\texttt{Cartesian\_translation\_y}, \texttt{Cartesian\_rotation},
\texttt{linear\_translation}, \texttt{polar\_radial}) of the diagnostic
suite at four additional temperatures
$\tau \in \{0.25, 0.5, 0.75, 1.0\}$ across the five frontier models
(Opus~4.6, Gemini~2.5~Pro, GPT-5.4, GPT-4o, Grok-4.20) at $n{=}50$ samples
per cell. Open base controls (Qwen2.5-VL-7B, LLaVA-NeXT-7B) at the same
temperatures are reported alongside in
Appendix~\ref{app:temp-ablation-open} when available. Only the decode
temperature varies. The prompt, vocabulary, image rendering, parser, and
seed are identical to the temp$=$0 evaluation reported in the main
diagnostic heatmap. Figure~\ref{fig:temp-ablation-curves} plots the
resulting curves and Table~\ref{tab:temp-ablation-summary} reports the
qualitative summary.

\begin{table}[t]
\centering
\small
\caption{Temperature ablation summary across the $5$-spec subset for
six frontier VLMs. ``Total cells'' counts the (coordinate system,
question type) cells with a defined accuracy ($11$ per model after
excluding quantitative cells, where constant rate is not directly
interpretable). ``Collapsed @ $\tau{=}0$'' = constant rate $\geq 85\%$ on
the dominant parsed answer. ``Released @ $\tau{=}1.0$'' = constant rate
drops below $85\%$ at temp$=$1.0. Grok-4.20 collapses on every cell at
greedy decode; $4$ release at temp$=$1.0 (predominantly categorical
cells), while binary cells stay above $85\%$ at every temperature.
GPT-4o is the only frontier model whose mean accuracy degrades
substantially with temperature ($-11$\,pp). Mean $\Delta$acc is over
all $11$ cells.}
\label{tab:temp-ablation-summary}
\begin{tabular}{@{}lcccc@{}}
\toprule
Model              & Total cells & Collapsed @ $\tau{=}0$ & Released @ $\tau{=}1.0$ & Mean $\Delta$acc \\
                   &             &                        &                          & ($\tau{=}0\!\to\!1.0$)\\
\midrule
Opus 4.6           & 11 & 0  & ---     & $+1$\,pp \\
Sonnet 4.6         & 11 & 1  & 1 of 1  & $+1$\,pp \\
Gemini 2.5 Pro     & 11 & 0  & ---     & $+3$\,pp \\
GPT-5.4            & 11 & 0  & ---     & $+5$\,pp \\
GPT-4o             & 11 & 3  & 3 of 3  & $-11$\,pp \\
\textbf{Grok-4.20} & \textbf{11} & \textbf{11} & \textbf{4 of 11} & \textbf{$+5$\,pp} \\
\bottomrule
\end{tabular}
\end{table}

\paragraph{Interpretation.}
Three findings emerge. (i) For frontier models that succeed at greedy
decode (Opus~4.6, Sonnet~4.6, Gemini~2.5~Pro, GPT-5.4), mean accuracy is approximately
flat across the temperature sweep ($+1$ to $+5$\,pp from $\tau{=}0$ to
$\tau{=}1.0$). Greedy decoding is the appropriate primary setting for
these models on Sophon, not a setting that disadvantages them. The
clearest counter pattern is GPT-4o, whose mean accuracy drops
$11$\,pp from greedy to $\tau{=}1.0$, with individual cells losing as
much as $29$\,pp (Cartesian trans y binary, $92$ to $64\%$). GPT-4o is the
only frontier model in our set that visibly degrades under stochastic
decoding. (ii) Grok-4.20's constant token attractor is \emph{partially}
released under stochastic decoding but \emph{persistent in aggregate}.
$4$ of $11$ collapsed cells drop below the $85\%$ threshold at
$\tau{=}1.0$ (predominantly categorical cells, where constant rate falls
$\sim 30$\,pp on Cartesian trans x, Cartesian trans y, and Cartesian rotation), while
all binary cells stay $\geq 96\%$ collapsed across the sweep. The
categorical cells where the attractor releases also recover modest
accuracy ($+16$ to $+19$\,pp on Cartesian trans x and Cartesian trans y
categorical), consistent with the single spec spot check that motivated
this ablation. The binary cells, where Grok answers ``yes'' on
essentially every sample regardless of stimulus, do not respond to
temperature at all and remain at $\sim 47\%$ (below the $50\%$ chance
floor due to the asymmetric ``yes''-bias). (iii) Several models gain
$15$--$19$\,pp on \emph{categorical cells} under temperature: Grok-4.20 on
Cartesian trans x and Cartesian trans y categorical, GPT-5.4 on Cartesian trans x
categorical ($+18$\,pp), Gemini~2.5~Pro on Cartesian trans x ($+17$\,pp) and
Cartesian rotation ($+16$\,pp) categorical, and Qwen-7B on Cartesian rotation
categorical ($+16$\,pp). Categorical cells are restricted vocabulary
($2$ candidate direction words). Greedy can lock into the wrong
candidate, while sampling lets the second choice token through. This
effect is absent on binary cells (which use ``yes/no'' and have a
strong language prior toward ``yes''), explaining the categorical vs 
binary asymmetry in Grok's release pattern.

The substantive consequence for the paper's three tier framing is that
Grok-4.20 sits between the frontier reasoner regime and the open base
degeneracy regime on Sophon. Its categorical cell collapse is partially
decode induced and partially perceptual, while its binary cell collapse
is fully perceptual (a fixed ``yes'' that no decode setting recovers).
The categorical recovery pattern observed on several frontier models
suggests that Sophon's per-cell accuracies could be marginally improved
($\sim 5$--$10$\,pp on the affected cells) by averaging across multiple
sampled outputs rather than by greedy decoding. We retain greedy as the
primary protocol for reproducibility, noting that this choice slightly
penalises models with vocabulary token attractor effects.

\paragraph{Reproduction.}
The temperature sweep used the same image prompt parser pipeline as the
main diagnostic heatmap, with only \texttt{temperature} varied per
process. Predictions are stored under
\texttt{results/exp1\_temp\_ablation/temp\_$\tau$/} alongside the
greedy decode predictions in \texttt{results/exp1/}. The per-cell curve
data backing Figure~\ref{fig:temp-ablation-curves} is saved as
\texttt{results/exp1\_temp\_ablation/full\_curves.json}, and the figure
itself is regenerated from those files via
\texttt{scripts/make\_temp\_ablation\_figure.py}.

\section{Per-Architecture Probe Profiles}
\label{app:probes-architecture}

\begin{figure}[!htbp]
\centering
\includegraphics[width=0.96\linewidth]{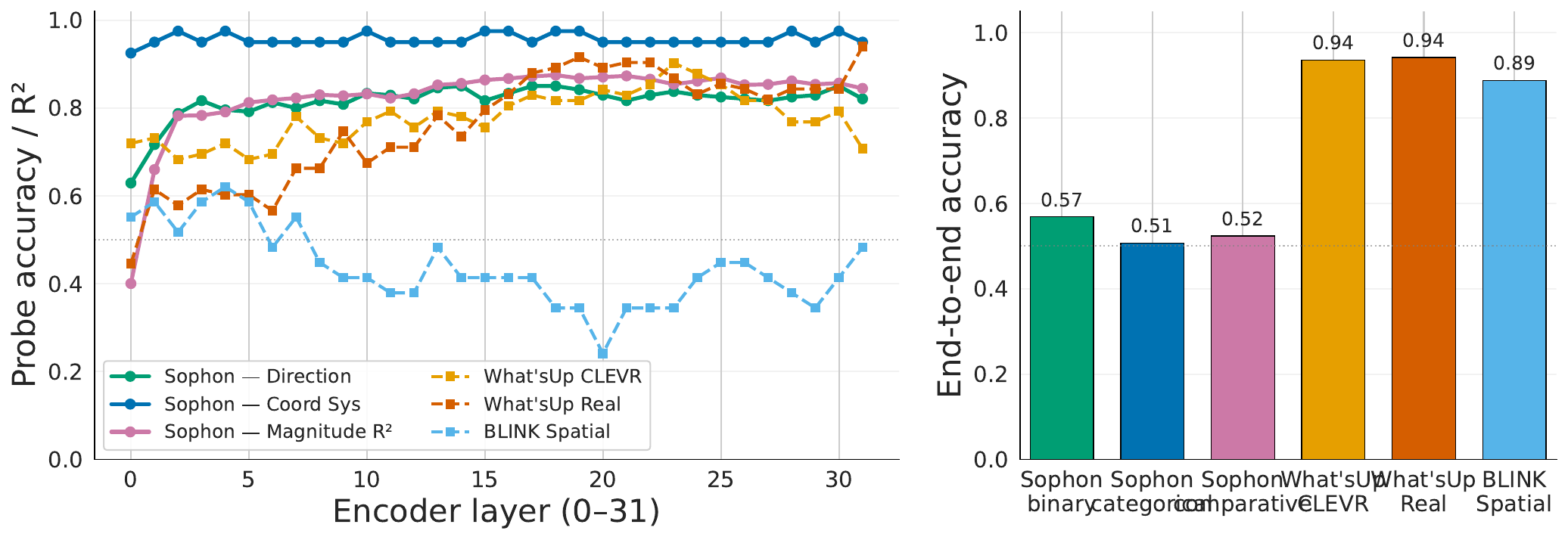}
\caption{Linear probe accuracy on frozen Qwen2.5-VL ViT features across
four benchmarks (Sophon + What's~Up + BLINK). \emph{Left of each panel}:
probe accuracy at every encoder layer, with end-to-end VLM accuracy as
the dotted reference. \emph{Right}: end-to-end VLM accuracy bar.}
\label{fig:probe-qwen}
\end{figure}

Figure~\ref{fig:probe-qwen} shows layer wise probe accuracy on the
frozen Qwen2.5-VL ViT across Sophon plus the two external benchmarks
(What's~Up and BLINK), the per-architecture detail referenced from
\S\ref{sec:exp3}. Sophon spatial content is recoverable at
85--97\% at the top of the encoder, well above the VLM's $\sim$50\%
end-to-end ceiling. The encoder has the information that the readout
does not surface. On BLINK, the probe is below end-to-end VLM accuracy
because BLINK questions can be solved by combining visual features with
object/scene priors that the probe alone cannot exploit. Probes on the
fine tuned encoder are essentially unchanged from the base.

\begin{figure}[!htbp]
\centering
\includegraphics[width=0.96\linewidth]{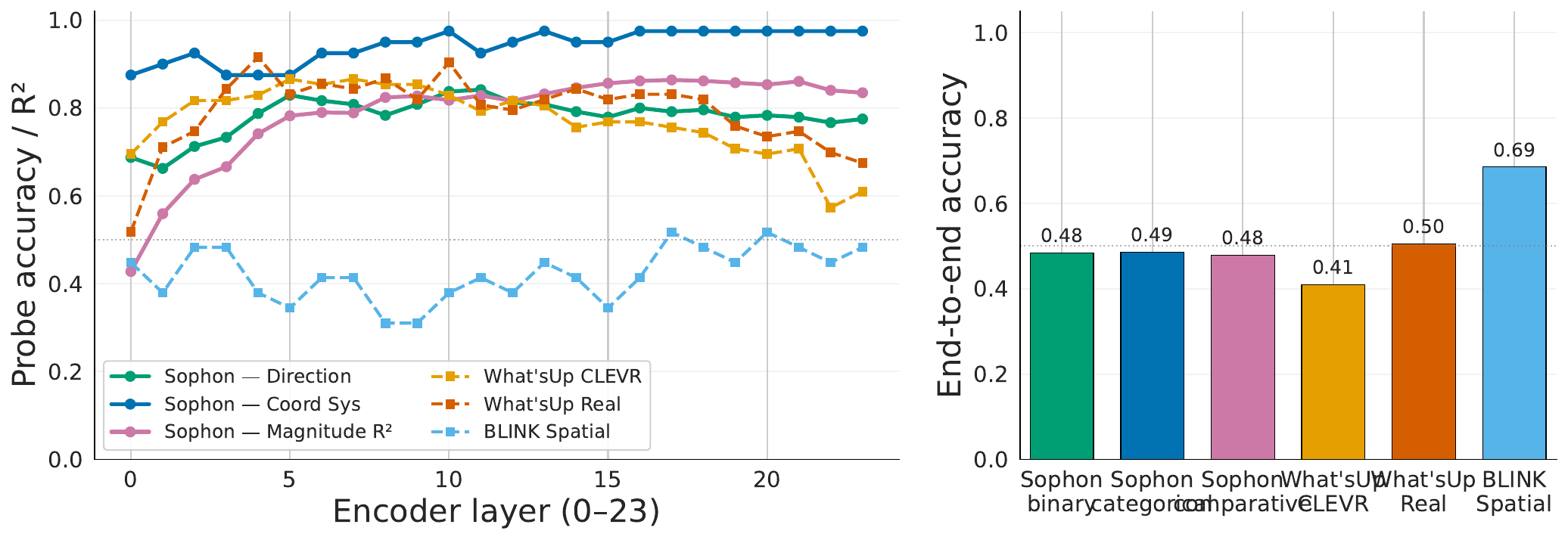}
\caption{CLIP encoder probes (LLaVA-NeXT). Note that CLIP's spatial
content peaks early (L5--7) and degrades toward the final layer.}
\label{fig:probe-clip}
\end{figure}

\begin{figure}[!htbp]
\centering
\includegraphics[width=0.96\linewidth]{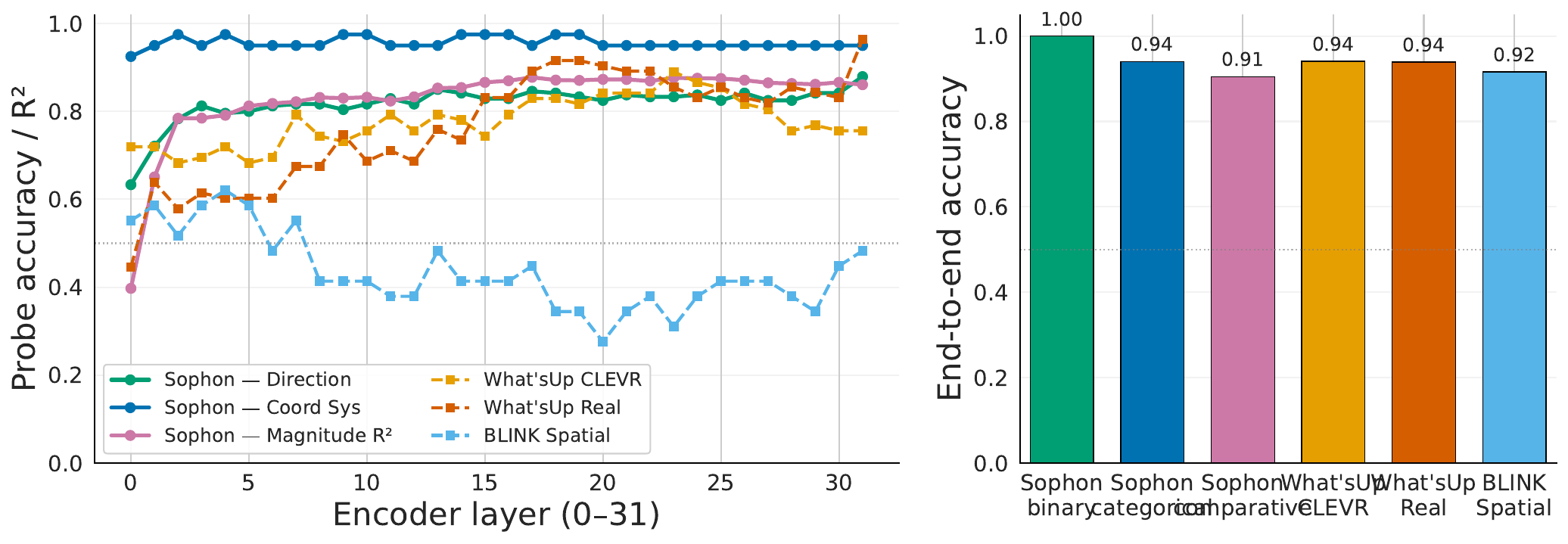}
\caption{Qwen-VL ViT probes after 10k Sophon fine tuning
(LoRA 10k). The shift in Sophon direction peak from L14 to L31
relative to the base (Fig.~\ref{fig:probe-qwen}) is small relative to
the decoder side end-to-end gain.}
\label{fig:probe-qwen-ft10k}
\end{figure}

Figure~\ref{fig:probe-clip} shows the CLIP encoder profile from
LLaVA-NeXT for comparison, and Figure~\ref{fig:probe-qwen-ft10k} shows the
Qwen-VL ViT after 10k Sophon fine tuning. The 10k-trained encoder shifts
Sophon direction peak accuracy from L14 to L31 ($+2.9$\,pp) and gains
$+5.8$\,pp at the final layer, while remaining $\sim$1\,pp on the
external benchmarks, substantially less than the simultaneous decoder
gain ($\sim$35\,pp end-to-end).

\section{Encoder Probe vs.\ End-to-End Accuracy}
\label{app:probe-vs-e2e}

Table~\ref{tab:probes} reports the headline contrast. Linear probes on
frozen CLIP features achieve $85{-}97\%$ accuracy on Sophon spatial
tasks, while LLaVA-NeXT end-to-end on the same images achieves
$41{-}48\%$. The spatial content the encoder represents does not reach
the answer token.

\paragraph{Text only control: prompt and coordinate format.}
\label{app:text-only-protocol}
The text only control replaces the rendered image with a textual
description of the same point coordinates. The model receives two
text only messages, first the coordinate dump and then the question.
The coordinate dump uses the format:
\begin{quote}\small
\texttt{Blue points (reference set):} \\
\texttt{Points in a 2D Cartesian coordinate system with} \\
\texttt{x $\in$ [0, 1] and y $\in$ [0, 1]:} \\
\texttt{~~Point 1: (0.123, 0.456)} \\
\texttt{~~Point 2: (0.234, 0.567)} \\
\texttt{~~...} \\
\texttt{} \\
\texttt{Red points (transformed set):} \\
\texttt{[same format as above]} \\
\texttt{} \\
\texttt{The coordinate system is cartesian\_2d.}
\end{quote}
Coordinates are rendered to three decimal places, and points are sorted
by their first coordinate (x for cartesian, $r$ for polar). The
question text is then appended verbatim from the same template the
image mode evaluation uses. Polar and 1D linear stimuli use the
analogous coordinate system specific format
(\texttt{sophon.coordinates.\{cartesian,polar,linear\_1d\}.describe\_points}).
The same model is queried with the same vocabulary restrictions and
parser. Only the input modality changes.

\begin{table}[t]
\centering
\small
\caption{Encoder probe vs.\ end-to-end accuracy on Sophon tasks. The
end-to-end magnitude $R^2$ is computed over the responses that parse
to a number. LLaVA-NeXT emits out of scale values on a substantial
fraction of magnitude items (e.g., predicting $180^\circ$ or
$360^\circ$ when the ground truth rotation is under $45^\circ$),
yielding $R^2 \ll 0$. The probe $R^2$ is from sklearn ridge regression
over the same training/test split. Per cell parse rates and
prediction range diagnostics are in
Appendix~\ref{app:per-cell}.}
\label{tab:probes}
\begin{tabular}{@{}lcc@{}}
\toprule
Task & CLIP Probe & LLaVA-NeXT End-to-End \\
\midrule
Coordinate system identification & 97.5\% & 44\% \\
Direction classification         & 81\%   & 48\% \\
Magnitude regression ($R^2$)     & 0.83   & $-54.8$ \\
\bottomrule
\end{tabular}
\end{table}

\paragraph{Encoder is nearly invariant under LoRA fine tuning.}
Re-running the probe on Qwen LoRA-1k shows $<$$1$\,pp drift at every
layer on every dataset, meaning that the encoder is essentially identical to
base. At LoRA-10k we observe a small late layer shift: Sophon
direction peak accuracy moves from L14 (base) to L31 (10k) and gains
$+2.9$\,pp; What'sUp Real gains $+2.4$\,pp at L31. These are an order
of magnitude smaller than the simultaneous decoder side gains
($+34.5$\,pp end-to-end at 10k), confirming the language decoder is
the dominant site of learning.

\paragraph{The BLINK probe inversion.}
On BLINK, both CLIP and Qwen probes hover near chance ($48{-}62\%$)
while the full VLM scores $69{-}89\%$. The VLM outperforms its own
encoder's linear spatial readout because BLINK questions are
language guided (``is the car beneath the cat?'') and can be solved by
leveraging object identity plus language priors, without isolated
geometric spatial competence. This makes BLINK accuracy a confounded
measure. High scores may reflect semantic or linguistic reasoning
rather than spatial perception.

\section{Chain of Thought Per-Model Breakdown}
\label{app:cot}

The two turn CoT protocol (\S\ref{sec:exp3}) elicits a free form
description of the image in turn 1 and a structured answer in turn 2.
Table~\ref{tab:cot} reports per-model accuracy split by transform
type, alongside two diagnostics. \textbf{Description to GT} is the fraction of
turn 1 descriptions that mention the correct direction word.
\textbf{Contradict.} is the subset of those samples whose turn 2 answer
disagrees with the description's stated direction. The contradiction
rate isolates the readout failure cleanly. The model can name the
correct transformation in description and still emit the wrong final
answer.

\begin{table}[t]
\centering
\small
\caption{Two turn CoT accuracy by transform type.
\textbf{Description to GT}: fraction of categorical samples whose Turn 1
description mentions the correct direction word.
\textbf{Contradict.}: subset of those samples where the Turn 2 final
answer disagrees with the description's stated direction.}
\label{tab:cot}
\begin{tabular}{@{}lcccccc@{}}
\toprule
Model & Overall & Rot. & Trans. & Scale & Description to GT & Contradict. \\
\midrule
LLaVA base                & 53\% & 46\% & 55\% & 58\% & 57\% & 29\%\\
LLaVA LoRA               & 52\% & 48\% & 52\% & 57\% & 41\% & 23\%\\
Qwen base                 & 52\% & 45\% & 51\% & 58\% & 63\% & 27\%\\
Qwen LoRA (1k)           & 57\% & 46\% & 54\% & 70\% & 57\% & 23\%\\
Qwen LoRA (10k)          & 58\% & 43\% & 52\% & 71\% & 67\% & 25\%\\
Qwen-32B-AWQ              & 52\% & 44\% & 56\% & 52\% & 54\% & 41\%\\
OneVision base            & 47\% & 46\% & 52\% & 40\% & 62\% & 25\%\\
OneVision LoRA           & 57\% & 46\% & 56\% & 61\% & 33\% & 11\%\\
\textbf{Opus 4.6}         & \textbf{78\%} & \textbf{54\%} & \textbf{76\%} & \textbf{96\%} & \textbf{84\%} & \textbf{33\%} \\
\bottomrule
\end{tabular}
\end{table}

\section{Fine Tuning: Volume Curve and Per-Architecture Detail}
\label{app:fine-tuning}

This appendix collects the per-volume and per-architecture detail expanding on the compressed §\ref{sec:exp4} narrative.

\paragraph{The projector is not the lever.}
\textsc{Projector}-only training lands at base level on OneVision
($49.1\%$ vs base $46.6\%$) and Idefics3 ($51.9\%$ vs $53.5\%$), and
\emph{below} base on Qwen2.5-VL ($43.0\%$ vs $56.5\%$, with trans\_y
at $30\%$). The alignment interface neither carries the gain nor
tolerates being trained alone on Qwen. Balanced evaluation, $200$ per
direction.

\begin{figure}[!htbp]
\centering
\includegraphics[width=0.98\linewidth]{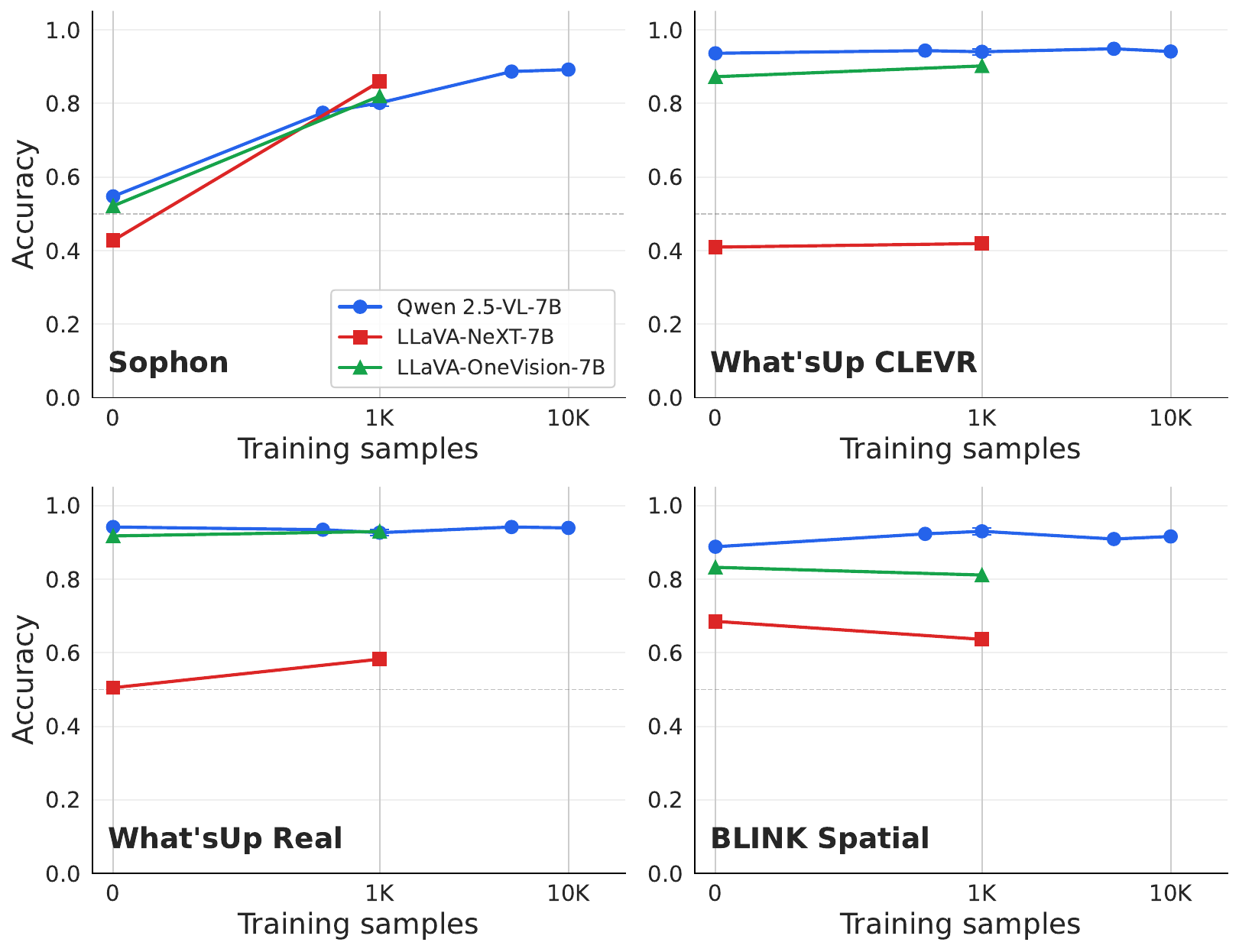}
\caption{Qwen2.5-VL-7B training volume curve, plus 1k points for the
two other architectures. Error bars on the Qwen 1k point are 1$\sigma$
across seeds 42/43/44, and other points are single seed. Sophon saturates
at 5k, What'sUp is stable, and BLINK transfer peaks at 1k and
regresses at higher volumes.}
\label{fig:volume-curve}
\end{figure}

\paragraph{Volume curve (Qwen2.5-VL-7B).}
Figure~\ref{fig:volume-curve} and Table~\ref{tab:volume-app} report
Sophon, What'sUp, and BLINK accuracy at four training volumes (500,
1k, 5k, 10k Sophon samples; full LoRA, full vocabulary random training,
merged adapters). Sophon saturates at 5k ($+0.5$ to $+2$\,pp from 5k
 to 10k), What'sUp is stable across the curve, and BLINK transfer
peaks at 1k ($+4.4 \pm 0.8$\,pp over base across $n{=}3$ seeds) and
regresses at higher volumes.

\begin{table}[t]
\centering
\small
\caption{Qwen2.5-VL-7B volume curve, full LoRA, full vocabulary random
training, merged evals. Sophon saturates at 5k and BLINK transfer peaks
at 1k. Values percent.}
\label{tab:volume-app}
\begin{tabular}{@{}lcccc@{}}
\toprule
Volume & Sophon & WU-CLEVR & WU-Real & BLINK \\
\midrule
Base ($n{=}1$) & 54.7 & 93.6 & 94.2 & 88.8 \\
500 ($n{=}3$)  & 78.7 $\pm$ 1.7 & 94.0 $\pm$ 0.4 & 93.4 $\pm$ 0.2 & 91.1 $\pm$ 1.1 \\
1k ($n{=}3$)   & 80.2 $\pm$ 0.8 & 94.0 $\pm$ 0.8 & 92.6 $\pm$ 0.8 & \textbf{93.2 $\pm$ 0.8} \\
5k ($n{=}1$)   & 88.7 & 94.9 & 94.2 & 90.9 \\
10k ($n{=}1$)  & 89.2 & 94.1 & 93.9 & 91.6 \\
\bottomrule
\end{tabular}
\end{table}

\paragraph{External benchmarks across the retrained grid.}
Table~\ref{tab:external-clean} reports the external benchmarks for
base, \textsc{Workspace}, and \textsc{Full} on all four architectures.
These are the same merged checkpoints as the balanced Sophon grid,
single seed. Sophon training leaves these external spatial benchmarks
intact. We do not evaluate non-spatial abilities, so this bounds
forgetting within the spatial domain only.
On three architectures every move is within sampling noise (BLINK
$n{=}143$, s.e.\ ${\approx}3$\,pp; What'sUp $n{\approx}410$, s.e.\
${\approx}2$\,pp). The What'sUp moves are mostly small gains.
LLaVA-NeXT trades $7$\,pp of BLINK for $+6$ to $+9$\,pp on both
What'sUp splits. No cell collapses.

\begin{table}[t]
\centering
\small
\caption{External benchmarks across the retrained grid. Same merged
checkpoints as the balanced Sophon evaluation, single seed. Values
percent.}
\label{tab:external-clean}
\begin{tabular}{@{}llccc@{}}
\toprule
Architecture & Condition & WU-CLEVR & WU-Real & BLINK \\
\midrule
OneVision   & Base              & 87.3 & 91.7 & 83.2 \\
            & \textsc{Workspace} & 90.2 & 93.4 & 81.1 \\
            & \textsc{Full}      & 90.7 & 93.9 & 81.8 \\
\midrule
Qwen2.5-VL  & Base              & 93.6 & 94.2 & 88.8 \\
            & \textsc{Workspace} & 93.4 & 93.7 & 90.2 \\
            & \textsc{Full}      & 94.1 & 92.7 & 92.3 \\
\midrule
LLaVA-NeXT  & Base              & 40.9 & 50.5 & 68.5 \\
            & \textsc{Workspace} & 46.8 & 59.5 & 60.8 \\
            & \textsc{Full}      & 47.3 & 59.2 & 61.5 \\
\midrule
Idefics3    & Base              & 73.0 & 73.1 & 85.0 \\
            & \textsc{Workspace} & 73.8 & 75.0 & 83.2 \\
            & \textsc{Full}      & 74.0 & 74.8 & 82.5 \\
\bottomrule
\end{tabular}
\end{table}

\paragraph{Cross-coord transfer detail (Qwen).}
Table~\ref{tab:crosscoord} reports Qwen's full per-coord eval matrix.
Cartesian only training brings Cartesian 2D to $93.0 \pm 0.5\%$ but leaves polar
at $51.7 \pm 2.6\%$, within $1$\,pp of base ($50.8\%$). Polar only
training transfers $+22.2 \pm 1.1$\,pp to Cartesian 2D. Every polar only seed
lands between $77.0$ and $78.9\%$ on Cartesian 2D. Every Cartesian only seed lands
between $49.5$ and $54.5\%$ on polar.

\begin{table}[t]
\centering
\small
\caption{Cross-coordinate transfer for Qwen2.5-VL-7B ($n{=}3$ seeds).
Cartesian only training drives Cartesian specs above mixed-coord training but
leaves polar at base. Polar only training drives polar specs up modestly
and \emph{also} transfers substantially to Cartesian. Mean accuracy
across question types per coordinate family.}
\label{tab:crosscoord}
\begin{tabular}{@{}lcccc@{}}
\toprule
Eval coord & Base & LoRA all (1k) & LoRA Cartesian & LoRA polar \\
\midrule
Cartesian 2D       & 55.4\% & 88.0\% & \textbf{93.0 $\pm$ 0.5\%} & 77.6 $\pm$ 1.1\% \\
Linear 1D     & 45.0\% & 79.5\% & 66.2 $\pm$ 1.3\%          & \textbf{72.8 $\pm$ 3.1\%} \\
Polar 2D      & 50.8\% & 62.0\% & \textbf{51.7 $\pm$ 2.6\%} & 67.3 $\pm$ 0.8\% \\
\bottomrule
\end{tabular}
\end{table}

\paragraph{Cartesian to polar transfer per architecture.}
Cartesian only training brings Qwen's Cartesian 2D to $93.0\%$ but leaves polar at
$51.7\%$, identical to base. Polar categorical drops to $26\%$, below
the $16.7\%$ 6-way chance floor. OneVision and LLaVA-NeXT show smaller
but non-zero Cartesian to polar transfer ($+8.5 \pm 1.4$\,pp and
$+13.1 \pm 2.2$\,pp respectively), suggesting their bindings are
looser. The three plausible contributors to the polar to Cartesian asymmetry
(vocabulary overlap, polar's coupled radial$\times$angular visual
variety, and polar primitives re-projecting into Cartesian while the reverse
needs additional context) are discussed alongside the readout pathway framing
(\S\ref{sec:mechanism}). Per-architecture cross-coord heatmaps are
in Appendix~\ref{app:crosscoord-heatmap}.

\paragraph{Question prompts.}
Every Sophon sample is a single turn user message containing one PNG and
one of the following text templates verbatim:

\begin{itemize}
  \item \textbf{Binary:} ``Are the blue and red point sets aligned to within
        approximately one dot width? Answer yes or no.''
  \item \textbf{Categorical (default):} ``Which direction should the red
        points move to align with the blue points? Answer with one of:
        \{directions\}.''
  \item \textbf{Categorical (rotation):} ``Which direction should the red
        points move over the shortest arc to align with the blue points?
        Answer with one of: \{directions\}.''
  \item \textbf{Comparative:} ``Which image shows better alignment between
        the blue and red point sets? Answer A or B.''
  \item \textbf{Quantitative (rotation):} ``Estimate the rotation angle in
        degrees between the blue and red point sets. State your estimate
        in degrees as a number.''
  \item \textbf{Quantitative (scale):} ``By what factor are the red points
        scaled relative to the blue points? State your estimate as a
        numeric factor.''
\end{itemize}

The categorical templates accept a \textbf{restricted} vocabulary by
default (only the direction pair relevant to the active transform, e.g.
``left, right'' for an x translation, or ``clockwise, counterclockwise''
for a rotation), with an optional \textbf{full} vocabulary mode that
exposes the coordinate system's complete six-way direction set
(\S\ref{sec:vocab-ablation} reports an ablation between the two).
Vocabulary order is canonical by default and can be deterministically
shuffled (or placed correct last) via \texttt{vocab\_order} for
positional shortcut analysis (Appendix~\ref{app:vocab-order}).

\paragraph{Response parsing.}
\label{app:parser}
Free form VLM responses are mapped to structured answers by the
\texttt{sophon.parsing} module, with type specific rules:

\begin{itemize}
  \item \textbf{Binary} uses \emph{last keyword wins}, the later of any
        \texttt{\textbackslash byes\textbackslash b} or
        \texttt{\textbackslash b(no$\vert$not)\textbackslash b} match in
        the response. Models often restate the question, so the conclusion
        wins.
  \item \textbf{Categorical} uses last keyword wins over the supplied
        vocabulary (e.g.~``shifted right \ldots move left'' parses to
        \texttt{left}).
  \item \textbf{Comparative} runs three priorities in order: (0) the
        last explicit answer marker (\texttt{(?:final\textbackslash s+)?answer\ldots}
        regex with optional bolding) anywhere in the response; (1) a
        bolded \texttt{**A**} or \texttt{**B**}; (2) the first standalone
        \texttt{A} or \texttt{B}. Priority 0 was added during this paper
        to handle structured response models (e.g.~Qwen2.5-VL-32B) whose
        ``\textbf{Image A:} \ldots \textbf{Image B:} \ldots
        \textbf{Final Answer: B}'' format would otherwise grab the
        ``A'' from the section header. The fix moved 122 of 200
        compound comparative samples from incorrect to correct on 32B
        without changing any other model's accuracy by more than $0.5$
        pp.
  \item \textbf{Quantitative} prefers unit adjacent numbers
        (\emph{30 degrees}, \emph{2.3$\times$}, \emph{factor of 0.85})
        over bare numbers, and rejects list/step markers
        (\emph{1.~Normalize\ldots}) that would otherwise be scraped as
        answers.
\end{itemize}

\paragraph{Failure mode notes.}
Three parser pathologies are worth flagging for re-implementers.
(i)~Models that emit a verbose chain ending with an answer marker need
the comparative Priority 0 to avoid first keyword pickup.
(ii)~Quantitative responses with multiple plausible numbers
(``approximately 30 degrees, more precisely 32.5'') return the last
unit adjacent number. This is the most defensible heuristic but worth
checking on a new model class.
(iii)~Categorical responses with rotation tokens
(``clockwise''/``counterclockwise'') in the body of a translation
discussion will be misclassified if the full vocabulary is offered.
This is the vocabulary distractor finding in \S\ref{sec:vocab-ablation}
and motivates the restricted vocab default.

\section{Cross-Coordinate Transfer: Per Architecture Detail}
\label{app:crosscoord-heatmap}

\begin{figure}[!htbp]
\centering
\includegraphics[width=0.85\linewidth]{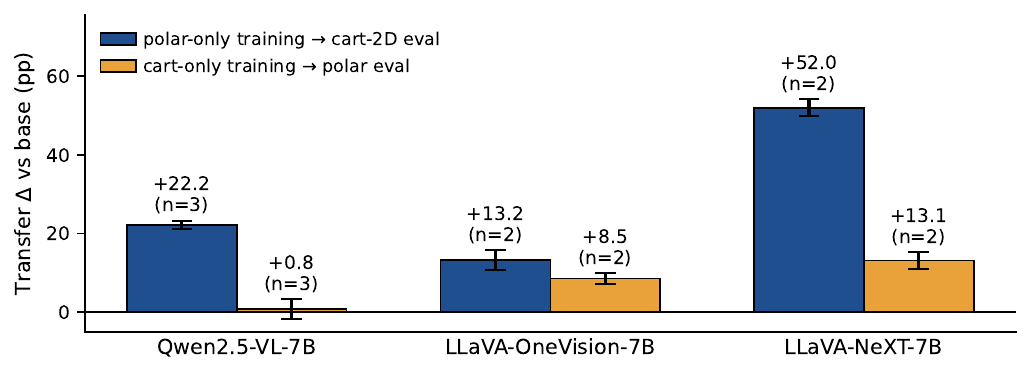}
\caption{Cross-coordinate transfer asymmetry across three open
architectures we test.}
\label{fig:crosscoord}
\end{figure}

Figure~\ref{fig:crosscoord} summarizes the asymmetry as a 6 bar chart.
Polar only training (left bars) transfers $+13$ to $+50$\,pp to
Cartesian 2D evaluation. Cartesian only training (right bars) transfers $+1$ to
$+13$\,pp to polar evaluation. Error bars are $1\sigma$ across seeds
where $n{\geq}2$. Qwen $n{=}3$ train side (seeds 42, 43, 44 full
retrains). LLaVA-NeXT and OneVision $n{=}3$ with $n{=}2$ train side
(seeds 42, 43) plus an additional eval side data point at seed 44 on
the seed 42 adapter, since per condition retrain wall clock for the
non Qwen architectures was prohibitive (Appendix~\ref{app:limitations}).
The off diagonal pattern that matters most for the shortcut claim is
the polar row's positive transfer to the Cartesian 2D column versus
the Cartesian row's near zero transfer to the polar column.

\section{Why Rotation Stays Near Chance for VLMs}
\label{app:rotation-discussion}

Even Opus reaches only $54\%$ on rotation categorical (barely above the
$50\%$ binary chance floor), and 1000~rotation only training samples
cannot move 7B accuracy off chance. We do not have a complete account
of this gap. What we can say is that the gap is a real model side
failure, not an intrinsic property of unordered point clouds. The human
baseline clears polar rotation at $r{=}0.69$
(\S\ref{sec:human-baseline}), the only perceiver class result in our
entire quantitative grid, so the global pattern of points carries
enough rotation signal for a perceiver to recover direction. Plausible
contributors include the relative absence of explicit rotation
supervision in pretraining mixtures and the difficulty of binding
continuous angular variation to discrete CW/CCW direction words.
Sophon's other transforms (translation, scale) do not exhibit the same
gap. Opus reaches $76\% / 96\%$ on translation categorical / scale
binary, which makes rotation a specific readout failure rather
than a general perception limitation. An ordered point cloud Sophon variant would be a useful
control to separate ``cannot perceive rotation'' from ``cannot bind
perceived rotation to language.''

\section{Appearance Battery: Full Results}
\label{app:rendering-battery}

Each variant re-renders the same point geometry (identical
coordinates at fixed stimulus seed) with one appearance change. Models
are evaluated zero-shot with no retraining across three stimulus seeds. We evaluate $50$ stimuli per task per variant per seed and study x and y translation and radial scale. Table~\ref{tab:battery-families} reports suite-family aggregates across conditions. Table~\ref{tab:battery-variants} splits the \textsc{Workspace} cells to variant level.
Table~\ref{tab:battery-npts} reports the point-count sweep per task and
Table~\ref{tab:battery-res} the render-resolution sweep.

\begin{table}[t]
\centering
\small
\caption{Suite-family aggregates, mean over three stimulus seeds $\pm$
seed range. The standard family mixes the standard rendering with
marker and sparsity variants. Table~\ref{tab:battery-variants} splits
it. \textsc{Workspace} rows use each architecture's workspace-band
cell.}
\label{tab:battery-families}
\begin{tabular}{@{}lcccccc@{}}
\toprule
Arch / condition & standard & colors & dot count & role rev. & grid off & black bg \\
\midrule
OneVision base & 52$\pm$1 & 50$\pm$0 & 50$\pm$0 & 50$\pm$0 & 50$\pm$0 & 50$\pm$0 \\
OneVision \textsc{Reader} & 54$\pm$2 & 54$\pm$3 & 53$\pm$1 & 51$\pm$3 & 54$\pm$1 & 53$\pm$5 \\
OneVision \textsc{Workspace} & 84$\pm$1 & 74$\pm$3 & 87$\pm$1 & 82$\pm$2 & 93$\pm$1 & 92$\pm$2 \\
OneVision \textsc{Language} & 87$\pm$2 & 77$\pm$4 & 87$\pm$1 & 84$\pm$2 & 91$\pm$3 & 89$\pm$2 \\
OneVision \textsc{Full} & 86$\pm$2 & 74$\pm$3 & 86$\pm$1 & 83$\pm$1 & 94$\pm$5 & 89$\pm$5 \\
\midrule
Qwen2.5-VL base & 57$\pm$1 & 55$\pm$2 & 62$\pm$0 & 56$\pm$3 & 52$\pm$3 & 57$\pm$5 \\
Qwen2.5-VL \textsc{Workspace} & 75$\pm$1 & 66$\pm$0 & 83$\pm$2 & 70$\pm$3 & 66$\pm$2 & 77$\pm$7 \\
\midrule
NeXT-Mistral base & 38$\pm$1 & 25$\pm$3 & 56$\pm$1 & 47$\pm$2 & 44$\pm$3 & 50$\pm$4 \\
NeXT-Mistral \textsc{Workspace} & 90$\pm$2 & 82$\pm$2 & 89$\pm$2 & 96$\pm$1 & 97$\pm$3 & 94$\pm$5 \\
\midrule
Idefics3 base & 56$\pm$4 & 54$\pm$4 & 58$\pm$1 & 57$\pm$2 & 53$\pm$3 & 60$\pm$13 \\
Idefics3 \textsc{Workspace} & 91$\pm$0 & 81$\pm$1 & 86$\pm$1 & 87$\pm$1 & 97$\pm$1 & 98$\pm$1 \\
\bottomrule
\end{tabular}
\end{table}

\begin{table}[t]
\centering
\small
\caption{Variant-level accuracy for the \textsc{Workspace} cells, mean
over three stimulus seeds, tasks pooled. Values percent.}
\label{tab:battery-variants}
\begin{tabular}{@{}lcccc@{}}
\toprule
Variant & OneVision & Qwen2.5-VL & NeXT-Mistral & Idefics3 \\
\midrule
standard & 95 & 79 & 97 & 98 \\
marker square & 92 & 77 & 94 & 94 \\
marker triangle & 91 & 80 & 86 & 96 \\
dots sparse & 89 & 87 & 90 & 89 \\
color green/orange & 65 & 56 & 72 & 84 \\
palette purple/orange & 70 & 68 & 86 & 78 \\
$n$ points $=1$ & 81 & 81 & 78 & 68 \\
$n$ points $=30$ & 94 & 79 & 93 & 97 \\
role reverse (red$\leftrightarrow$blue) & 69 & 62 & 95 & 77 \\
grid off & 93 & 66 & 97 & 97 \\
black background & 92 & 77 & 94 & 98 \\
\bottomrule
\end{tabular}
\end{table}

\paragraph{Invariance to geometry-preserving changes is universal.}
Marker shape, dot count, sparsity, grid removal, and background
inversion move the trained workspace models by at most a few points
(drops $\leq$$4$\,pp, one $-11$ cell on NeXT-Mistral's triangle
marker). Base and \textsc{Reader} models sit at or below chance under
every variant. They are appearance-invariant only in that they never
perceive.

\paragraph{Color is the systematic sensitivity.}
Novel palettes cost $11$--$30$\,pp at variant level in all four
architectures, always degrading toward but never below chance.
Red/blue role reversal splits the architectures. OneVision loses
$26$\,pp, Idefics3 $21$\,pp, and Qwen $17$\,pp, while NeXT-Mistral is
untouched ($-2$\,pp). The red/blue role convention is part of what
three of the four architectures learned. Both effects are consistent
with the routing account. Fine-tuning binds direction words partly to
the training appearance, with that binding weakening as the appearance diverges.

\paragraph{Two architecture-specific notes.}
Qwen's grid-off cell drops $13$\,pp where the other architectures gain
$0$--$5$. Qwen's trans\_y also interacts with point count ($95\%$ at
$n{=}1$ falling to $52\%$ at $n{=}30$) while its trans\_x stays flat
($87$--$97\%$). No other architecture shows either effect. These are
pathologies of Qwen's damaged trans\_y axis, not stimulus effects.
Separately, NeXT-Mistral's base sits at $25\%$ on the color family,
below the $50\%$ chance floor. This is constant-token bias, the same
failure family as the frontier constant-token collapse in
\S\ref{sec:exp1}.

\paragraph{The point-count sweep separates translation from scale.} Translation is readable from a single point in every architecture (trans\_y $84$--$97\%$ at $n{=}1$). Displacement is a first-order statistic. Scale climbs monotonically with point count in every architecture ($55$--$71\%$ at $n{=}1$ to $87$--$95\%$ at $n{=}30$). We do not consider the climb to be an information requirement, as a small number of points already determines the radial change.  Rather, we think this is a property of whatever statistic the trained readout consumes, whose reliability grows with dot count.  The pooled ``$n{=}1$ degrades everyone'' reading is entirely the scale component. This gives the benchmark's difficulty knob a mechanistic interpretation. Translation is carried by any single point, the cues for
scale thin as dots are removed, and the stimulus generator can starve
each task selectively.

\begin{table}[t]
\centering
\small
\caption{Point-count sweep, per task, \textsc{Workspace} cells, mean
over three stimulus seeds. Values percent.}
\label{tab:battery-npts}
\begin{tabular}{@{}lcccccc@{}}
\toprule
Arch / task & $n{=}1$ & $n{=}3$ & $n{=}5$ & $n{=}10$ & $n{=}20$ & $n{=}30$ \\
\midrule
OneVision trans\_x & 90 & 83 & 87 & 95 & 93 & 99 \\
OneVision trans\_y & 97 & 88 & 87 & 91 & 94 & 98 \\
OneVision scale & 57 & 57 & 55 & 71 & 81 & 87 \\
\midrule
Qwen trans\_x & 87 & 92 & 89 & 95 & 97 & 92 \\
Qwen trans\_y & 95 & 97 & 96 & 85 & 58 & 52 \\
Qwen scale & 61 & 73 & 85 & 91 & 93 & 94 \\
\midrule
NeXT trans\_x & 75 & 89 & 85 & 89 & 91 & 91 \\
NeXT trans\_y & 87 & 89 & 89 & 95 & 97 & 97 \\
NeXT scale & 71 & 70 & 72 & 85 & 90 & 93 \\
\midrule
Idefics3 trans\_x & 65 & 75 & 75 & 83 & 90 & 97 \\
Idefics3 trans\_y & 84 & 95 & 91 & 95 & 97 & 99 \\
Idefics3 scale & 55 & 61 & 67 & 81 & 86 & 95 \\
\bottomrule
\end{tabular}
\end{table}

\paragraph{Render resolution is a non-factor.}
Three architectures are flat within noise from half to double the
training resolution. OneVision gains mildly above it ($85\%$ at $256$px
to $96{-}97\%$ at $768$--$1024$px). No architecture degrades at
off-training resolutions.

\begin{table}[t]
\centering
\small
\caption{Render-resolution sweep, \textsc{Workspace} cells, mean over
three stimulus seeds, tasks pooled. Training render is $512$px. Values
percent.}
\label{tab:battery-res}
\begin{tabular}{@{}lccccc@{}}
\toprule
Arch & 256 & 384 & 512 (train) & 768 & 1024 \\
\midrule
OneVision & 85 & 86 & 95 & 97 & 96 \\
Qwen2.5-VL & 81 & 80 & 79 & 81 & 78 \\
NeXT-Mistral & 96 & 97 & 97 & 97 & 98 \\
Idefics3 & 98 & 98 & 98 & 98 & 98 \\
\bottomrule
\end{tabular}
\end{table}

\section{Comparison to Neighboring Spatial Benchmarks}
\label{app:benchmark-comparison}

Table~\ref{tab:benchmark-comparison} situates Sophon against the
spatial benchmarks discussed in \S\ref{sec:related}. The prior-stripped
phenomenon are considered by FUGU and BlindTest. No prior benchmark treats the coordinate system as a manipulated variable, crosses it factorially with parametric transforms, and pairs the result with an in-benchmark failure localization. The existence of a perception-readout gap is known to the literature; Sophon allows for a factorial dissection and localization of that gap.

\begin{table}[t]
\centering
\scriptsize
\setlength{\tabcolsep}{3pt}
\caption{Sophon vs.\ neighboring spatial benchmarks. What'sUp
\citep{kamath2023whatsup}, VSR \citep{liu2023vsr}, BLINK
\citep{fu2024blink}, Eyes Wide Shut / MMVP \citep{tong2024eyes},
BlindTest \citep{rahmanzadehgervi2024blind}, FUGU
\citep{tartaglini2025diagnosing}.}
\label{tab:benchmark-comparison}
\begin{tabular}{@{}>{\raggedright\arraybackslash}p{2.9cm}*{7}{>{\raggedright\arraybackslash}p{1.3cm}}@{}}
\toprule
 & \textbf{Sophon} & What'sUp & VSR & BLINK & Eyes Wide Shut & BlindTest & FUGU \\
\midrule
Stimuli & point clouds & object / CLEVR photos & natural images & natural images & CLIP-blind pairs & 2D primitives & scatter plots, labeled axes \\
Obj./scene/lang.\ priors & \textbf{none} & objects, language & objects, scene, language & objects, scene & object semantics & mostly stripped & mostly stripped \\
Ground truth & procedural, exact & template & annotation & annotation & curated & procedural & procedural, exact \\
Coord-system factorial & \textbf{yes} & no & no & no & no & no & no \\
Parametric transforms & \textbf{T, R, S} & relational only & relational only & mixed & no & no & no \\
Continuous difficulty knob & \textbf{yes} & no & no & no & no & partial & partial (point count) \\
Human + frontier solvability & \textbf{yes} & partial & no & models only & models only & models only & models only \\
In-paper failure localization & \textbf{yes} & no & no & no & CLIP-feature diff & no & yes (handoff) \\
\bottomrule
\end{tabular}
\end{table}

\paragraph{Limitations compared to other benchmarks.}
The prior-stripping designed to provide clean geometry also removes
dimensions that some neighboring benchmarks cover. 
\begin{itemize}
\item \emph{Naturalistic validity.} What'sUp, VSR, BLINK, and Eyes
Wide Shut use real photographs. Sophon's abstraction does not by
itself establish real-world spatial competence.
\item \emph{Compositional spatial language.} What'sUp and VSR test
relational language tied to objects. Sophon uses fixed templates with
minimal linguistic variety.
\item \emph{Multi-step relational chains.} Sophon is single-transform,
single-question.
\item \emph{Answer-format diversity.} Sophon's answer formats are
deliberately constrained.
\end{itemize}

\section{Human Baseline Protocol}
\label{app:human-baseline}

\subsection{Recruitment and compensation}
Participants were recruited via Prolific (\texttt{prolific.com}). 31 raters
were recruited and 30 passed attention check criteria ($\geq$2 of 3
attention check items correct). One rater with 2 failures was excluded,
and one additional record from an internal demo session was excluded
prior to analysis. Recruitment required raters fluent in English with a
$\geq$95\% prior approval rate on Prolific, $\geq$10 prior submissions,
normal or corrected to normal vision, and
normal color vision (stimuli use red and blue point sets and require
red/blue discrimination). No country of residence restriction was applied.
Each participant was paid \$8.25 USD for a study estimated at 15 minutes (effective
rate $\approx$\$33/hr). 

\subsection{Study protocol}
Each rater completed a single 31 item session of 24 Sophon items (3 per
categorical, binary, or comparative cell across 8 cells), 4 quantitative
items (one per quantitative cell, sampled at moderate magnitude), and 3
attention check items at fixed within session positions (8, 17, 26).
Within session item order was deterministically randomized per rater from a
per rater seed. Stimuli were generated on demand by the Sophon generator
(the same code path used for the model evaluation set). Each rater saw
freshly rendered images with the same generator produced ground truth used
to score the VLMs. Raters answered in a Next.js web UI optimized for
desktop and mobile. Per item render and submit times were logged
server side.

\subsection{Attention checks and exclusion}
Three items per session were attention checks, Sophon formatted items with
trivially obvious answers (e.g., a binary alignment question on a stimulus
where the two point sets are visibly identical or visibly far apart).
Workers failing $>$1 of 3 attention checks were excluded from analysis but
still paid in full. Of 31 recruited raters, 26 had zero attention failures,
4 had a single failure (retained), and 1 had two failures (excluded). The
attention check exclusion rate was therefore $3.2\%$.

\subsection{Quantitative scale prompt and unit ambiguity}
For the four quantitative cells we asked each rater a single magnitude
question per cell. Rotation prompts asked for an angle in degrees and were
scored against generator produced ground truth in radians (converted to
degrees for the comparison). Human MAE was $24.1^\circ$ for Cartesian
rotation and $20.8^\circ$ for polar rotation. Scale prompts asked for the
factor by which the second point set was scaled relative to the first. We
observed substantial heterogeneity in answer interpretation, with raters
giving answers consistent with several different readings of the prompt.
The actual top response values across $n{=}30$ raters per cell illustrate
the issue. On Cartesian scale the most common answers were
$\{2.0~(\!\times{}8),~3.0~(\!\times{}4),~1.5~(\!\times{}3),~10~(\!\times{}2),~20~(\!\times{}2),~-4~(\!\times{}2)\}$
against a GT factor range of $[0.5, 2.0]$, with one rater entering values
as large as 50 and several entering negative numbers. Rotation answers
behaved as expected (the degree unit is unambiguous) and clustered on
round numbered psychophysical estimates such as $\{60^\circ, 45^\circ,
5^\circ, 90^\circ\}$. We therefore exclude human scale numbers from the
main paper quantitative comparison. Descriptively, mean absolute deviation
against the factor GT was $5.1$ for Cartesian scale and $3.5$ for polar
scale, but these values are not interpretable as a perception measurement
because the prompt admits factor, percent, and signed fold change readings
that produce wildly different numeric outputs from the same perceptual
state. The data we collect therefore supports a strong conclusion about
\emph{rotation} (humans achieve $r{=}0.69$ on polar rotation, the only
perceiver across all (model, cell) combinations) and a weak conclusion
about \emph{scale} (rater unit interpretation dominates the signal). A
follow up study with explicit unit instruction (e.g., ``enter the factor
$f$ such that red points are $f$ times the blue spread, where $f{=}1$
means equal'') would resolve the confound.

\subsection{Ethics and consent}
Participants saw an informed consent page before beginning. Consent
required an affirmative checkbox and was a precondition for entering the
study. The study collected only anonymous task responses (Prolific PID,
study/session IDs, item level answers and timings).  No personally
identifying information beyond what Prolific's platform retains.
Participants could withdraw at any point. As the authors are not
affiliated with an institution that maintains an IRB, no IRB review
applied. The protocol followed standard ethics practice for low risk
anonymous behavioral research and Prolific's research ethics guidelines.

\subsection{Data release}
Per rater item level responses (with rater IDs replaced by per session
opaque hashes) are released alongside the benchmark, including answer,
ground truth, scoring outcome, item type, cell ID, and timing fields.

\section{Vocabulary Distractor Detail}
\label{app:vocab-distractor}

\begin{figure}[!htbp]
\centering
\includegraphics[width=\linewidth]{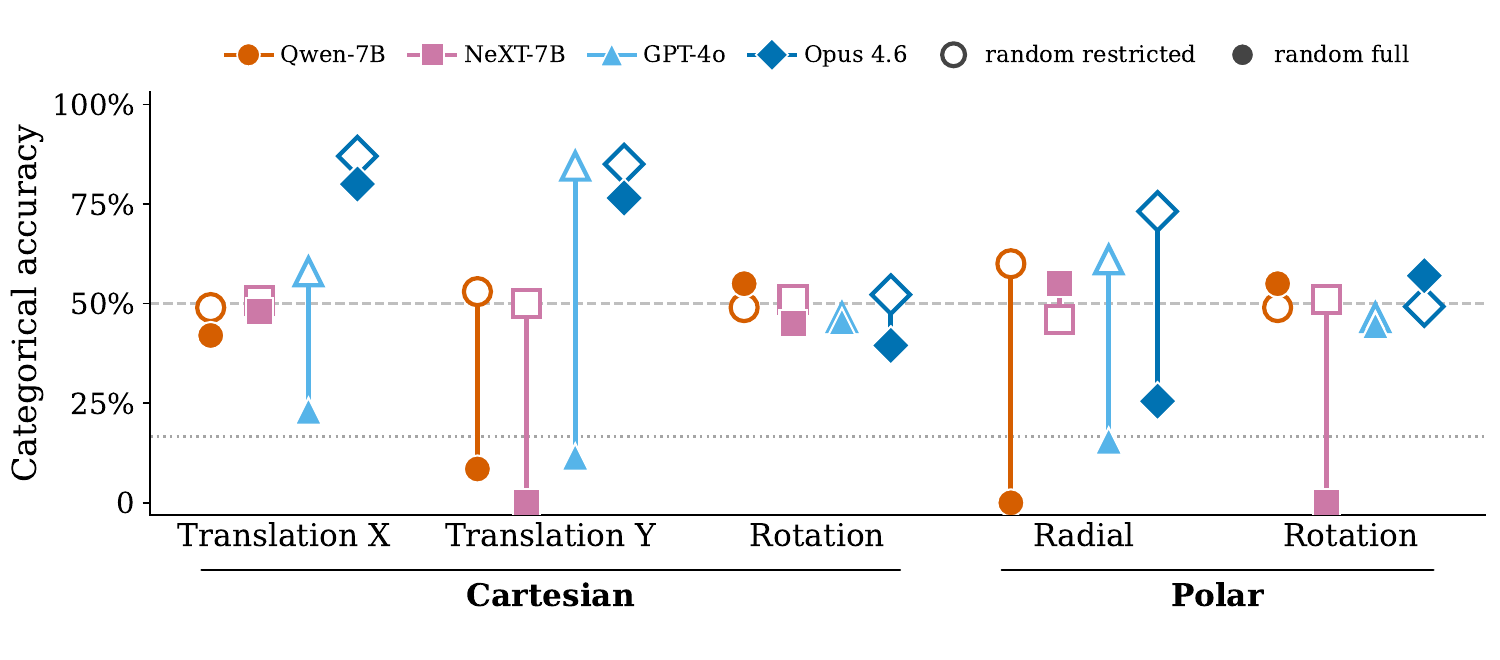}
\caption{Vocabulary distractor effect across five categorical specs.
Each dumbbell connects \textit{random restricted} accuracy (open
marker, 2 transform appropriate direction words offered) to
\textit{random full} accuracy (filled marker, all 6 direction words
offered, order randomized in both conditions). The translation
(Cartesian \textit{Translation X} / \textit{Translation Y}) and polar
\textit{Radial} (radial translation, answer choices
\emph{inward} / \emph{outward}) cells plunge toward the 6-way chance
floor under the full vocabulary across both open (Qwen-7B, NeXT-7B) and
frontier (GPT-4o, Opus 4.6) tiers.  The extra rotation tokens
(\textit{clockwise}, \textit{counterclockwise}) draw probability mass
away from the correct translation answer even though the underlying
transform is not rotational. The Cartesian \textit{Rotation} control
stays flat because rotation vocabulary is
already among the choices in both conditions, so widening the option
set adds no novel distractors. The pattern is consistent across
capability tiers, supporting the claim that the failure is at the
answer set interface rather than at perception.}
\label{fig:vocab-distractor}
\end{figure}

Figure~\ref{fig:vocab-distractor} shows the full per-(spec, model)
vocabulary distractor effect referenced in \S\ref{sec:vocab-ablation}.
For each (spec, model) pair the dumbbell connects accuracy under
restricted (transform appropriate 2-way; open marker) vocabulary to
accuracy under the full 6-way vocabulary (filled marker). Translation
and polar radial cells plunge toward the 6-way chance floor under the
full vocabulary across both open (Qwen-7B, NeXT-7B) and frontier
(GPT-4o, Opus 4.6) tiers, while the Cartesian rotation control stays
flat (rotation vocabulary is already among the choices in both
conditions).

\paragraph{Positional ordering shortcut.}
\label{app:vocab-order}
Positional ordering of the 6-way options reveals that some
fine tuned open models have learned a \emph{positional} shortcut on
top of the vocabulary shortcut. LLaVA-NeXT fine tuned on
fixed order 2-way vocab jumps from $53\%$ (random) to $90\%$
(correct last) on Cartesian rotation, but only because the
correct last position aligns with the learned rotation vocab
attraction. Translation specs remain at $0\%$ in correct last,
confirming the effect is not pure position bias but semantic
attraction amplified by position.

\section{Range Hints: A Crutch That Hurts Frontier Estimation}
\label{app:range-hint}

Range hints (telling the model ``the factor is between 0.70 and
1.30'') are a common prompt engineering practice for quantitative
tasks. Their effect depends on whether the underlying model estimates
genuinely or by anchoring. For open models, hints shift
the modal answer toward the hint center and often lower MAE because the
new anchor sits closer to the ground truth mean without changing the
near zero correlation with actual stimuli. For Opus~4.6, which \emph{does}
estimate, the same hint \emph{hurts} (Table~\ref{tab:rangehint}).

\begin{table}[t]
\centering
\small
\caption{Opus~4.6 magnitude estimation, no hint vs.\ range hint. The hint
collapses Opus~4.6's stimulus sensitive estimation toward canonical values
($15^\circ$, 0.85, 0.95) and drops correlation sharply on polar scale.
Numbers filtered to plausible range. Pixel coordinate leakage in raw
responses fixed upstream (\S\ref{sec:setup}).}
\label{tab:rangehint}
\begin{tabular}{@{}lcc|cc@{}}
\toprule
 & \multicolumn{2}{c|}{No hint} & \multicolumn{2}{c}{With hint} \\
Spec & MAE & Pearson $r$ & MAE & Pearson $r$ \\
\midrule
Cartesian rotation (deg) & 10.3 & 0.43 & 17.2 & $-0.19$ \\
Cartesian scale (factor) & 0.09 & 0.55 & 0.13 & 0.55 \\
Polar rotation (deg)     & 15.2 & 0.29 & 15.1 & 0.10 \\
Polar scale (factor)     & \textbf{0.06} & \textbf{0.86} & 0.35 & 0.31 \\
\bottomrule
\end{tabular}
\end{table}

The polar scale row is the decisive comparison. Without hints, Opus~4.6's
answer distribution tracks the ground truth distribution (Pearson
$r{=}0.86$, MAE 0.06), providing the cleanest evidence in the study that a VLM
can perform genuine continuous valued magnitude estimation from visual
content. With a range hint, correlation drops to $r{=}0.31$ and MAE
rises 6$\times$. The hint pushes Opus~4.6 to emit canonical value answers
(0.85, 0.95) and overrides its stimulus driven estimation. Range hints
help a model that was already anchoring and hurt a model that was
succeeding.  This is a clean diagnostic criterion for separating the two.

\section{Quantitative Answer Distribution Taxonomy: All Models}
\label{app:quant-taxonomy}

\begin{figure}[!htbp]
\centering
\includegraphics[width=\linewidth]{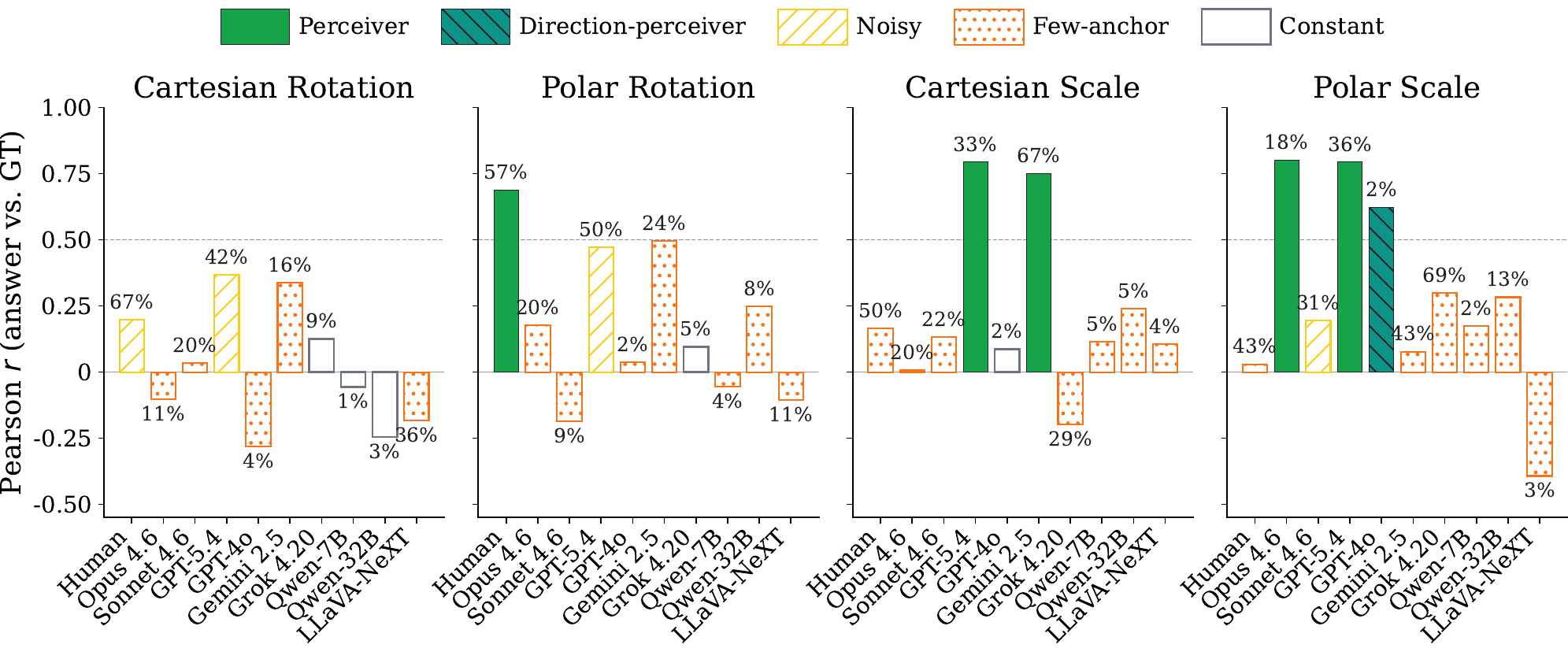}
\caption{All models version of the quant response taxonomy
(Fig.~\ref{fig:quant-taxonomy} shows only Human plus the four
current/recent closed weight flagships). Adds Grok 4.20 and four open
$7$B--$32$B models. The five classes and the color scheme match the
main-paper figure: \textit{Perceiver}, \textit{Direction perceiver},
\textit{Noisy}, \textit{Few anchor}, \textit{Constant} (neutral grey).
Numerical labels are the percentage of responses that are unique values.
None of the additional models reaches the perceiver or
direction perceiver class on any cell. Qwen2.5-VL-7B Cartesian rotation
($u{=}1\%$, top-1${=}98\%$) and Grok 4.20 polar-rotation
($u{=}5\%$, top-1${=}97\%$) are canonical \textit{Constant} cells.
LLaVA-NeXT polar rotation ($u{=}11\%$, top-1${\sim}37\%$, $r{=}-0.11$) is
the noisiest entry, with many distinct guesses and none correlated with GT.}
\label{fig:quant-taxonomy-all}
\end{figure}

The main text Fig.~\ref{fig:quant-taxonomy} shows only Human plus
the current/recent closed weight flagships. Figure~\ref{fig:quant-taxonomy-all}
extends the taxonomy to Grok-4.20 and four open VLMs ($7$B--$32$B)
for completeness. The headline observation is that none of the
additional models reaches \textit{perceiver} or
\textit{direction perceiver} on any cell. Quantitative magnitude
tracking is a frontier-only capability across the systems we test,
and the across the board collapse to \textit{constant} or
\textit{few anchor} on the open VLM side gives a different reading
of MAE than the categorical cell accuracy alone provides.

\section{Limitations: Full Discussion}
\label{app:limitations}

This appendix expands each limitation listed in
\S\ref{sec:limitations}.

\paragraph{Controlled abstract regime.}
Sophon deliberately operates in a controlled abstract stimulus regime,
plain background point clouds with no objects, scenes, or textures.
This is the property that makes the benchmark diagnostic. The same
property prevents it from \emph{measuring} how a VLM behaves on the
naturalistic stimuli that constitute most real world deployment
contexts. Claims here are therefore about whether a VLM can perform
spatial reasoning \emph{when language and object priors are unavailable},
not about its naturalistic spatial competence as a whole. A naturalistic
comparator that strips object/language priors with the same control as
Sophon is methodologically non-trivial (the very priors we strip are
near impossible to remove from natural images without altering content)
and is out of scope for this paper. Sophon is intended to complement,
not replace, naturalistic spatial benchmarks.

\paragraph{Open model scale.}
Our open model experiments span 7B (Qwen2.5-VL, LLaVA-NeXT-Mistral,
LLaVA-OneVision) and 32B (Qwen2.5-VL-AWQ, zero-shot and CoT v2;
fine tuning of the AWQ-quantized 32B at the 1k volume point is omitted
because the standard LoRA path is incompatible with AWQ-quantized
weights and re-loading the unquantized 32B exceeds a single A6000).
We do not test 72B or larger open VLMs. The frontier contrast (Opus~4.6,
GPT-5.4) already spans the most extreme capability gap, and the
mechanism we document (vocabulary shortcuts, encoder probe ceiling,
readout-routing bottleneck) is not obviously tied to a specific parameter count. The 32B zero-shot results we already
report (mean $54.7\%$ vs.\ 7B's $51.1\%$) show no broad improvement on
Sophon at $4\times$ language side capacity, which is itself
evidence against a capacity bound hypothesis.

\paragraph{Cross-coordinate seed coverage.}
Cross-coord variance is reported at $n=3$ for Qwen2.5-VL and $n=2$ for
LLaVA-OneVision and LLaVA-NeXT. The asymmetric transfer pattern is striking
enough that additional seeds are unlikely to flip the qualitative
finding, but tighter error bars on the two non-Qwen architectures would
firm up the magnitude estimates.

\paragraph{Projector specific claims.}
Our readout-routing argument treats the encoder-decoder interface as a
single unit. The architectures we test use MLP, patch-merger, and
pixel-shuffle projectors. We
do not test cross-attention or Q-Former projectors, which may behave
differently. Our probes demonstrate that the encoder output is
informative. We do not separately probe post projector features, which
would localize the failure to the projector versus decoder attention.
This is a productive direction for future work.

\paragraph{Volume curve seed coverage.}
The fine tuning volume curve (500/1k/5k/10k) reports $n=3$ seeds at the
1k point and $n=1$ at the others. The 1k BLINK peak ($+4.4 \pm 0.8$\,pp
over base) is robustly above seed noise, but we do not have multi seed
variance for the 5k and 10k points. A full $3\times 4$ matrix would
require approximately 60 additional GPU hours and was beyond submission
budget.

\paragraph{Rotation: an unexplained model side failure.}
See Appendix~\ref{app:rotation-discussion} for an extended discussion
of why rotation stays near chance for VLMs while humans clear polar
rotation at $r{=}0.69$.

\paragraph{Frontier model coverage.}
We report extensive results for Claude~Opus~4.6 and GPT-5.4 and partial
results for GPT-4o, Gemini~2.5~Pro, and Grok~4.20. With
GPT-5.4 matching Opus's factor level scale estimation
(MAE~$\approx$~0.06 on both Cartesian and polar scale), the genuine
magnitude estimation finding is no longer a single model anchor. Broader
frontier coverage (in particular Claude~Sonnet~4.6 and GPT-4.1)
would establish whether the gap between current and prior generation
flagships seen here (GPT-4o 74.8\%, GPT-5.4 81.4\%) reflects a clean
generational improvement or a single axis closure on Sophon's specific
task structure. A $5$-spec temperature ablation
($\tau \in \{0, 0.25, 0.5, 0.75, 1.0\}$) for Grok-4.20 alongside the
other four frontier models is reported in
Appendix~\ref{app:grok-degen}, and confirms that Grok's constant token
collapse is a persistent attractor not relieved by stochastic decoding.
Extending the ablation to all $23$ cells, with matched decode controls
on additional open base models, is out of scope for this paper.

\paragraph{Human baseline scope.}
Our human baseline is $n{=}30$ Prolific raters who completed a 30 item
session each. This is sufficient for tight 95\% Wilson confidence intervals
on per-cell categorical, binary, and comparative accuracy ($\pm$5--10\,pp
depending on cell) but does not provide a per-magnitude difficulty curve at
full statistical power for the quantitative cells, where each rater answered
only one item per cell. A larger psychophysics study (per-rater 200 item
sessions, per-magnitude binning, and explicit unit instruction for scale
prompts) would be more authoritative for quantitative magnitude estimation.
We view our baseline as a benchmark anchor rather than a perception research
grade study.

\section{Per-Cell Accuracy and MAE Tables}
\label{app:per-cell}

The diagnostic heatmap (Fig.~\ref{fig:diagnostic-heatmap}) and Table~\ref{tab:main}
report aggregates over the 23 (coordinate system, question type) cells of
the Sophon suite. Tables~\ref{tab:per-cell-frontier},
\ref{tab:per-cell-open}, and \ref{tab:per-cell-mae} give the underlying
per-cell numbers across all evaluated models. ``Cmp Bin'' /
``Cmp Comp'' refer to compound transform binary and comparative cells;
``Rot'' / ``Scl'' / ``TrX'' / ``TrY'' / ``Rad'' / ``RadScl'' map to the
rotation, scale, translation x, translation y, polar radial, and
polar radial scale specs respectively.

\begin{table}[t]
\centering
\scriptsize
\caption{Per-cell accuracy and quantitative MAE, aggregated across specs to match Fig.~\ref{fig:diagnostic-heatmap}. Accuracy in $\%$. Rotation MAE in degrees, scale MAE in factor units. Human values are from $n{=}30$ Prolific raters; human scale-MAE is omitted because raters interpreted the factor prompt heterogeneously (factor vs.~percent), see Appendix~\ref{app:human-baseline}. OneVision polar-scale MAE is computed over the $30\%$ of items that parsed to a number (constant non-numeric tokens on the rest). Cross-seed standard deviations $\pm s$ are shown in the heatmap (this table reports point estimates).}
\label{tab:per-cell}
\begin{tabular}{@{}l r rrrr rrrr@{}}
\toprule
 & Human & \multicolumn{4}{c}{Frontier} & \multicolumn{4}{c}{Open} \\
\cmidrule(lr){3-6} \cmidrule(lr){7-10}
Cell
  &
  & \rotatebox{45}{Opus 4.6}
  & \rotatebox{45}{Gemini 2.5 Pro}
  & \rotatebox{45}{GPT-5.4}
  & \rotatebox{45}{Grok 4.20}
  & \rotatebox{45}{Qwen-7B}
  & \rotatebox{45}{Qwen-32B}
  & \rotatebox{45}{LLaVA-NeXT}
  & \rotatebox{45}{OneVision} \\
\midrule
1D Lin Bin & 86 & 76 & 64 & 52 & 45 & 53 & 54 & 48 & 64 \\
1D Lin Cat & 94 & 92 & 84 & 98 & 57 & 44 & 92 & 40 & 46 \\
2D Cart Bin & 87 & 88 & 89 & 95 & 48 & 60 & 49 & 49 & 54 \\
2D Cart Cat & 88 & 78 & 76 & 48 & 48 & 50 & 71 & 51 & 40 \\
2D Cart Comp & 84 & 68 & 86 & 92 & 53 & 52 & 51 & 48 & 56 \\
Polar Bin & 88 & 91 & 92 & 96 & 47 & 52 & 52 & 48 & 56 \\
Polar Cat & 71 & 61 & 65 & 54 & 50 & 55 & 45 & 48 & 52 \\
Polar Comp & 88 & 61 & 73 & 95 & 51 & 53 & 53 & 48 & 72 \\
\midrule
Cart Rot MAE ($^\circ$) & 24.1 & 16.1 & 19.8 & 14.7 & 22.7 & 22.3 & 26.7 & 63.2 & 21.4 \\
Polar Rot MAE ($^\circ$) & 20.8 & 25.5 & 18.5 & 29.8 & 28.3 & 19.1 & 111.0 & 100.9 & 21.4 \\
Cart Scale MAE & --- & 0.87 & 0.15 & 0.06 & 9.32 & 0.70 & 0.76 & 1.95 & 0.26 \\
Polar Scale MAE & --- & 0.07 & 0.23 & 0.07 & 82.83 & 0.60 & 0.69 & 0.80 & 0.65$^{\dagger}$ \\
\bottomrule
\end{tabular}
\end{table}

\begin{table}[t]
\centering
\scriptsize
\caption{Per-cell accuracy ($\%$): frontier models. Cells are coordinate-system $\times$ spec $\times$ question type. Reparsed comparative scores used where available (\S\ref{sec:exp1}).}
\label{tab:per-cell-frontier}
\begin{tabular}{@{}lrrrrr@{}}
\toprule
Cell
  & \rotatebox{45}{Opus 4.6}
  & \rotatebox{45}{Gemini 2.5 Pro}
  & \rotatebox{45}{GPT-5.4}
  & \rotatebox{45}{GPT-4o}
  & \rotatebox{45}{Grok 4.20} \\
\midrule
1D-Lin $\cdot$ Bin & 47 & 71 & 52 & 72 & 45 \\
1D-Lin $\cdot$ Cat & 94 & 84 & 98 & 64 & 57 \\
\midrule
2D-Cart $\cdot$ Cmp Bin & 87 & 90 & 98 & 92 & 53 \\
2D-Cart $\cdot$ Cmp Comp & 69 & 84 & 92 & 78 & 59 \\
2D-Cart $\cdot$ Rot Bin & 85 & 93 & 93 & 82 & 55 \\
2D-Cart $\cdot$ Rot Cat & 52 & 47 & 46 & 46 & 48 \\
2D-Cart $\cdot$ Scl Bin & 78 & 73 & 85 & 82 & 39 \\
2D-Cart $\cdot$ Scl Comp & 66 & 94 & 91 & 72 & 45 \\
2D-Cart $\cdot$ TrX Bin & 89 & 93 & 97 & 92 & 47 \\
2D-Cart $\cdot$ TrX Cat & 87 & 80 & 82 & 58 & 48 \\
2D-Cart $\cdot$ TrY Bin & 95 & 91 & 100 & 92 & 47 \\
2D-Cart $\cdot$ TrY Cat & 85 & 93 & 86 & 84 & 49 \\
\midrule
Polar $\cdot$ Rad Bin & 97 & 93 & 100 & 92 & 37 \\
Polar $\cdot$ Rad Cat & 73 & 80 & 57 & 61 & 52 \\
Polar $\cdot$ Rad Comp & 56 & 66 & 92 & 65 & 56 \\
Polar $\cdot$ RadScl Bin & 88 & 91 & 90 & 84 & 43 \\
Polar $\cdot$ RadScl Comp & 66 & 91 & 97 & 62 & 46 \\
Polar $\cdot$ Rot Bin & 87 & 95 & 99 & 97 & 60 \\
Polar $\cdot$ Rot Cat & 49 & 50 & 52 & 46 & 48 \\
\bottomrule
\end{tabular}
\end{table}

\begin{table}[t]
\centering
\scriptsize
\caption{Per-cell accuracy ($\%$): open base and fine-tuned models. LoRA-language variants apply LoRA only to the language tower; LoRA-full updates both vision and language towers (full-vocab fine-tune, 1k samples). Reparsed comparative scores used where available.}
\label{tab:per-cell-open}
\begin{tabular}{@{}l rr rrr r@{}}
\toprule
 & \multicolumn{2}{c}{LLaVA-NeXT-7B} & \multicolumn{3}{c}{Qwen2.5-VL-7B} & \\
\cmidrule(lr){2-3} \cmidrule(lr){4-6}
Cell
  & base & LoRA-lang
  & base & LoRA-full & LoRA-lang
  & Qwen2.5-VL-32B \\
\midrule
1D-Lin $\cdot$ Bin & 47 & 86 & 57 & 75 & 57 & 50 \\
1D-Lin $\cdot$ Cat & 40 & 88 & 44 & 96 & 82 & 92 \\
\midrule
2D-Cart $\cdot$ Cmp Bin & 49 & 100 & 60 & 100 & 86 & 48 \\
2D-Cart $\cdot$ Cmp Comp & 49 & 90 & 51 & 90 & 94 & 62 \\
2D-Cart $\cdot$ Rot Bin & 56 & 100 & 52 & 100 & 85 & 56 \\
2D-Cart $\cdot$ Rot Cat & 51 & 49 & 49 & 51 & 51 & 48 \\
2D-Cart $\cdot$ Scl Bin & 44 & 100 & 56 & 100 & 73 & 44 \\
2D-Cart $\cdot$ Scl Comp & 47 & 100 & 53 & 98 & 100 & 63 \\
2D-Cart $\cdot$ TrX Bin & 48 & 100 & 63 & 98 & 92 & 48 \\
2D-Cart $\cdot$ TrX Cat & 51 & 73 & 49 & 59 & 62 & 82 \\
2D-Cart $\cdot$ TrY Bin & 48 & 100 & 72 & 100 & 97 & 48 \\
2D-Cart $\cdot$ TrY Cat & 50 & 8 & 53 & 48 & 52 & 82 \\
\midrule
Polar $\cdot$ Rad Bin & 44 & 100 & 56 & 100 & 94 & 47 \\
Polar $\cdot$ Rad Cat & 46 & 80 & 60 & 78 & 92 & 35 \\
Polar $\cdot$ Rad Comp & 48 & 98 & 52 & 86 & 92 & 52 \\
Polar $\cdot$ RadScl Bin & 44 & 100 & 56 & 100 & 77 & 48 \\
Polar $\cdot$ RadScl Comp & 48 & 97 & 53 & 84 & 92 & 45 \\
Polar $\cdot$ Rot Bin & 56 & 100 & 44 & 100 & 96 & 60 \\
Polar $\cdot$ Rot Cat & 51 & 48 & 49 & 51 & 51 & 54 \\
\bottomrule
\end{tabular}
\end{table}

\begin{table}[t]
\centering
\scriptsize
\caption{Per-cell quantitative MAE (rotation in degrees, scale as factor). Lower is better. Dashes indicate the model emitted a constant token (zero variance) on that cell, which makes MAE-vs-truth a misleading summary; see the per-model degeneracy notes in \S\ref{sec:exp1}.}
\label{tab:per-cell-mae}
\begin{tabular}{@{}lrrrr@{}}
\toprule
Model & Cart Rot ($^\circ$) & Polar Rot ($^\circ$) & Cart Scale & Polar Scale \\
\midrule
\multicolumn{5}{@{}l}{\textit{Frontier}} \\
Opus 4.6           & 16 & 26  & 0.87 & 0.07 \\
Gemini 2.5 Pro     & 20 & 19  & ---  & ---  \\
GPT-4o             & 48 & 32  & 0.96 & 0.75 \\
Grok 4.20          & 23 & 28  & 9.32 & 82.83 \\
\midrule
\multicolumn{5}{@{}l}{\textit{Open}} \\
LLaVA-NeXT-7B            & 63  & 101 & 1.95 & 0.80 \\
LLaVA-NeXT LoRA-language & 16  & 15  & 0.05 & 0.05 \\
Qwen2.5-VL-7B            & 22  & 19  & 0.70 & 0.60 \\
Qwen LoRA-full           & 10  & 11  & 0.20 & 0.21 \\
Qwen LoRA-language       & 9   & 15  & 0.05 & 0.04 \\
Qwen2.5-VL-32B           & 27  & 111 & ---  & ---  \\
\bottomrule
\end{tabular}
\end{table}

\section{Opus~4.6 Magnitude Estimation Distribution}
\label{app:opus-magnitude}

\begin{figure}[!htbp]
\centering
\includegraphics[width=\linewidth]{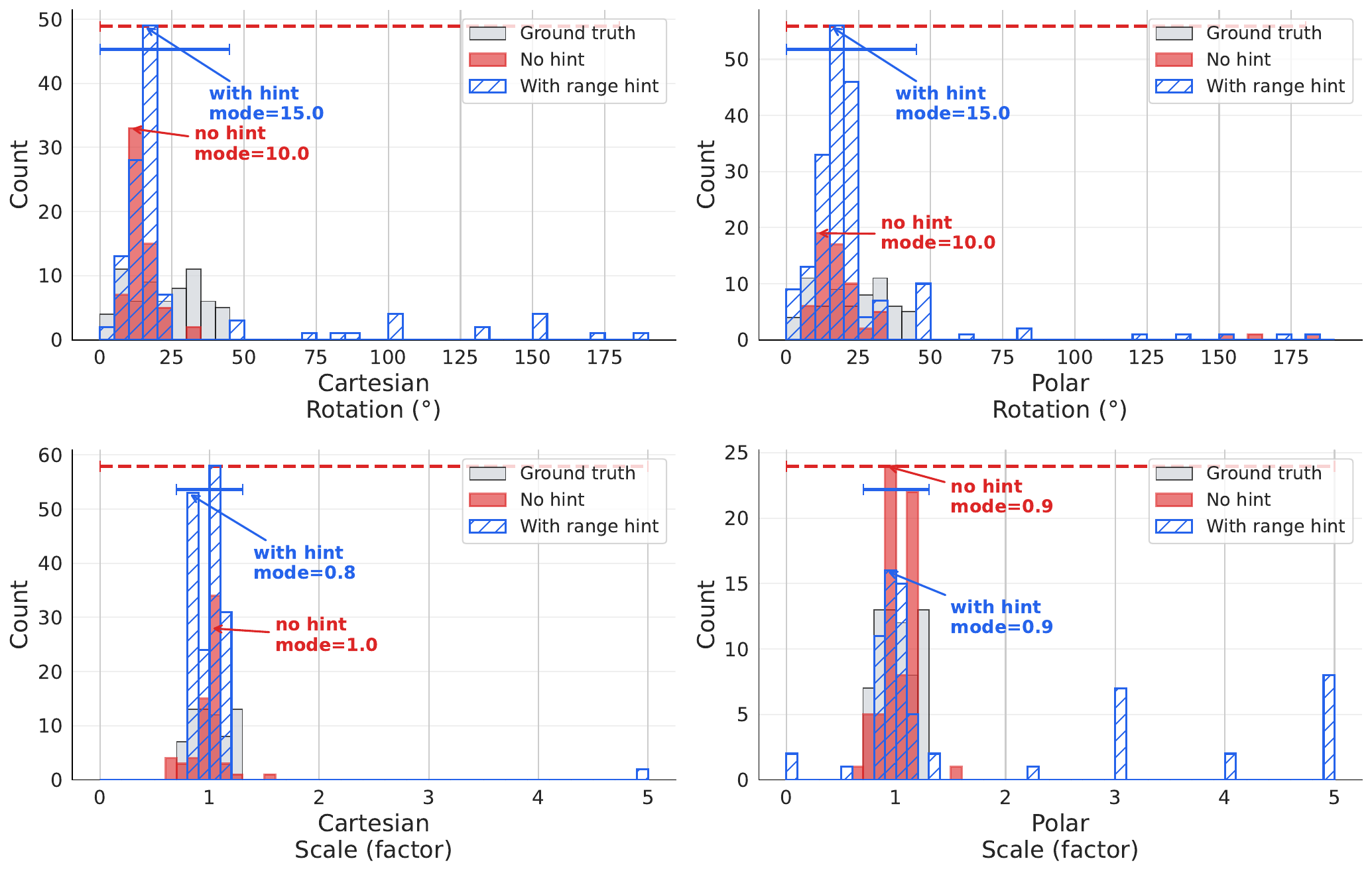}
\caption{\textbf{Opus 4.6.} Quantitative answer distribution vs.\
ground truth on the four (coordinate system, transform) cells. Each
panel overlays the ground truth histogram (gray fill) with no hint
predictions (red, solid edge) and range hint predictions (blue,
hatched). The two horizontal range bars at the top show the no hint
stimulus support (red dashed) and the explicit range supplied by the
hint (blue solid), and mode of response annotations point at the dominant
text anchor in each condition. The polar scale panel is the only cell
where Opus~4.6 tracks the stimulus continuously ($r{=}0.80$, perceiver
class). The other three panels collapse to a single mode near
$90^\circ$ or $\sim$1.0 regardless of GT (canonical \textit{few anchor}).
Range hints shrink the support of the predicted distribution but do
not move the mode toward GT, which is the basis for our claim that
hints constrain the answer set rather than reveal perception.}
\label{fig:magnitude-anchoring}
\end{figure}

Figure~\ref{fig:magnitude-anchoring} shows Opus 4.6's predicted vs. ground truth scatter on the four quantitative cells (Cartesian and
polar, rotation and scale). The polar scale panel exhibits the
$r{=}0.86$ continuous correlation referenced in \S\ref{sec:exp1}. A
unit disambiguated re-prompt drops this to $r{=}0.70$ but stays
above the perceiver threshold (\S\ref{sec:quant-taxonomy}). The
remaining three panels show the failure modes characterized by the
taxonomy. Cartesian scale is \textit{noisy} ($r{=}0.01$), and
Cartesian rotation and polar rotation are \textit{few anchor},
clustering near $90^\circ$ regardless of GT. Even the strongest
single model signal we observe on Sophon does not generalize across
the four (coordinate system, transform) cells.

\begin{figure}[!htbp]
\centering
\includegraphics[width=\linewidth]{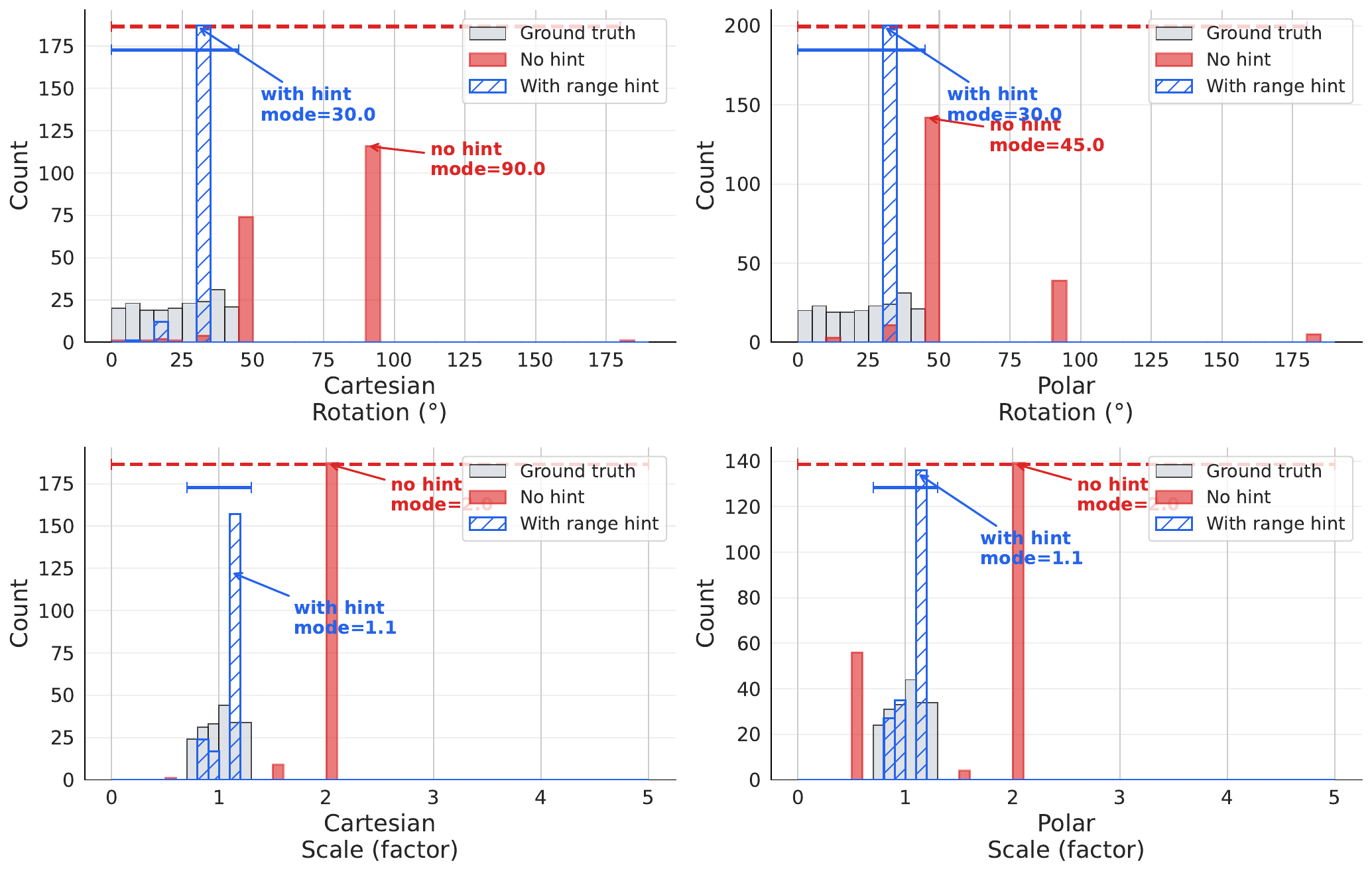}
\caption{\textbf{GPT-4o.} Same layout as Fig.~\ref{fig:magnitude-anchoring}.
Polar scale is the lone non degenerate cell ($r{=}0.62$,
direction perceiver, meaning the model tracks scale direction up vs.\ down
but does not produce continuous magnitude). Cartesian scale collapses
to a single anchor ($r{=}0.09$, constant), and both rotation cells are
\textit{few anchor} at $\sim$$45^\circ$.}
\label{fig:magnitude-anchoring-gpt4o}
\end{figure}

\begin{figure}[!htbp]
\centering
\includegraphics[width=\linewidth]{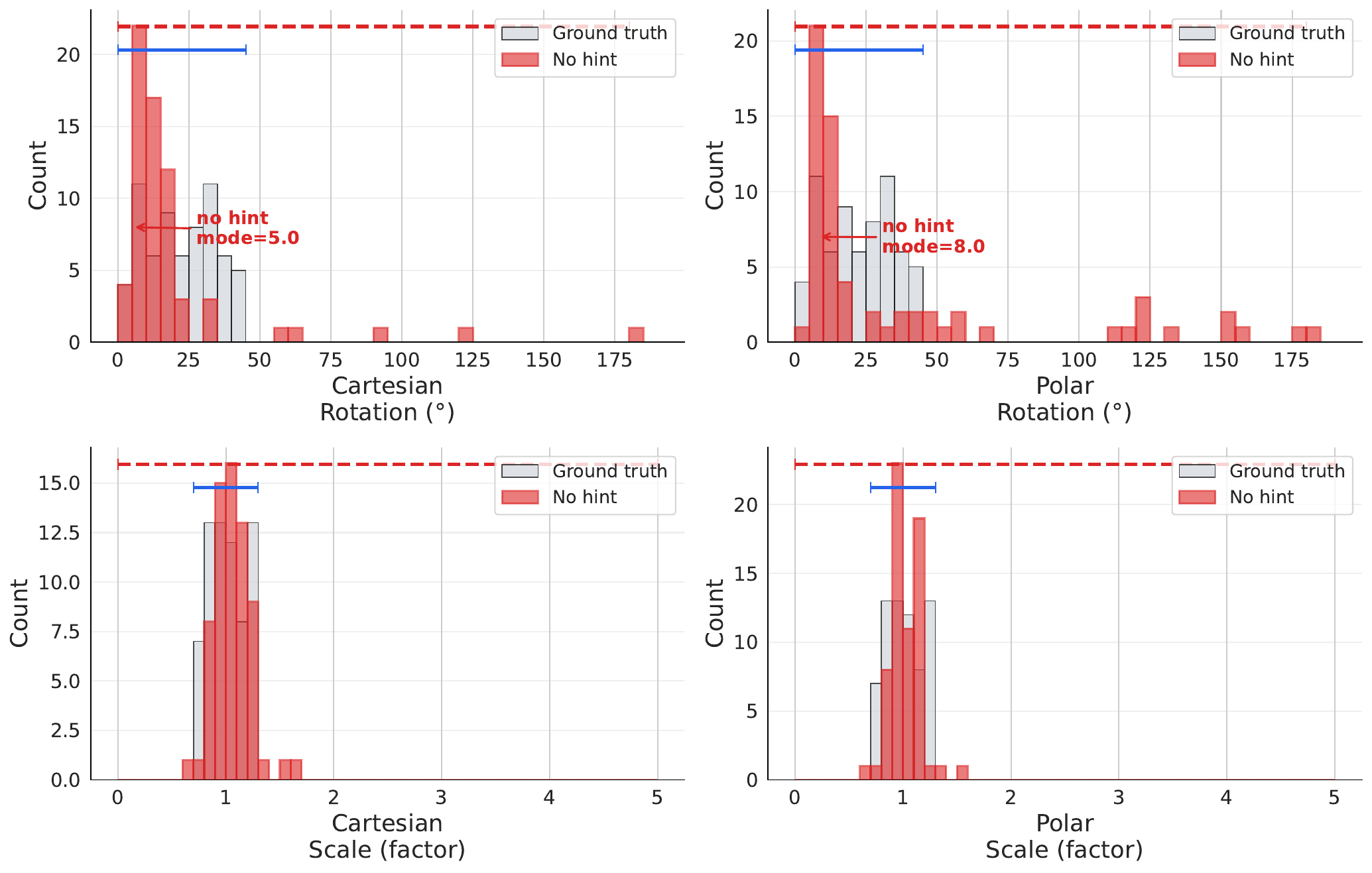}
\caption{\textbf{GPT-5.4.} Same layout as
Fig.~\ref{fig:magnitude-anchoring} (no range hint condition was run
for GPT-5.4, so the blue series is absent). Both scale cells reach
\textit{perceiver} ($r{=}0.79$ each), the only frontier model in our
suite that does so on Cartesian scale. Rotation distributions remain
\textit{noisy} ($r{=}0.37$ Cartesian, $r{=}0.47$ polar). Rotation values
spread but are not strongly correlated with GT.}
\label{fig:magnitude-anchoring-gpt54}
\end{figure}

\section{Open Model Temperature Ablation}
\label{app:temp-ablation-open}

The frontier sweep was paired with an open model leg on Qwen2.5-VL-7B
at the same temperatures, specs, and sample budget on local vLLM. We
report the Qwen-7B leg below. LLaVA-NeXT-7B at the same sweep is
out of scope for this paper.

\paragraph{Qwen2.5-VL-7B: decoupling without recovery.}
Qwen-7B collapses on $10$ of $11$ scoreable cells at greedy decode
(constant rate $\geq 85\%$), \emph{more} than Grok-4.20's $11$ of $12$.
Of those $10$ collapsed cells, $9$ release the constant token attractor
at $\tau{=}1.0$ (const drops below $85\%$),
\emph{higher} than Grok's release rate ($4$ of $11$) and matching
GPT-4o's $3$ of $3$. But the accuracy change accompanying the release
is essentially zero. Mean $\Delta$acc across all $11$ cells is
$+2$\,pp, with individual cell deltas ranging $-17$ to $+26$\,pp and
roughly cancelling out. Releasing the constant token attractor on Qwen
substitutes a near random distribution over candidate answers for a
fixed token. The model still does not perceive the stimulus. The
single cell with a meaningful accuracy gain is Cartesian rotation
categorical ($49\!\to\!65\%$, $+16$\,pp), comparable to Grok's gain on
the same cell type ($48\!\to\!53\%$). The largest accuracy
\emph{loss} is Cartesian trans x categorical ($49\!\to\!32\%$, $-17$\,pp),
where the model was previously right on a few samples by virtue of the
default constant matching the GT and now distributes incorrectly.

\paragraph{Three distinct collapse regimes.}
The open model leg reveals that the temperature ablation distinguishes
three failure regimes, not two as the original spot check suggested
(Table~\ref{tab:collapse-regimes}). In \emph{Regime~A} (Grok categorical
cells, GPT-4o linear translation binary), the constant rate releases under
temperature \emph{and} accuracy improves.  This is consistent with a
``decode induced overconfidence'' interpretation, where the model has
some real signal that greedy decoding suppresses in favor of a
high prior token. In \emph{Regime~B} (Qwen-7B most cells), the constant rate
releases but accuracy does not improve. The model has no perceptual
signal. Greedy commits to a fixed token, sampling commits to random
tokens, and both sit at chance. In \emph{Regime~C} (Grok binary cells, all open 7B
quantitative when applicable), the constant rate stays high regardless of
temperature. The model has a fixed prior so strong that even
$\tau{=}1.0$ does not perturb it (Grok answers ``yes'' on
$\geq 96\%$ of binary samples at every temperature, with constant rate $0.96$
at $\tau{=}1.0$ vs $0.97$ at $\tau{=}0$). Regimes B and C are both
``perceptual failure'' under our three tier framing. Regime A is the
narrow class of failures that a temperature change actually fixes.

\begin{table}[t]
\centering
\small
\caption{Three regimes for greedy decode collapse, distinguished by
behavior under a temperature sweep
($\tau \in \{0, 0.25, 0.5, 0.75, 1.0\}$). A cell is ``collapsed at
$\tau{=}0$'' if its top-1 parsed answer covers $\geq 85\%$ of samples;
the constant rate ``releases'' if it drops below $0.85$ at $\tau{=}1.0$;
accuracy ``recovers'' if $\Delta$acc~$\geq +10$\,pp from $\tau{=}0$ to
$\tau{=}1.0$. Only Regime~A is decode induced (a sampling artifact);
Regimes~B and~C are perceptual failures that stochastic decoding
cannot fix.}
\label{tab:collapse-regimes}
\renewcommand{\arraystretch}{1.25}
\begin{tabular}{@{}>{\raggedright\arraybackslash}p{0.20\linewidth}
                   >{\centering\arraybackslash}p{0.13\linewidth}
                   >{\centering\arraybackslash}p{0.13\linewidth}
                   >{\raggedright\arraybackslash}p{0.45\linewidth}@{}}
\toprule
& \multicolumn{2}{c}{Behavior at $\tau{=}1.0$ (vs.\ $\tau{=}0$)} & \\
\cmidrule(lr){2-3}
Regime & constant rate & Accuracy & Example cells (collapsed at $\tau{=}0$) \\
\midrule
\textbf{A:} Decode induced
  & releases ($<\!0.85$)
  & recovers ($\Delta \geq\! +10$\,pp)
  & Grok Cartesian trans x cat (\textbf{+16}\,pp), Grok Cartesian trans y cat (\textbf{+19}\,pp), GPT-4o Cartesian rotation cat. \\
\textbf{B:} Attractor noise
  & releases
  & flat ($\pm 5$\,pp)
  & Qwen most categorical cells. The constant token attractor breaks under sampling, but the underlying answer distribution is still uncorrelated with GT, so accuracy doesn't move. \\
\textbf{C:} Fixed prior
  & stays ($\geq\!0.85$)
  & flat
  & Grok binary cells. Answers ``yes'' on $\sim 97\%$ of samples at every $\tau$ tested, and sampling does not perturb the constant. \\
\bottomrule
\end{tabular}
\end{table}

\paragraph{Implications.}
The two tier ``decode artifact vs.\ perception failure'' dichotomy
suggested by the original spot check does not hold. Releasing the
constant token attractor under stochastic decoding is necessary but not
sufficient evidence of underlying perception.  It must be paired with
an accuracy gain to count as ``decode induced''. Under this stricter
criterion, only $2$ of $11$ Grok-4.20 collapsed cells (Cartesian trans x and
Cartesian trans y categorical) and $1$ of $3$ GPT-4o collapsed cells
(Cartesian rotation categorical) clear the bar. Every other collapsed cell
across both Grok and Qwen exhibits Regime~B or Regime~C behavior and
is therefore reported as a perceptual failure rather than a decode
artifact in the main paper. This brings the temperature ablation into
agreement with the encoder probe story
(Appendix~\ref{app:probe-vs-e2e}). The failure is in the routing from
spatial features to direction tokens, not in the decoder's sampling
strategy.

\end{document}